# Decision-Theoretic Planning with non-Markovian Rewards

**Sylvie Thiébaux** SYLVIE.THIEBAUX@ANU.EDU.AU
**Charles Gretton** CHARLES.GRETTON@ANU.EDU.AU
**John Slaney** JOHN.SLANEY@ANU.EDU.AU
**David Price** DAVID.PRICE@ANU.EDU.AU
*National ICT Australia &*
*The Australian National University*
*Canberra, ACT 0200, Australia*

**Froduald Kabanza** KABANZA@USHERBROOKE.CA
*Département d'Informatique*
*Université de Sherbrooke*
*Sherbrooke, Québec J1K 2R1, Canada*

## Abstract

A decision process in which rewards depend on history rather than merely on the current state is called a decision process with non-Markovian rewards (NMRDP). In decision-theoretic planning, where many desirable behaviours are more naturally expressed as properties of execution sequences rather than as properties of states, NMRDPs form a more natural model than the commonly adopted fully Markovian decision process (MDP) model. While the more tractable solution methods developed for MDPs do not directly apply in the presence of non-Markovian rewards, a number of solution methods for NMRDPs have been proposed in the literature. These all exploit a compact specification of the non-Markovian reward function in temporal logic, to automatically translate the NMRDP into an equivalent MDP which is solved using efficient MDP solution methods. This paper presents NMRDPP(Non-Markovian Reward Decision Process Planner), a software platform for the development and experimentation of methods for decision-theoretic planning with non-Markovian rewards. The current version of NMRDPP implements, under a single interface, a family of methods based on existing as well as new approaches which we describe in detail. These include dynamic programming, heuristic search, and structured methods. Using NMRDPP, we compare the methods and identify certain problem features that affect their performance. NMRDPP's treatment of non-Markovian rewards is inspired by the treatment of domain-specific search control knowledge in the TLPlan planner, which it incorporates as a special case. In the First International Probabilistic Planning Competition, NMRDPP was able to compete and perform well in both the domain-independent and hand-coded tracks, using search control knowledge in the latter.

# 1. Introduction

## 1.1 The Problem

Markov decision processes (MDPs) are now widely accepted as the preferred model for decision-theoretic planning problems (Boutilier, Dean, & Hanks, 1999). The fundamental assumption behind the MDP formulation is that not only the system dynamics but also the reward function are *Markovian*. Therefore, all information needed to determine the reward at a given state must be encoded in the state itself.

This requirement is not always easy to meet for planning problems, as many desirable behaviours are naturally expressed as properties of execution *sequences* (see e.g., Drummond, 1989; Haddawy & Hanks, 1992; Bacchus & Kabanza, 1998; Pistore & Traverso, 2001). Typical cases include rewards for the maintenance of some property, for the periodic achievement of some goal, for the achievement of a goal within a given number of steps of the request being made, or even simply for the very first achievement of a goal which becomes irrelevant afterwards.

For instance, consider a health care robot which assists ederly or disabled people by achieving simple goals such as reminding them to do important tasks (e.g. taking a pill), entertaining them, checking or transporting objects for them (e.g. checking the stove's temperature or bringing coffee), escorting them, or searching (e.g. for glasses or for the nurse) (Cesta et al., 2003). In this domain, we might want to reward the robot for making sure a given patient takes his pill exactly once every 8 hours (and penalise it if it fails to prevent the patient from doing this more than once within this time frame!), we may reward it for repeatedly visiting all rooms in the ward in a given order and reporting any problem it detects, it may also receive a reward once for each patient's request answered within the appropriate time-frame, etc. Another example is the elevator control domain (Koehler & Schuster, 2000), in which an elevator must get passengers from their origin to their destination as efficiently as possible, while attempting to satisfying a range of other conditions such as providing priority services to critical customers. In this domain, some trajectories of the elevator are more desirable than others, which makes it natural to encode the problem by assigning rewards to those trajectories.

A decision process in which rewards depend on the sequence of states passed through rather than merely on the current state is called a decision process with *non-Markovian rewards* (NMRDP) (Bacchus, Boutilier, & Grove, 1996). A difficulty with NMRDPs is that the most efficient MDP solution methods do not directly apply to them. The traditional way to circumvent this problem is to formulate the NMRDP as an equivalent MDP, whose states result from augmenting those of the original NMRDP with extra information capturing enough history to make the reward Markovian. Hand crafting such an MDP can however be very difficult in general. This is exacerbated by the fact that the size of the MDP impacts the effectiveness of many solution methods. Therefore, there has been interest in automating the translation into an MDP, starting from a natural specification of non-Markovian rewards and of the system's dynamics (Bacchus et al., 1996; Bacchus, Boutilier, & Grove, 1997). This is the problem we focus on.

### 1.2 Existing Approaches

When solving NMRDPs in this setting, the central issue is to define a non-Markovian reward specification language and a translation into an MDP *adapted* to the class of MDP solution methods and representations we would like to use for the type of problems at hand. More precisely, there is a tradeoff between the effort spent in the translation, e.g. in producing a *small* equivalent MDP without many irrelevant history distinctions, and the effort required to solve it. Appropriate resolution of this tradeoff depends on the type of representations and solution methods envisioned for the MDP. For instance, *structured* representations and solution methods which have some ability to ignore irrelevant information may cope with a crude translation, while *state-based* (flat) representations and methods will require a more sophisticated translation producing an MDP as small as feasible.

Both the two previous proposals within this line of research rely on past linear temporal logic (PLTL) formulae to specify the behaviours to be rewarded (Bacchus et al., 1996, 1997). A nice feature of PLTL is that it yields a straightforward semantics of non-Markovian rewards, and lends itself to a range of translations from the crudest to the finest. The two proposals adopt very different translations adapted to two very different types of solution methods and representations. The first (Bacchus et al., 1996) targets classical state-based solution methods such as policy iteration (Howard, 1960) which generate *complete* policies at the cost of enumerating all states in the entire MDP. Consequently, it adopts an expensive translation which attempts to produce a *minimal* MDP. By contrast, the second translation (Bacchus et al., 1997) is very efficient but crude, and targets structured solution methods and representations (see e.g., Hoey, St-Aubin, Hu, & Boutilier, 1999; Boutilier, Dearden, & Goldszmidt, 2000; Feng & Hansen, 2002), which do not require explicit state enumeration.

### 1.3 A New Approach

The first contribution of this paper is to provide a language and a translation adapted to another class of solution methods which have proven quite effective in dealing with large MDPs, namely *anytime* state-based heuristic search methods such as LAO* (Hansen & Zilberstein, 2001), LRTDP (Bonet & Geffner, 2003), and ancestors (Barto, Bardtke, & Singh, 1995; Dean, Kaelbling, Kirman, & Nicholson, 1995; Thiébaux, Hertzberg, Shoaff, & Schneider, 1995). These methods typically start with a compact representation of the MDP based on probabilistic planning operators, and search forward from an initial state, constructing new states by expanding the envelope of the policy as time permits. They may produce an approximate and even incomplete policy, but explicitly construct and explore only a fraction of the MDP. Neither of the two previous proposals is well-suited to such solution methods, the first because the cost of the translation (most of which is performed prior to the solution phase) annihilates the benefits of anytime algorithms, and the second because the size of the MDP obtained is an obstacle to the applicability of state-based methods. Since here both the cost of the translation and the size of the MDP it results in will severely impact on the quality of the policy obtainable by the deadline, we need an appropriate resolution of the tradeoff between the two.

Our approach has the following main features. The translation is entirely embedded in the anytime solution method, to which full control is given as to which parts of the MDP will be explicitly constructed and explored. While the MDP obtained is not minimal, it is of the minimal size achievable without stepping outside of the anytime framework, i.e., without enumerating parts of the state space that the solution method would not necessarily explore. We formalise this relaxed notion of minimality, which we call *blind minimality* in reference to the fact that it does not require any lookahead (beyond the fringe). This is appropriate in the context of anytime state-based solution methods, where we want the minimal MDP achievable without expensive pre-processing.

When the rewarding behaviours are specified in PLTL, there does not appear to be a way of achieving a relaxed notion of minimality as powerful as blind minimality without a prohibitive translation. Therefore instead of PLTL, we adopt a variant of *future* linear temporal logic (FLTL) as our specification language, which we extend to handle rewards. While the language has a more complex semantics than PLTL, it enables a natural translation into a blind-minimal MDP by simple *progression* of the reward formulae. Moreover, search control knowledge expressed in FLTL (Bacchus & Kabanza, 2000) fits particularly nicely in this framework, and can be used to dramatically reduce the fraction of the search space explored by the solution method.

### 1.4 A New System

Our second contribution is NMRDPP, the first reported implementation of NMRDP solution methods. NMRDPP is designed as a software platform for their development and experimentation under a common interface. Given a description of the actions in a domain, NMRDPP lets the user play with and compare various encoding styles for non-Markovian rewards and search control knowledge, various translations of the resulting NMRDP into MDP, and various MDP solution methods. While solving the problem, it can be made to record a range of statistics about the space and time behaviour of the algorithms. It also supports the graphical display of the MDPs and policies generated.

While NMRDPP's primary interest is in the treatment of non-Markovian rewards, it is also a competitive platform for decision-theoretic planning with purely Markovian rewards. In the First International Probabilistic Planning Competition, NMRDPP was able to enrol in both the domain-independent and hand-coded tracks, attempting all problems featuring in the contest. Thanks to its use of search control-knowledge, it scored a second place in the hand-coded track which featured probabilistic variants of blocks world and logistics problems. More surprisingly, it also scored second in the domain-independent subtrack consisting of all problems that were *not* taken from the blocks world and logistic domains. Most of these latter problems had not been released to the participants prior to the competition.

### 1.5 A New Experimental Analysis

Our third contribution is an experimental analysis of the factors that affect the performance of NMRDP solution methods. Using NMRDPP, we compared their behaviours under the influence of parameters such as the structure and degree of uncertainty in the dynamics, the type of rewards and the syntax used to described them, reachability of the conditions tracked, and relevance of rewards to the optimal policy. We were able to identify a number of general trends in the behaviours of the methods and provide advice concerning which are best suited in certain circumstances. Our experiments also lead us to rule out one of the methods as systematically underperforming, and to identify issues with the claim of minimality made by one of the PLTL approaches.

### 1.6 Organisation of the Paper

The paper is organised as follows. Section 2 begins with background material on MDPs, NMRDPs, and existing approaches. Section 3 describes our new approach and Section 4 presents NMRDPP. Sections 5 and 6 report our experimental analysis of the various approaches. Section 7 explains how we used NMRDPP in the competition. Section 8 concludes with remarks about related and future work. Appendix B gives the proofs of the theorems. Most of the material presented is compiled from a series of recent conference and workshop papers (Thiébaux, Kabanza, & Slaney, 2002a, 2002b; Gretton, Price, & Thiébaux, 2003a, 2003b). Details of the logic we use to represent rewards may be found in our 2005 paper (Slaney, 2005).

## 2. Background

### 2.1 MDPs, NMRDPs, Equivalence

We start with some notation and definitions. Given a finite set $S$ of states, we write $S^*$ for the set of finite sequences of states over $S$, and $S^\omega$ for the set of possibly infinite state sequences. Where '$\Gamma$' stands for a possibly infinite state sequence in $S^\omega$ and $i$ is a natural number, by '$\Gamma_i$' we mean the state of index $i$ in $\Gamma$, by '$\Gamma(i)$' we mean the prefix $\langle \Gamma_0, \ldots, \Gamma_i \rangle \in S^*$ of $\Gamma$. $\Gamma; \Gamma'$ denotes the concatenation of $\Gamma \in S^*$ and $\Gamma' \in S^\omega$.

#### 2.1.1 MDPs

A Markov decision process of the type we consider is a 5-tuple $\langle S, s_0, A, \Pr, R \rangle$, where $S$ is a finite set of fully observable states, $s_0 \in S$ is the initial state, $A$ is a finite set of actions ($A(s)$ denotes the subset of actions applicable in $s \in S$), $\{\Pr(s, a, \bullet) \mid s \in S, a \in A(s)\}$ is a family of probability distributions over $S$, such that $\Pr(s, a, s')$ is the probability of being in state $s'$ after performing action $a$ in state $s$, and $R : S \mapsto \mathbb{R}$ is a reward function such that $R(s)$ is the immediate reward for being in state $s$. It is well known that such an MDP can be compactly represented using dynamic Bayesian networks (Dean & Kanazawa, 1989; Boutilier et al., 1999) or probabilistic extensions of traditional planning languages (see e.g., Kushmerick, Hanks, & Weld, 1995; Thiébaux et al., 1995; Younes & Littman, 2004).

A stationary policy for an MDP is a function $\pi : S \mapsto A$, such that $\pi(s) \in A(s)$ is the action to be executed in state $s$. The value $V_\pi$ of the policy at $s_0$, which we seek to maximise, is the sum of the expected future rewards over an infinite horizon, discounted by how far into the future they occur:

$$V_\pi(s_0) = \lim_{n \to \infty} \mathsf{E} \left[ \sum_{i=0}^{n} \beta^i R(\Gamma_i) \mid \pi, \Gamma_0 = s_0 \right]$$

where $0 \leq \beta < 1$ is the discount factor controlling the contribution of distant rewards.

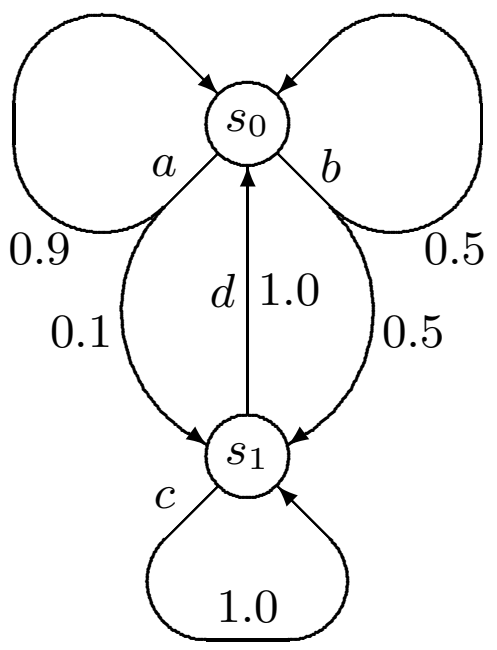

In the initial state $s_0$, $p$ is false and two actions are possible: $a$ causes a transition to $s_1$ with probability 0.1, and no change with probability 0.9, while for $b$ the transition probabilities are 0.5. In state $s_1$, $p$ is true, and actions $c$ and $d$ ("stay" and "go") lead to $s_1$ and $s_0$ respectively with probability 1.
A reward is received the first time $p$ is true, but not subsequently. That is, the rewarded state sequences are:

$\langle s_0, s_1 \rangle$
$\langle s_0, s_0, s_1 \rangle$
$\langle s_0, s_0, s_0, s_1 \rangle$
$\langle s_0, s_0, s_0, s_0, s_1 \rangle$ etc.

Figure 1: A Simple NMRDP

#### 2.1.2 NMRDPs

A decision process with non-Markovian rewards is identical to an MDP except that the domain of the reward function is $S^*$. The idea is that if the process has passed through state sequence $\Gamma(i)$ up to stage $i$, then the reward $R(\Gamma(i))$ is received at stage $i$. Figure 1 gives an example. Like the reward function, a policy for an NMRDP depends on history, and is a mapping from $S^*$ to $A$. As before, the value of policy $\pi$ is the expectation of the discounted cumulative reward over an infinite horizon:

$$V_\pi(s_0) = \lim_{n\to\infty} \mathsf{E}\left[\sum_{i=0}^{n} \beta^i R(\Gamma(i)) \mid \pi, \Gamma_0 = s_0\right]$$

For a decision process $D = \langle S, s_0, A, \Pr, R\rangle$ and a state $s \in S$, we let $\widetilde{D}(s)$ stand for the set of state sequences rooted at $s$ that are *feasible* under the actions in $D$, that is: $\widetilde{D}(s) = \{\Gamma \in S^\omega \mid \Gamma_0 = s \text{ and } \forall i\ \exists a \in A(\Gamma_i) \quad \Pr(\Gamma_i, a, \Gamma_{i+1}) > 0\}$. Note that the definition of $\widetilde{D}(s)$ does not depend on $R$ and therefore applies to both MDPs and NMRDPs.

#### 2.1.3 Equivalence

The clever algorithms developed to solve MDPs cannot be directly applied to NMRDPs. One way of dealing with this problem is to translate the NMRDP into an equivalent MDP with an expanded state space (Bacchus et al., 1996). The expanded states in this MDP (*e-states*, for short) augment the states of the NMRDP by encoding additional information sufficient to make the reward history-independent. For instance, if we only want to reward the very first achievement of goal $g$ in an NMRDP, the states of an equivalent MDP would carry one extra bit of information recording whether $g$ has already been true. An e-state can be seen as labelled by a state of the NMRDP (via the function $\tau$ in Definition 1 below) and by history information. The dynamics of NMRDPs being Markovian, the actions and their probabilistic effects in the MDP are exactly those of the NMRDP. The following definition, adapted from that given by Bacchus et al. (1996), makes this concept of equivalent MDP precise. Figure 2 gives an example.

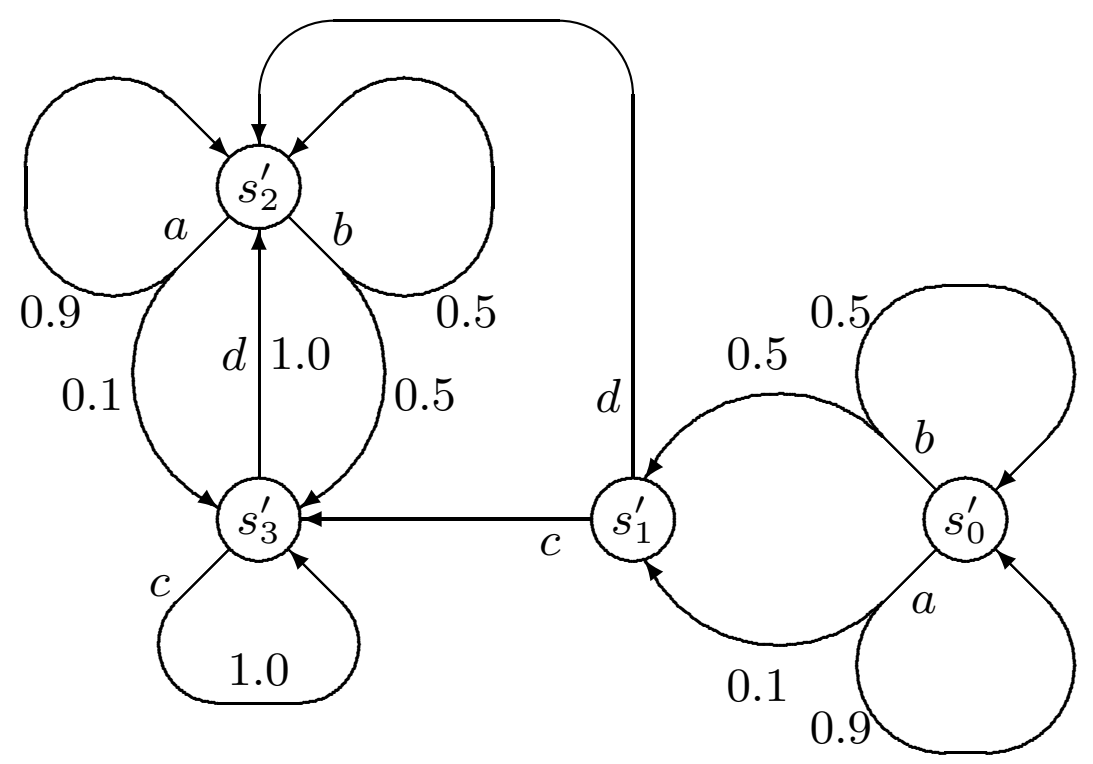

Figure 2: An MDP Equivalent to the NMRDP in Figure 1. $\tau(s'_0) = \tau(s'_2) = s_0$. $\tau(s'_1) = \tau(s'_3) = s_1$. The initial state is $s'_0$. State $s'_1$ is rewarded; the other states are not.

**Definition 1** *MDP $D'=\langle S', s'_0, A', \Pr', R'\rangle$ is equivalent to NMRDP $D = \langle S, s_0, A, \Pr, R\rangle$ if there exists a mapping $\tau : S' \mapsto S$ such that:*[1]

1. $\tau(s'_0) = s_0$.
2. *For all $s' \in S'$, $A'(s') = A(\tau(s'))$.*
3. *For all $s_1, s_2 \in S$, if there is $a \in A(s_1)$ such that $\Pr(s_1, a, s_2) > 0$, then for all $s'_1 \in S'$ such that $\tau(s'_1) = s_1$, there exists a unique $s'_2 \in S'$, $\tau(s'_2) = s_2$, such that for all $a' \in A'(s'_1)$, $\Pr'(s'_1, a', s'_2) = \Pr(s_1, a', s_2)$.*
4. *For any feasible state sequence $\Gamma \in \widetilde{D}(s_0)$ and $\Gamma' \in \widetilde{D}'(s'_0)$ such that $\tau(\Gamma'_i) = \Gamma_i$ for all $i$, we have: $R'(\Gamma'_i) = R(\Gamma(i))$ for all $i$.*

Items 1–3 ensure that there is a bijection between feasible state sequences in the NMRDP and feasible e-state sequences in the MDP. Therefore, a stationary policy for the MDP can be reinterpreted as a non-stationary policy for the NMRDP. Furthermore, item 4 ensures that the two policies have identical values, and that consequently, solving an NMRDP optimally reduces to producing an equivalent MDP and solving it optimally (Bacchus et al., 1996):

**Proposition 1** *Let $D$ be an NMRDP, $D'$ an equivalent MDP for it, and $\pi'$ a policy for $D'$. Let $\pi$ be the function defined on the sequence prefixes $\Gamma(i) \in \widetilde{D}(s_0)$ by $\pi(\Gamma(i)) = \pi'(\Gamma'_i)$, where for all $j \leq i$ $\tau(\Gamma'_j) = \Gamma_j$. Then $\pi$ is a policy for $D$ such that $V_\pi(s_0) = V_{\pi'}(s'_0)$.*

1. Technically, the definition allows the sets of actions $A$ and $A'$ to be different, but any action in which they differ must be inapplicable in reachable states in the NMRDP and in all e-states in the equivalent MDP. For all practical purposes, $A$ and $A'$ can be seen as identical.

### 2.2 Existing Approaches

Both existing approaches to NMRDPs (Bacchus et al., 1996, 1997) use a temporal logic of the past (PLTL) to compactly represent non-Markovian rewards and exploit this compact representation to translate the NMRDP into an MDP amenable to off-the-shelf solution methods. However, they target different classes of MDP representations and solution methods, and consequently, adopt different styles of translations.

Bacchus et al. (1996) target state-based MDP representations. The equivalent MDP is first generated entirely – this involves the enumeration of all e-states and all transitions between them. Then, it is solved using traditional dynamic programming methods such as value or policy iteration. Because these methods are extremely sensitive to the number of states, attention is paid to producing a minimal equivalent MDP (with the least number of states). A first simple translation which we call pltlsim produces a large MDP which can be post-processed for minimisation before being solved. Another, which we call pltlmin, directly results in a minimal MDP, but relies on an expensive pre-processing phase.

The second approach (Bacchus et al., 1997), which we call pltlstr, targets structured MDP representations: the transition model, policies, reward and value functions are represented in a compact form, e.g. as trees or algebraic decision diagrams (ADDs) (Hoey et al., 1999; Boutilier et al., 2000). For instance, the probability of a given proposition (state variable) being true after the execution of an action is specified by a tree whose internal nodes are labelled with the state variables on whose previous values the given variable depends, whose arcs are labelled by the possible previous values ($\top$ or $\bot$) of these variables, and whose leaves are labelled with probabilities. The translation amounts to augmenting the structured MDP with new *temporal* variables tracking the relevant properties of state sequences, together with the compact representation of (1) their dynamics, e.g. as trees over the previous values of relevant variables, and (2) of the non-Markovian reward function in terms of the variables' current values. Then, structured solution methods such as structured policy iteration or the SPUDD algorithm are run on the resulting structured MDP. Neither the translation nor the solution methods explicitly enumerates the states.

We now review these approaches in some detail. The reader is referred to the respective papers for additional information.

#### 2.2.1 Representing Rewards with PLTL

The syntax of PLTL, the language chosen to represent rewarding behaviours, is that of propositional logic, augmented with the operators $\ominus$ (previously) and $\mathsf{S}$ (since) (see Emerson, 1990). Whereas a classical propositional logic formula denotes a set of states (a subset of $S$), a PLTL formula denotes a set of finite *sequences* of states (a subset of $S^*$). A formula without temporal modality expresses a property that must be true of the current state, i.e., the last state of the finite sequence. $\ominus f$ specifies that $f$ holds in the previous state (the state one before the last). $f_1 \mathsf{S} f_2$, requires $f_2$ to have been true at some point in the sequence, and, unless that point is the present, $f_1$ to have held ever since. More formally, the modelling relation $\models$ stating whether a formula $f$ holds of a finite sequence $\Gamma(i)$ is defined recursively as follows:

- $\Gamma(i) \models p$ iff $p \in \Gamma_i$, for $p \in \mathcal{P}$, the set of atomic propositions

- $\Gamma(i) \models \neg f$ iff $\Gamma(i) \not\models f$
- $\Gamma(i) \models f_1 \wedge f_2$ iff $\Gamma(i) \models f_1$ and $\Gamma(i) \models f_2$
- $\Gamma(i) \models \ominus f$ iff $i > 0$ and $\Gamma(i-1) \models f$
- $\Gamma(i) \models f_1 \mathsf{S} f_2$ iff $\exists j \leq i, \Gamma(j) \models f_2$ and $\forall k, j < k \leq i, \Gamma(k) \models f_1$

From $\mathsf{S}$, one can define the useful operators $\diamondminus f \equiv \top \mathsf{S} f$ meaning that $f$ has been true at some point, and $\boxminus f \equiv \neg \diamondminus \neg f$ meaning that $f$ has always been true. E.g, $g \wedge \neg \ominus \diamondminus g$ denotes the set of finite sequences ending in a state where $g$ is true for the first time in the sequence. Other useful abbreviation are $\ominus^k$ ($k$ times ago), for $k$ iterations of the $\ominus$ modality, $\diamondminus_k f$ for $\vee_{i=1}^{k} \ominus^i f$ ($f$ was true at some of the $k$ last steps), and $\boxminus_k f$ for $\wedge_{i=1}^{k} \ominus^i f$ ($f$ was true at all the $k$ last steps).

Non-Markovian reward functions are described with a set of pairs $(f_i : r_i)$ where $f_i$ is a PLTL reward formula and $r_i$ is a real, with the semantics that the reward assigned to a sequence in $S^*$ is the sum of the $r_i$'s for which that sequence is a model of $f_i$. Below, we let $F$ denote the set of reward formulae $f_i$ in the description of the reward function. Bacchus et al. (1996) give a list of behaviours which it might be useful to reward, together with their expression in PLTL. For instance, where $f$ is an atemporal formula, $(f : r)$ rewards with $r$ units the achievement of $f$ whenever it happens. This is a Markovian reward. In contrast $(\diamondminus f : r)$ rewards every state following (and including) the achievement of $f$, while $(f \wedge \neg \ominus \diamondminus f : r)$ only rewards the first occurrence of $f$. $(f \wedge \boxminus_k \neg f : r)$ rewards the occurrence of $f$ at most once every $k$ steps. $(\ominus^n \neg \ominus \perp : r)$ rewards the $n^{\text{th}}$ state, independently of its properties. $(\ominus^2 f_1 \wedge \ominus f_2 \wedge f_3 : r)$ rewards the occurrence of $f_1$ immediately followed by $f_2$ and then $f_3$. In reactive planning, so-called *response* formulae which describe that the achievement of $f$ is triggered by a condition (or command) $c$ are particularly useful. These can be written as $(f \wedge \diamondminus c : r)$ if every state in which $f$ is true following the first issue of the command is to be rewarded. Alternatively, they can be written as $(f \wedge \ominus(\neg f \mathsf{S} c) : r)$ if only the first occurrence of $f$ is to be rewarded after each command. It is common to only reward the achievement $f$ within $k$ steps of the trigger; we write for example $(f \wedge \diamondminus_k c : r)$ to reward all such states in which $f$ holds.

From a theoretical point of view, it is known (Lichtenstein, Pnueli, & Zuck, 1985) that the behaviours representable in PLTL are exactly those corresponding to star-free regular languages. Non star-free behaviours such as $(pp)*$ (reward an even number of states all containing $p$) are therefore not representable. Nor, of course, are non-regular behaviours such as $p^n q^n$ (e.g. reward taking equal numbers of steps to the left and right). We shall not speculate here on how severe a restriction this is for the purposes of planning.

### 2.2.2 Principles Behind the Translations

All three translations into an MDP (pltlsim, pltlmin, and pltlstr) rely on the equivalence $f_1 \mathsf{S} f_2 \equiv f_2 \vee (f_1 \wedge \ominus(f_1 \mathsf{S} f_2))$, with which we can decompose temporal modalities into a requirement about the last state $\Gamma_i$ of a sequence $\Gamma(i)$, and a requirement about the prefix $\Gamma(i-1)$ of the sequence. More precisely, given state $s$ and a formula $f$, one can com-pute in[2] $\mathcal{O}(||f||)$ a new formula $\text{Reg}(f, s)$ called the regression of $f$ through $s$. Regression has the property that, for $i > 0$, $f$ is true of a finite sequence $\Gamma(i)$ ending with $\Gamma_i = s$ iff $\text{Reg}(f, s)$ is true of the prefix $\Gamma(i-1)$. That is, $\text{Reg}(f, s)$ represents what must have been true previously for $f$ to be true now. Reg is defined as follows:

- $\text{Reg}(p, s) = \top$ iff $p \in s$ and $\bot$ otherwise, for $p \in \mathcal{P}$
- $\text{Reg}(\neg f, s) = \neg\text{Reg}(f, s)$
- $\text{Reg}(f_1 \wedge f_2, s) = \text{Reg}(f_1, s) \wedge \text{Reg}(f_2, s)$
- $\text{Reg}(\circleddash f, s) = f$
- $\text{Reg}(f_1 \mathsf{S} f_2, s) = \text{Reg}(f_2, s) \vee (\text{Reg}(f_1, s) \wedge (f_1 \mathsf{S} f_2))$

For instance, take a state $s$ in which $p$ holds and $q$ does not, and take $f = (\circleddash\neg q) \wedge (p \mathsf{S} q)$, meaning that $q$ must have been false 1 step ago, but that it must have held at some point in the past and that $p$ must have held since $q$ last did. $\text{Reg}(f, s) = \neg q \wedge (p \mathsf{S} q)$, that is, for $f$ to hold now, then at the previous stage, $q$ had to be false and the $p \mathsf{S} q$ requirement still had to hold. When $p$ and $q$ are both false in $s$, then $\text{Reg}(f, s) = \bot$, indicating that $f$ cannot be satisfied, regardless of what came earlier in the sequence.

For notational convenience, where $X$ is a set of formulae we write $\overline{X}$ for $X \cup \{\neg x \mid x \in X\}$. Now the translations exploit the PLTL representation of rewards as follows. Each expanded state (e-state) in the generated MDP can be seen as labelled with a set $\Psi \subseteq \overline{\text{Sub}(F)}$ of subformulae of the reward formulae in $F$ (and their negations). The subformulae in $\Psi$ must be (1) true of the paths leading to the e-state, and (2) sufficient to determine the current truth of all reward formulae in $F$, as this is needed to compute the current reward. Ideally the $\Psi$s should also be (3) small enough to enable just that, i.e. they should not contain subformulae which draw history distinctions that are irrelevant to determining the reward at one point or another. Note however that in the worst-case, the number of distinctions needed, even in the minimal equivalent MDP, may be exponential in $||F||$. This happens for instance with the formula $\circleddash^k f$, which requires $k$ additional bits of information memorising the truth of $f$ over the last $k$ steps.

#### 2.2.3 pltlsim

For the choice of the $\Psi$s, Bacchus et al. (1996) consider two cases. In the simple case, which we call pltlsim, an MDP obeying properties (1) and (2) is produced by simply labelling each e-state with the set of *all* subformulae in $\overline{\text{Sub}(F)}$ which are true of the sequence leading to that e-state. This MDP is generated forward, starting from the initial e-state labelled with $s_0$ and with the set $\Psi_0 \subseteq \overline{\text{Sub}(F)}$ of all subformulae which are true of the sequence $\langle s_0 \rangle$. The successors of any e-state labelled by NMRDP state $s$ and subformula set $\Psi$ are generated as follows: each of them is labelled by a successor $s'$ of $s$ in the NMRDP and by the set of subformulae $\{\psi' \in \overline{\text{Sub}(F)} \mid \Psi \models \text{Reg}(\psi', s')\}$.

For instance, consider the NMRDP shown in Figure 3. The set $F = \{q \wedge \circleddash\circleddash p\}$ consists of a single reward formula. The set $\overline{\text{Sub}(F)}$ consists of all subformulae of this reward formula,

2. The size $||f||$ of a reward formula is measured as its length and the size $||F||$ of a set of reward formulae $F$ is measured as the sum of the lengths of the formulae in $F$.

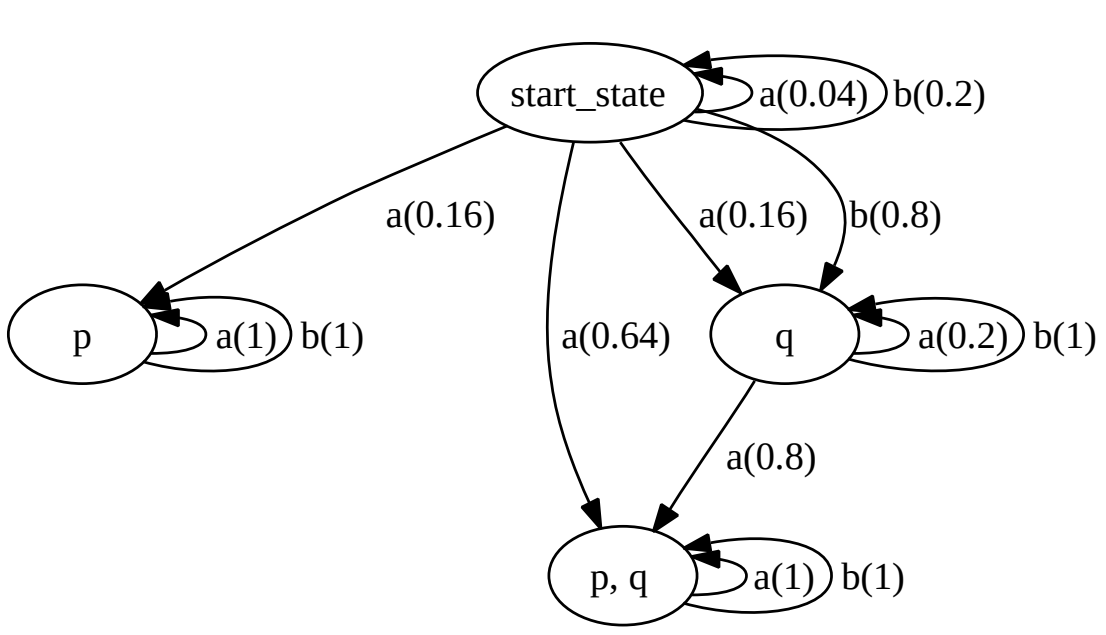

In the initial state, both $p$ and $q$ are false. When $p$ is false, action $a$ independently sets $p$ and $q$ to true with probability 0.8. When both $p$ and $q$ are false, action $b$ sets $q$ to true with probability 0.8. Both actions have no effect otherwise. A reward is obtained whenever $q \wedge \ominus\ominus p$. The optimal policy is to apply $b$ until $q$ gets produced, making sure to avoid the state on the left-hand side, then to apply $a$ until $p$ gets produced, and then to apply $a$ or $b$ indifferently forever.

Figure 3: Another Simple NMRDP

and their negations, that is $\overline{\mathrm{Sub}(F)} = \{p, q, \ominus p, \ominus\ominus p, q \wedge \ominus\ominus p, \neg p, \neg q, \neg\ominus p, \neg\ominus\ominus p, \neg(q \wedge \ominus\ominus p)\}$. The equivalent MDP produced by PLTLSIM is shown in Figure 4.

### 2.2.4 PLTLMIN

Unfortunately, the MDPs produced by PLTLSIM are far from minimal. Although they could be postprocessed for minimisation before invoking the MDP solution method, the above expansion may still constitute a serious bottleneck. Therefore, Bacchus et al. (1996) consider a more complex two-phase translation, which we call PLTLMIN, capable of producing an MDP also satisfying property (3). Here, a preprocessing phase iterates over all states in $S$, and computes, for each state $s$, a set $l(s)$ of subformulae, where the function $l$ is the solution of the fixpoint equation $l(s) = F \cup \{\mathrm{Reg}(\psi', s') \mid \psi' \in l(s'), s' \text{ is a successor of } s\}$. Only subformulae in $\overline{l(s)}$ will be candidates for inclusion in the sets labelling the respective e-states labelled with $s$. That is, the subsequent expansion phase will be as above, but taking $\Psi_0 \subseteq \overline{l(s_0)}$ and $\psi' \subseteq \overline{l(s')}$ instead of $\Psi_0 \subseteq \overline{\mathrm{Sub}(F)}$ and $\psi' \subseteq \overline{\mathrm{Sub}(F)}$. As the subformulae in $l(s)$ are exactly those that are relevant to the way feasible execution sequences starting from e-states labelled with $s$ are rewarded, this leads the expansion phase to produce a minimal equivalent MDP.

Figure 5 shows the equivalent MDP produced by PLTLMIN for the NMRDP example in Figure 3, together with the function $l$ from which the labels are built. Observe how this MDP is smaller than the PLTLSIM MDP: once we reach the state on the left-hand side in which $p$ is true and $q$ is false, there is no point in tracking the values of subformulae, because $q$ cannot become true and so the reward formula cannot either. This is reflected by the fact that $l(\{p\})$ only contains the reward formula.

In the worst case, computing $l$ requires a space, and a number of iterations through $S$, exponential in $||F||$. Hence the question arises of whether the gain during the expansion phase is worth the extra complexity of the preprocessing phase. This is one of the questions our experimental analysis in Section 5 will try to answer.

### 2.2.5 PLTLSTR

The PLTLSTR translation can be seen as a symbolic version of PLTLSIM. The set $T$ of added temporal variables contains the purely temporal subformulae $\mathrm{PTSub}(F)$ of the reward formulae in $F$, to which the $\ominus$ modality is prepended (unless already there): $T = \{\ominus\psi \mid \psi \in$

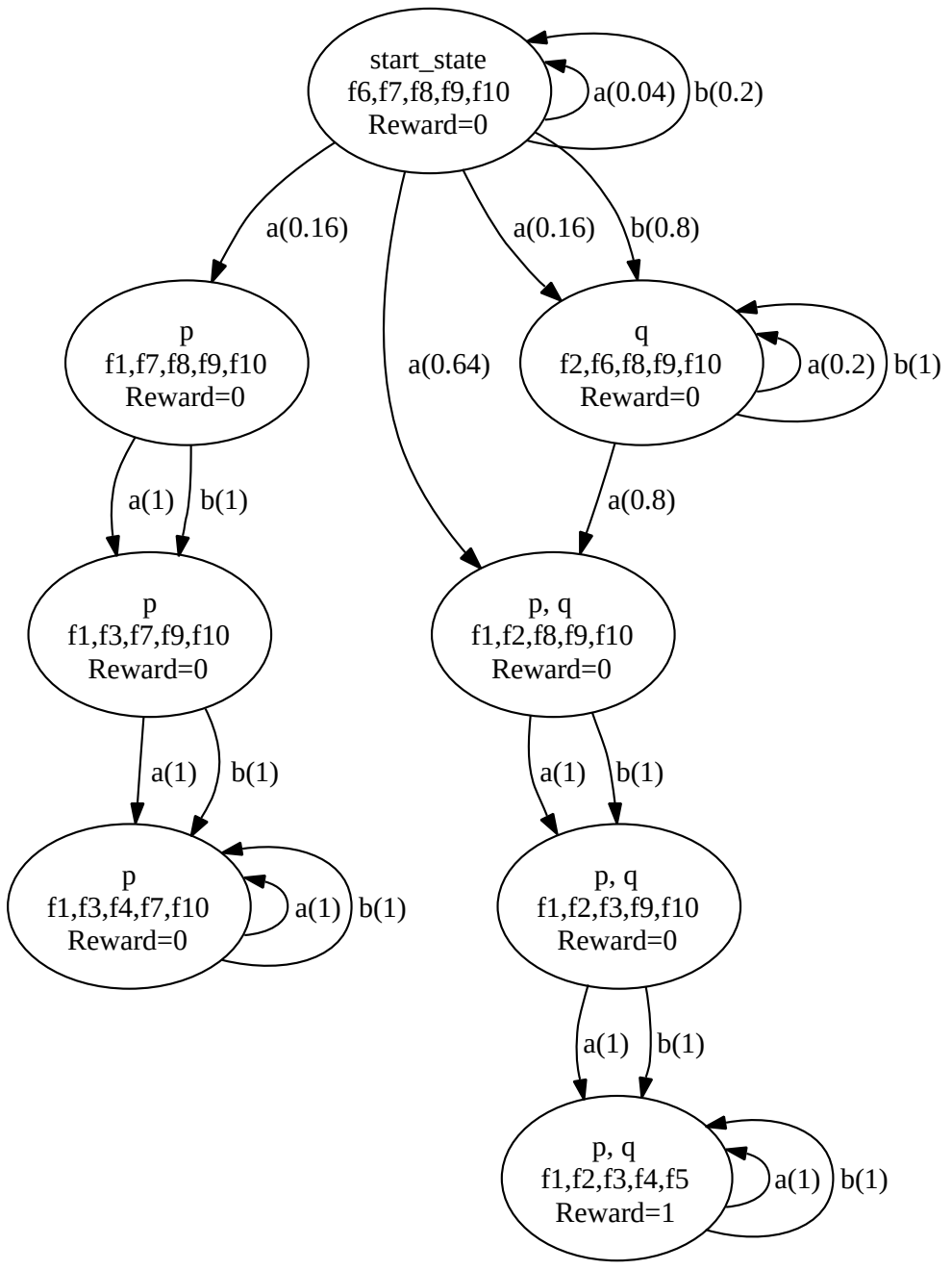

The following subformulae in $\overline{\text{Sub}(F)}$ label the e-states:
$f_1 : p$
$f_2 : q$
$f_3 : \ominus p$
$f_4 : \ominus \ominus p$
$f_5 : q \wedge \ominus \ominus p$
$f_6 : \neg p$
$f_7 : \neg q$
$f_8 : \neg \ominus p$
$f_9 : \neg \ominus \ominus p$
$f_{10} : \neg(q \wedge \ominus \ominus p)$

Figure 4: Equivalent MDP Produced by PLTLSIM

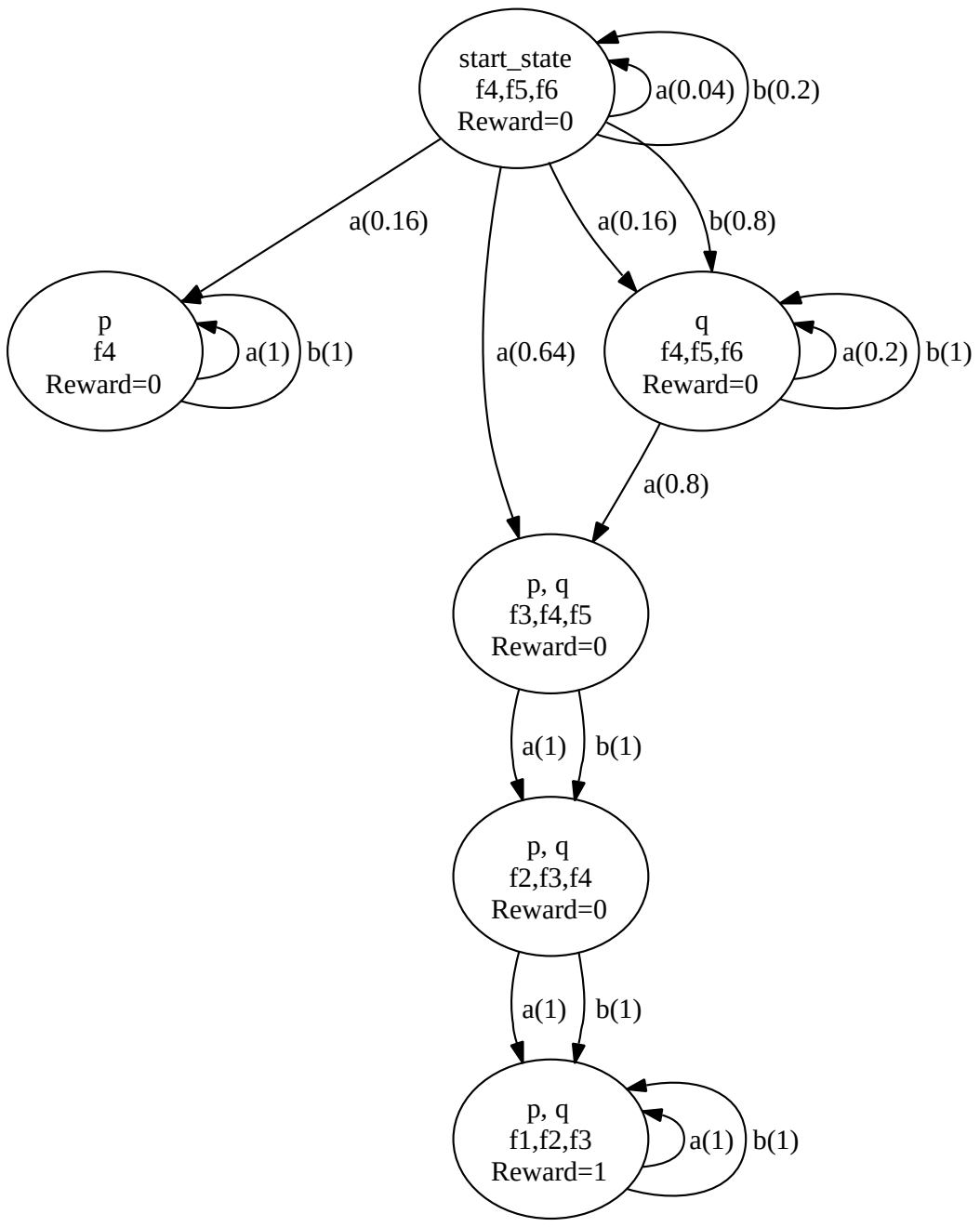

The function $l$ is given by:
$l(\{\}) = \{q \wedge \ominus \ominus p, \ominus p, p\}$
$l(\{p\}) = \{q \wedge \ominus \ominus p\}$
$l(\{q\}) = \{q \wedge \ominus \ominus p, \ominus p, p\}$
$l(\{p, q\}) = \{q \wedge \ominus \ominus p, \ominus p, p\}$

The following formulae label the e-states:
$f_1 : q \wedge \ominus \ominus p$
$f_2 : \ominus p$
$f_3 : p$
$f_4 : \neg(q \wedge \ominus \ominus p)$
$f_5 : \neg \ominus p$
$f_6 : \neg p$

Figure 5: Equivalent MDP Produced by PLTLMIN

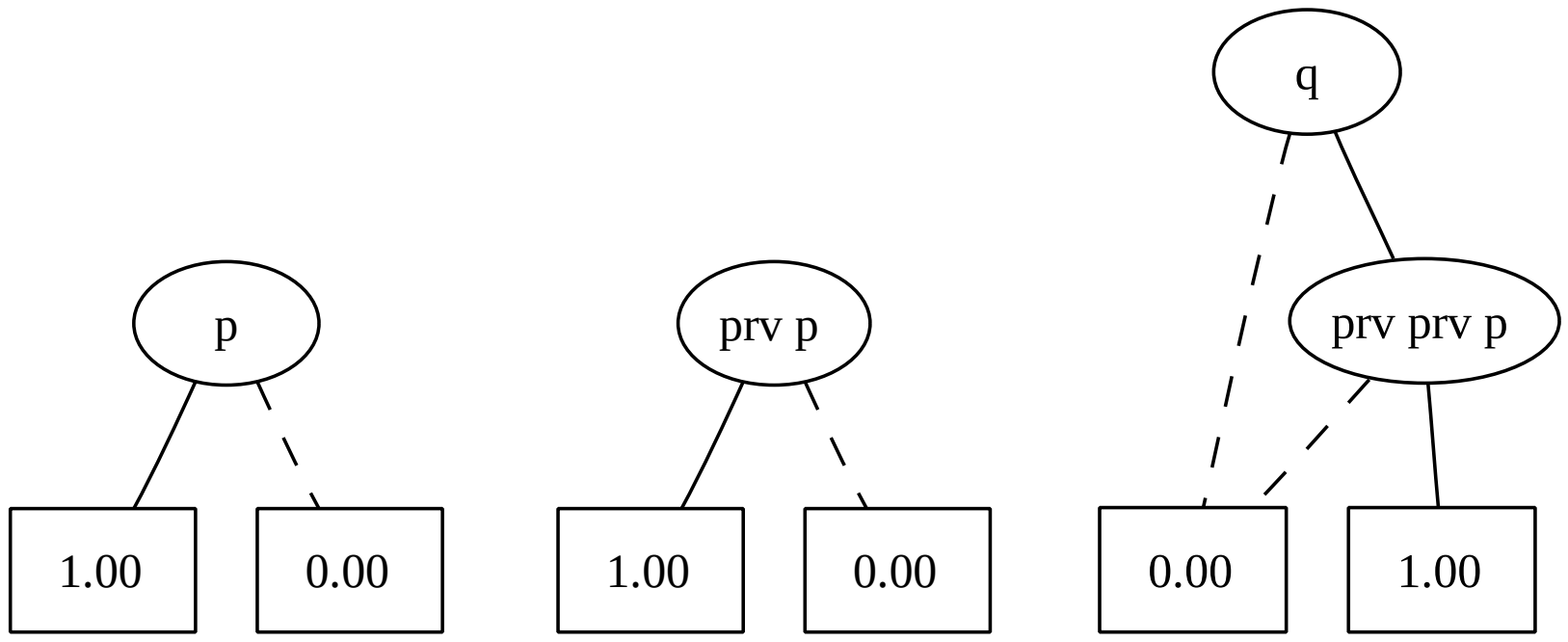

1. dynamics of $\ominus p$  2. dynamics of $\ominus\ominus p$  3. reward

Figure 6: ADDs Produced by PLTLSTR. prv (previously) stands for $\ominus$

$\text{PTSub}(F), \psi \neq \ominus\psi'\} \cup \{\ominus\psi \mid \ominus\psi \in \text{PTSub}(F)\}$. By repeatedly applying the equivalence $f_1 \mathsf{S} f_2 \equiv f_2 \vee (f_1 \wedge \ominus(f_1 \mathsf{S} f_2))$ to any subformula in PTSub($F$), we can express its current value, and hence that of reward formulae, as a function of the current values of formulae in $T$ and state variables, as required by the compact representation of the transition and reward models.

For our NMRDP example in Figure 3, the set of purely temporal variables is $\text{PTSub}(F) = \{\ominus p, \ominus\ominus p\}$, and $T$ is identical to PTSub($F$). Figure 6 shows some of the ADDs forming part of the symbolic MDP produced by PLTLSTR: the ADDs describing the dynamics of the temporal variables, i.e., the ADDs describing the effects of the actions $a$ and $b$ on their respective values, and the ADD describing the reward.

As a more complex illustration, consider this example (Bacchus et al., 1997) in which

$$F = \{\diamondminus(p\,\mathsf{S}\ (q \vee \ominus r))\} \equiv \{\top\,\mathsf{S}\ (p\,\mathsf{S}\ (q \vee \ominus r))\}$$

We have that

$$\text{PTSub}(F) = \{\top\,\mathsf{S}\ (p\,\mathsf{S}\ (q \vee \ominus r)), p\,\mathsf{S}\ (q \vee \ominus r), \ominus r\}$$

and so the set of temporal variables used is

$$T = \{t_1 : \ominus(\top\,\mathsf{S}\ (p\,\mathsf{S}\,(q \vee \ominus r))), t_2 : \ominus(p\,\mathsf{S}\ (q \vee \ominus r)), t3 : \ominus r\}$$

Using the equivalences, the reward can be decomposed and expressed by means of the propositions $p, q$ and the temporal variables $t_1, t_2, t_3$ as follows:

$$\begin{array}{ll} \top\,\mathsf{S}\ (p\,\mathsf{S}\ (q \vee \ominus r)) & \equiv \\ (p\,\mathsf{S}\ (q \vee \ominus r)) \vee \ominus(\top\,\mathsf{S}\ (p\,\mathsf{S}\ (q \vee \ominus r))) & \equiv \\ (q \vee \ominus r) \vee (p \wedge \ominus(p\,\mathsf{S}\ (q \vee \ominus r))) \vee t_1 & \equiv \\ (q \vee t_3) \vee (p \wedge t_2) \vee t_1 & \end{array}$$

As with PLTLSIM, the underlying MDP produced by PLTLSTR is far from minimal – the encoded history features do not even vary from one state to the next. However, size is not as problematic as with state-based approaches, because structured solution methods do not enumerate states and are able to dynamically ignore some of the variables that become irrelevant during policy construction. For instance, when solving the MDP, they may be able to determine that some temporal variables have become irrelevant because the situation they track, although possible in principle, is too costly to be realised under a good policy. This *dynamic* analysis of rewards contrast with pltlmin's *static* analysis (Bacchus et al., 1996) which must encode enough history to determine the reward at all reachable future states under any policy.

One question that arises is that of the circumstances under which this analysis of irrelevance by structured solution methods, especially the dynamic aspects, is really effective. This is another question our experimental analysis will try to address.

## 3. fltl: A Forward-Looking Approach

As noted in Section 1 above, the two key issues facing approaches to NMRDPs are how to specify the reward functions compactly and how to exploit this compact representation to automatically translate an NMRDP into an equivalent MDP amenable to the chosen solution method. Accordingly, our goals are to provide a reward function specification language and a translation that are adapted to anytime state-based solution methods. After a brief reminder of the relevant features of these methods, we consider these two goals in turn. We describe the syntax and semantics of the language, the notion of formula progression for the language which will form the basis of our translation, the translation itself, its properties, and its embedding into the solution method. We call our approach fltl. We finish the section with a discussion of the features that distinguish fltl from existing approaches.

### 3.1 Anytime State-Based Solution Methods

The main drawback of traditional dynamic programming algorithms such as policy iteration (Howard, 1960) is that they explicitly enumerate all states that are reachable from $s_0$ in the entire MDP. There has been interest in other state-based solution methods, which may produce incomplete policies, but only enumerate a fraction of the states that policy iteration requires.

Let $E(\pi)$ denote the *envelope* of policy $\pi$, that is the set of states that are reachable (with a non-zero probability) from the initial state $s_0$ under the policy. If $\pi$ is defined at all $s \in E(\pi)$, we say that the policy is *complete*, and that it is incomplete otherwise. The set of states in $E(\pi)$ at which $\pi$ is undefined is called the *fringe* of the policy. The fringe states are taken to be absorbing, and their value is heuristic. A common feature of anytime state-based algorithms is that they perform a forward search, starting from $s_0$ and repeatedly expanding the envelope of the current policy one step forward by adding one or more fringe states. When provided with admissible heuristic values for the fringe states, they eventually converge to the optimal policy without necessarily needing to explore the entire state space. In fact, since planning operators are used to compactly represent the state space, they may not even need to *construct* more than a small subset of the MDP before returning the optimal policy. When interrupted before convergence, they return a possibly incomplete but often useful policy.

These methods include the envelope expansion algorithm (Dean et al., 1995), which deploys policy iteration on judiciously chosen larger and larger envelopes, using each successive policy to seed the calculation of the next. The more recent LAO$^*$ algorithm (Hansen

& Zilberstein, 2001) which combines dynamic programming with heuristic search can be viewed as a clever implementation of a particular case of the envelope expansion algorithm, where fringe states are given admissible heuristic values, where policy iteration is run up to convergence between envelope expansions, and where the clever implementation only runs policy iteration on the states whose optimal value can actually be affected when a new fringe state is added to the envelope. Another example is a backtracking forward search in the space of (possibly incomplete) policies rooted at $s_0$ (Thiébaux et al., 1995), which is performed until interrupted, at which point the best policy found so far is returned. Real-time dynamic programming (RTDP) (Barto et al., 1995) is another popular anytime algorithm which is to MDPs what learning real-time A* (Korf, 1990) is to deterministic domains, and which has asymptotic convergence guarantees. The RTDP envelope is made up of sample paths which are visited with a frequency determined by the current greedy policy and the transition probabilities in the domain. RTDP can be run on-line, off-line for a given number of steps or until interrupted. A variant called LRTDP (Bonet & Geffner, 2003) incorporates mechanisms that focus the search on states whose value has not yet converged, resulting in convergence speed up and finite time convergence guarantees.

The FLTL translation we are about to present targets these anytime algorithms, although it could also be used with more traditional methods such as value and policy iteration.

### 3.2 Language and Semantics

Compactly representing non-Markovian reward functions reduces to compactly representing the behaviours of interest, where by *behaviour* we mean a set of finite sequences of states (a subset of $S^*$), e.g. the $\{\langle s_0, s_1\rangle, \langle s_0, s_0, s_1\rangle, \langle s_0, s_0, s_0, s_1\rangle \ldots\}$ in Figure 1. Recall that the reward is issued at the end of any prefix $\Gamma(i)$ in that set. Once behaviours are compactly represented, it is straightforward to represent non-Markovian reward functions as mappings from behaviours to real numbers – we shall defer looking at this until Section 3.6.

To represent behaviours compactly, we adopt a version of future linear temporal logic (FLTL) (see Emerson, 1990), augmented with a propositional constant '\$', intended to be read 'The behaviour we want to reward has just happened' or 'The reward is received now'. The language \$FLTL begins with a set of basic propositions $\mathcal{P}$ giving rise to literals:

$$\mathcal{L} ::= \mathcal{P} \mid \neg\mathcal{P} \mid \top \mid \bot \mid \$$$

where $\top$ and $\bot$ stand for 'true' and 'false', respectively. The connectives are classical $\wedge$ and $\vee$, and the temporal modalities $\bigcirc$ (next) and $\mathsf{U}$ (*weak* until), giving formulae:

$$\mathcal{F} ::= \mathcal{L} \mid \mathcal{F} \wedge \mathcal{F} \mid \mathcal{F} \vee \mathcal{F} \mid \bigcirc\mathcal{F} \mid \mathcal{F}\,\mathsf{U}\,\mathcal{F}$$

Our 'until' is weak: $f_1 \mathsf{U} f_2$ means $f_1$ will be true from now on until $f_2$ is, if ever. Unlike the more commonly used strong 'until', this does not imply that $f_2$ will eventually be true. It allows us to define the useful operator $\square$ (always): $\square f \equiv f \mathsf{U} \bot$ ($f$ will always be true from now on). We also adopt the notations $\bigcirc^k f$ for $k$ iterations of the $\bigcirc$ modality (f will be true in exactly $k$ steps), $\lozenge_k f$ for $\bigvee_{i=1}^{k} \bigcirc^i f$ (f will be true within the next $k$ steps), and $\square_k f$ for $\bigwedge_{i=1}^{k} \bigcirc^i f$ ($f$ will be true throughout the next $k$ steps).

Although negation officially occurs only in literals, i.e., the formulae are in negation normal form (NNF), we allow ourselves to write formulae involving it in the usual way, provided that they have an equivalent in NNF. Not every formula has such an equivalent, because there is no such literal as ¬$ and because eventualities ('f will be true some time') are not expressible. These restrictions are deliberate. If we were to use our notation and logic to *theorise* about the allocation of rewards, we would indeed need the means to say when rewards are not received or to express features such as liveness ('always, there will be a reward eventually'), but in fact we are using them only as a mechanism for ensuring that rewards are given where they should be, and for this restricted purpose eventualities and the negated dollar are not needed. In fact, including them would create technical difficulties in relating formulae to the behaviours they represent.

The semantics of this language is similar to that of FLTL, with an important difference: because the interpretation of the constant $ depends on the behaviour $B$ we want to reward (whatever that is), the modelling relation $\models$ must be indexed by $B$. We therefore write $(\Gamma, i) \models_B f$ to mean that formula $f$ holds at the $i$-th stage of an arbitrary sequence $\Gamma \in S^\omega$, relative to behaviour $B$. Defining $\models_B$ is the first step in our description of the semantics:

$(\Gamma, i) \models_B \$$ iff $\Gamma(i) \in B$

$(\Gamma, i) \models_B \top$

$(\Gamma, i) \not\models_B \bot$

$(\Gamma, i) \models_B p$, for $p \in \mathcal{P}$, iff $p \in \Gamma_i$

$(\Gamma, i) \models_B \neg p$, for $p \in \mathcal{P}$, iff $p \notin \Gamma_i$

$(\Gamma, i) \models_B f_1 \wedge f_2$ iff $(\Gamma, i) \models_B f_1$ and $(\Gamma, i) \models_B f_2$

$(\Gamma, i) \models_B f_1 \vee f_2$ iff $(\Gamma, i) \models_B f_1$ or $(\Gamma, i) \models_B f_2$

$(\Gamma, i) \models_B \bigcirc f$ iff $(\Gamma, i+1) \models_B f$

$(\Gamma, i) \models_B f_1 \mathsf{U} f_2$ iff $\forall k \geq i$ if $(\forall j, i \leq j \leq k \; (\Gamma, j) \not\models_B f_2)$ then $(\Gamma, k) \models_B f_1$

Note that except for subscript $B$ and for the first rule, this is just the standard FLTL semantics, and that therefore $-free formulae keep their FLTL meaning. As with FLTL, we say $\Gamma \models_B f$ iff $(\Gamma, 0) \models_B f$, and $\models_B f$ iff $\Gamma \models_B f$ for all $\Gamma \in S^\omega$.

The modelling relation $\models_B$ can be seen as specifying when a formula holds, on which reading it takes $B$ as input. Our next and final step is to use the $\models_B$ relation to define, for a formula $f$, the behaviour $B_f$ that it represents, and for this we must rather *assume* that $f$ holds, and then *solve* for $B$. For instance, let $f$ be $\Box(p \rightarrow \$)$, i.e., we get rewarded every time $p$ is true. We would like $B_f$ to be the set of all finite sequences ending with a state containing $p$. For an arbitrary $f$, we take $B_f$ to be the set of prefixes that *have* to be rewarded if $f$ is to hold in all sequences:

**Definition 2** $B_f \;\equiv\; \bigcap\{B \mid \;\models_B f\}$

To understand Definition 2, recall that $B$ contains prefixes at the end of which we get a reward and $ evaluates to true. Since $f$ is supposed to describe the way rewards will be received in an *arbitrary* sequence, we are interested in behaviours $B$ which make $ true in such a way as to make $f$ hold without imposing constraints on the evolution of the world. However, there may be many behaviours with this property, so we take their

intersection,[3] ensuring that $B_f$ will only reward a prefix if it has to because that prefix is in *every* behaviour satisfying $f$. In all but pathological cases (see Section 3.4), this makes $B_f$ coincide with the (set-inclusion) minimal behaviour $B$ such that $\models_B f$. The reason for this 'stingy' semantics, making rewards minimal, is that $f$ does not actually say that rewards are allocated to more prefixes than are required for its truth. For instance, $\Box(p \rightarrow \$)$ *says* only that a reward is given every time $p$ is true, even though a more generous distribution of rewards would be *consistent* with it.

### 3.3 Examples

It is intuitively clear that many behaviours can be specified by means of $FLTL formulae. While there is no simple way in general to translate between past and future tense expressions,[4] all of the examples used to illustrate PLTL in Section 2.2 above are expressible naturally in $FLTL, as follows.

The classical goal formula $g$ saying that a goal $p$ is rewarded whenever it happens is easily expressed: $\Box(p \rightarrow \$)$. As already noted, $B_g$ is the set of finite sequences of states such that $p$ holds in the last state. If we only care that $p$ is achieved once and get rewarded at each state from then on, we write $\Box(p \rightarrow \Box\$)$. The behaviour that this formula represents is the set of finite state sequences having at least one state in which $p$ holds. By contrast, the formula $\neg p \,\mathsf{U}\, (p \wedge \$)$ stipulates only that the first occurrence of $p$ is rewarded (i.e. it specifies the behaviour in Figure 1). To reward the occurrence of $p$ at most once every $k$ steps, we write $\Box((\bigcirc^{k+1} p \wedge \Box_k \neg p) \rightarrow \bigcirc^{k+1}\$)$.

For response formulae, where the achievement of $p$ is triggered by the command $c$, we write $\Box(c \rightarrow \bigcirc\Box(p \rightarrow \$))$ to reward every state in which $p$ is true following the first issue of the command. To reward only the first occurrence $p$ after each command, we write $\Box(c \rightarrow \bigcirc(\neg p \,\mathsf{U}\, (p \wedge \$)))$. As for bounded variants for which we only reward goal achievement within $k$ steps of the trigger command, we write for example $\Box(c \rightarrow \Box_k(p \rightarrow \$))$ to reward all such states in which $p$ holds.

It is also worth noting how to express simple behaviours involving past tense operators. To stipulate a reward if $p$ has always been true, we write $\$ \,\mathsf{U}\, \neg p$. To say that we are rewarded if $p$ has been true since $q$ was, we write $\Box(q \rightarrow (\$ \,\mathsf{U}\, \neg p))$.

Finally, we often find it useful to reward the holding of $p$ until the occurrence of $q$. The neatest expression for this is $\neg q \,\mathsf{U}\, ((\neg p \wedge \neg q) \vee (q \wedge \$))$.

### 3.4 Reward Normality

$FLTL is therefore quite expressive. Unfortunately, it is rather *too* expressive, in that it contains formulae which describe "unnatural" allocations of rewards. For instance, they may make rewards depend on future behaviours rather than on the past, or they may

3. If there is no $B$ such that $\models_B f$, which is the case for any \$-free $f$ which is not a logical theorem, then $B_f$ is $\bigcap \emptyset$ – i.e. $S^*$ following normal set-theoretic conventions. This limiting case does no harm, since \$-free formulae do not describe the attribution of rewards.

4. It is an open question whether the set of representable behaviours is the same for $FLTL as for PLTL, that is star-free regular languages. Even if the behaviours were the same, there is little hope that a practical translation from one to the other exists.

leave open a choice as to which of several behaviours is to be rewarded.[5] An example of dependence on the future is $\bigcirc p \rightarrow \$$, which stipulates a reward *now* if $p$ is going to hold *next*. We call such formula *reward-unstable*. What a reward-stable $f$ amounts to is that whether a particular prefix needs to be rewarded in order to make $f$ true does not depend on the future of the sequence. An example of an open choice of which behavior to reward is $\Box(p \rightarrow \$) \vee \Box(\neg p \rightarrow \$)$ which says we should *either* reward all achievements of the goal $p$ *or* reward achievements of $\neg p$ but does not determine which. We call such formula *reward-indeterminate*. What a reward-determinate $f$ amounts to is that the set of behaviours modelling $f$, i.e. $\{B \mid \models_B f\}$, has a unique minimum. If it does not, $B_f$ is insufficient (too small) to make $f$ true.

In investigating \$FLTL (Slaney, 2005), we examine the notions of reward-stability and reward-determinacy in depth, and motivate the claim that formulae that are both reward-stable and reward-determinate – we call them *reward-normal* – are precisely those that capture the notion of "no funny business". This is the intuition that we ask the reader to note, as it will be needed in the rest of the paper. Just for reference then, we define:

**Definition 3** *$f$ is* reward-normal *iff for every $\Gamma \in S^\omega$ and every $B \subseteq S^*$, $\Gamma \models_B f$ iff for every $i$, if $\Gamma(i) \in B_f$ then $\Gamma(i) \in B$.*

The property of reward-normality is decidable (Slaney, 2005). In Appendix A we give some simple syntactic constructions guaranteed to result in reward-normal formulae. While reward-abnormal formulae may be interesting, for present purposes we restrict attention to reward-normal ones. Indeed, we stipulate as part of our method that only reward-normal formulae should be used to represent behaviours. Naturally, all formulae in Section 3.3 are normal.

### 3.5 \$FLTL Formula Progression

Having defined a language to represent behaviours to be rewarded, we now turn to the problem of computing, given a reward formula, a minimum allocation of rewards to states actually encountered in an execution sequence, in such a way as to satisfy the formula. Because we ultimately wish to use anytime solution methods which generate state sequences incrementally via forward search, this computation is best done on the fly, while the sequence is being generated. We therefore devise an incremental algorithm based on a model-checking technique normally used to check whether a state sequence is a model of an FLTL formula (Bacchus & Kabanza, 1998). This technique is known as formula *progression* because it 'progresses' or 'pushes' the formula through the sequence.

Our progression technique is shown in Algorithm 1. In essence, it computes the modelling relation $\models_B$ given in Section 3.2. However,unlike the definition of $\models_B$, it is designed to be useful when states in the sequence become available one at a time, in that it defers the evaluation of the part of the formula that refers to the future to the point where the next state becomes available. Let $s$ be a state, say $\Gamma_i$, the last state of the sequence prefix $\Gamma(i)$

5. These difficulties are inherent in the use of linear-time formalisms in contexts where the principle of directionality must be enforced. They are shared for instance by formalisms developed for reasoning about actions such as the Event Calculus and LTL action theories (see e.g., Calvanese, De Giacomo, & Vardi, 2002).

that has been generated so far, and let $b$ be a boolean true iff $\Gamma(i)$ is in the behaviour $B$ to be rewarded. Let the \$FLTL formula $f$ describe the allocation of rewards over all possible futures. Then the progression of $f$ through $s$ given $b$, written $\text{Prog}(b, s, f)$, is a new formula which will describe the allocation of rewards over all possible futures of the *next* state, given that we have just passed through $s$. Crucially, the function Prog is Markovian, depending only on the current state and the single boolean value $b$. Note that Prog is computable in linear time in the length of $f$, and that for \$-free formulae, it collapses to FLTL formula progression (Bacchus & Kabanza, 1998), regardless of the value of $b$. We assume that Prog incorporates the usual simplification for sentential constants $\bot$ and $\top$: $f \wedge \bot$ simplifies to $\bot$, $f \wedge \top$ simplifies to $f$, etc.

**Algorithm 1** \$FLTL Progression

| | | |
|---|---|---|
| $\text{Prog}(\text{true}, s, \$)$ | $=$ | $\top$ |
| $\text{Prog}(\text{false}, s, \$)$ | $=$ | $\bot$ |
| $\text{Prog}(b, s, \top)$ | $=$ | $\top$ |
| $\text{Prog}(b, s, \bot)$ | $=$ | $\bot$ |
| $\text{Prog}(b, s, p)$ | $=$ | $\top$ iff $p \in s$ and $\bot$ otherwise |
| $\text{Prog}(b, s, \neg p)$ | $=$ | $\top$ iff $p \notin s$ and $\bot$ otherwise |
| $\text{Prog}(b, s, f_1 \wedge f_2)$ | $=$ | $\text{Prog}(b, s, f_1) \wedge \text{Prog}(b, s, f_2)$ |
| $\text{Prog}(b, s, f_1 \vee f_2)$ | $=$ | $\text{Prog}(b, s, f_1) \vee \text{Prog}(b, s, f_2)$ |
| $\text{Prog}(b, s, \bigcirc f)$ | $=$ | $f$ |
| $\text{Prog}(b, s, f_1 \mathsf{U} f_2)$ | $=$ | $\text{Prog}(b, s, f_2) \vee (\text{Prog}(b, s, f_1) \wedge f_1 \mathsf{U} f_2)$ |
| $\text{Rew}(s, f)$ | $=$ | true iff $\text{Prog}(\text{false}, s, f) = \bot$ |
| $\$\text{Prog}(s, f)$ | $=$ | $\text{Prog}(\text{Rew}(s, f), s, f)$ |

The fundamental property of Prog is the following. Where $b \Leftrightarrow (\Gamma(i) \in B)$:

**Property 1** $(\Gamma, i) \models_B f$ *iff* $(\Gamma, i+1) \models_B \text{Prog}(b, \Gamma_i, f)$

**Proof:** See Appendix B. □

Like $\models_B$, the function Prog seems to require $B$ (or at least $b$) as input, but of course when progression is applied in practice we only have $f$ and one new state at a time of $\Gamma$, and what we really want to do is *compute* the appropriate $B$, namely that represented by $f$. So, similarly as in Section 3.2, we now turn to the second step, which is to use Prog to decide on the fly whether a newly generated sequence prefix $\Gamma(i)$ is in $B_f$ and so should be allocated a reward. This is the purpose of the functions \$Prog and Rew, also given in Algorithm 1. Given $\Gamma$ and $f$, the function \$Prog in Algorithm 1 defines an infinite sequence of formulae $\langle f_0, f_1, \ldots \rangle$ in the obvious way:

$$f_0 = f$$
$$f_{i+1} = \$\text{Prog}(\Gamma_i, f_i)$$

To decide whether a prefix $\Gamma(i)$ of $\Gamma$ is to be rewarded, Rew first tries progressing the formula $f_i$ through $\Gamma_i$ with the boolean flag set to 'false'. If that gives a consistent result, we need not reward the prefix and we continue without rewarding $\Gamma(i)$, but if the result is $\bot$ then we know that $\Gamma(i)$ must be rewarded in order for $\Gamma$ to satisfy $f$. In that case, to obtain $f_{i+1}$ we must progress $f_i$ through $\Gamma_i$ again, this time with the boolean flag set to the value 'true'. To sum up, the behaviour corresponding to $f$ is $\{\Gamma(i)|\text{Rew}(\Gamma_i, f_i)\}$.

To illustrate the behaviour of \$FLTL progression, consider the formula $f = \neg p \,\mathsf{U}\, (p \wedge \$)$ stating that a reward will be received the first time $p$ is true. Let $s$ be a state in which $p$ holds, then $\text{Prog}(\text{false}, s, f) = \bot \vee (\bot \wedge \neg p \,\mathsf{U}\, (p \wedge \$)) \equiv \bot$. Therefore, since the formula has progressed to $\bot$, $\text{Rew}(s, f)$ is true and a reward is received. $\$\text{Prog}(s, f) = \text{Prog}(\text{true}, s, f) = \top \vee (\bot \wedge \neg p \,\mathsf{U}\, (p \wedge \$)) \equiv \top$, so the reward formula fades away and will not affect subsequent progression steps. If, on the other hand, $p$ is false in $s$, then $\text{Prog}(\text{false}, s, f) = \bot \vee (\top \wedge \neg p \,\mathsf{U}\, (p \wedge \$)) \equiv \neg p \,\mathsf{U}\, (p \wedge \$))$. Therefore, since the formula has not progressed to $\bot$, $\text{Rew}(s, f)$ is false and no reward is received. $\$\text{Prog}(s, f) = \text{Prog}(\text{false}, s, f) = \neg p \,\mathsf{U}\, (p \wedge \$)$, so the reward formula persists as is for subsequent progression steps.

The following theorem states that under weak assumptions, rewards are correctly allocated by progression:

**Theorem 1** *Let $f$ be reward-normal, and let $\langle f_0, f_1, \ldots\rangle$ be the result of progressing it through the successive states of a sequence $\Gamma$ using the function \$Prog. Then, provided no $f_i$ is $\bot$, for all $i$ $\text{Rew}(\Gamma_i, f_i)$ iff $\Gamma(i) \in B_f$.*

**Proof:** See Appendix B □

The premise of the theorem is that $f$ never progresses to $\bot$. Indeed if $f_i = \bot$ for some $i$, it means that even rewarding $\Gamma(i)$ does not suffice to make $f$ true, so something must have gone wrong: at some earlier stage, the boolean Rew was made false where it should have been made true. The usual explanation is that the original $f$ was not reward-normal. For instance $\bigcirc p \rightarrow \$$, which is reward unstable, progresses to $\bot$ in the next state if p is true there: regardless of $\Gamma_0$, $f_0 = \bigcirc p \rightarrow \$ = \bigcirc \neg p \vee \$$, $\text{Rew}(\Gamma_0, f_0) = \text{false}$, and $f_1 = \neg p$, so if $p \in \Gamma_1$ then $f_2 = \bot$. However, other (admittedly bizarre) possibilities exist: for example, although $\bigcirc p \rightarrow \$$ is reward-unstable, its substitution instance $\bigcirc\bigcirc\top \rightarrow \$$, which also progresses to $\bot$ in a few steps, is logically equivalent to \$ and is reward-normal.

If the progression method were to deliver the correct minimal behaviour in all cases (even in all reward-normal cases) it would have to backtrack on the choice of values for the boolean flags. In the interest of efficiency, we choose not to allow backtracking. Instead, our algorithm raises an exception whenever a reward formula progresses to $\bot$, and informs the user of the sequence which caused the problem. The onus is thus placed on the domain modeller to select sensible reward formulae so as to avoid possible progression to $\bot$. It should be noted that in the worst case, detecting reward-normality cannot be easier than the decision problem for \$FLTL so it is not to be expected that there will be a simple syntactic criterion for reward-normality. In practice, however, commonsense precautions such as avoiding making rewards depend explicitly on future tense expressions suffice to keep things normal in all routine cases. For a generous class of syntactically recognisable reward-normal formulae, see Appendix A.

### 3.6 Reward Functions

With the language defined so far, we are able to compactly represent behaviours. The extension to a non-Markovian reward function is straightforward. We represent such a function by a set[6] $\phi \subseteq \$FLTL \times \mathbb{R}$ of formulae associated with real valued rewards. We call $\phi$ a *reward function specification.* Where formula $f$ is associated with reward $r$ in $\phi$, we write '$(f : r) \in \phi$'. The rewards are assumed to be independent and additive, so that the reward function $R_\phi$ represented by $\phi$ is given by:

**Definition 4** $R_\phi(\Gamma(i)) = \sum_{(f:r)\in\phi} \{r \mid \Gamma(i) \in B_f\}$

E.g, if $\phi$ is $\{\neg p \,\mathsf{U}\, (p \wedge \$) : 5.2, \Box(q \rightarrow \Box\$) : 7.3\}$, we get a reward of 5.2 the first time that $p$ holds, a reward of 7.3 from the first time that $q$ holds onwards, a reward of 12.5 when both conditions are met, and 0 otherwise.

Again, we can progress a reward function specification $\phi$ to compute the reward at all stages i of $\Gamma$. As before, progression defines a sequence $\langle\phi_0, \phi_1, \ldots\rangle$ of reward function specifications, with $\phi_{i+1} = \text{RProg}(\Gamma_i, \phi_i)$, where RProg is the function that applies Prog to all formulae in a reward function specification:

$$\text{RProg}(s, \phi) = \{(\text{Prog}(s, f) : r) \mid (f : r) \in \phi\}$$

Then, the total reward received at stage $i$ is simply the sum of the real-valued rewards granted by the progression function to the behaviours represented by the formulae in $\phi_i$:

$$\sum_{(f:r)\in\phi_i} \{r \mid \text{Rew}(\Gamma_i, f)\}$$

By proceeding that way, we get the expected analog of Theorem 1, which states progression correctly computes non-Markovian reward functions:

**Theorem 2** *Let $\phi$ be a reward-normal[7] reward function specification, and let $\langle\phi_0, \phi_1 \ldots\rangle$ be the result of progressing it through the successive states of a sequence $\Gamma$ using the function* RProg. *Then, provided $(\bot : r) \notin \phi_i$ for any $i$, then $\sum_{(f:r)\in\phi_i} \{r \mid \text{Rew}(\Gamma_i, f)\} = R_\phi(\Gamma(i))$.*

**Proof:** Immediate from Theorem 1. □

### 3.7 Translation Into MDP

We now exploit the compact representation of a non-Markovian reward function as a reward function specification to translate an NMRDP into an equivalent MDP amenable to state-based anytime solution methods. Recall from Section 2 that each e-state in the MDP is labelled by a state of the NMRDP and by history information sufficient to determine the immediate reward. In the case of a compact representation as a reward function specification $\phi_0$, this additional information can be summarised by the progression of $\phi_0$ through the sequence of states passed through. So an e-state will be of the form $\langle s, \phi\rangle$, where $s \in S$ is

6. Strictly speaking, a multiset, but for convenience we represent it as a set, with the rewards for multiple occurrences of the same formula in the multiset summed.
7. We extend the definition of reward-normality to reward specification functions in the obvious way, by requiring that all reward formulae involved be reward normal.

a state, and $\phi \subseteq \$FLTL \times \mathbb{R}$ is a reward function specification (obtained by progression). Two e-states $\langle s, \phi \rangle$ and $\langle t, \psi \rangle$ are equal if $s = t$, the immediate rewards are the same, and the results of progressing $\phi$ and $\psi$ through $s$ are semantically equivalent.[8]

**Definition 5** *Let $D = \langle S, s_0, A, \Pr, R \rangle$ be an NMRDP, and $\phi_0$ be a reward function specification representing $R$ (i.e., $R_{\phi_0} = R$, see Definition 4). We translate $D$ into the MDP $D' = \langle S', s'_0, A', \Pr', R' \rangle$ defined as follows:*

1. $S' \subseteq S \times 2^{\$FLTL \times \mathbb{R}}$
2. $s'_0 = \langle s_0, \phi_0 \rangle$
3. $A'(\langle s, \phi \rangle) = A(s)$
4. *If* $a \in A'(\langle s, \phi \rangle)$, *then* $\Pr'(\langle s, \phi \rangle, a, \langle s', \phi' \rangle) = \begin{cases} \Pr(s, a, s') & \textit{if } \phi' = \mathrm{RProg}(s, \phi) \\ 0 & \textit{otherwise} \end{cases}$
   *If* $a \notin A'(\langle s, \phi \rangle)$, *then* $\Pr'(\langle s, \phi \rangle, a, \bullet)$ *is undefined*
5. $R'(\langle s, \phi \rangle) = \sum_{(f:r) \in \phi} \{ r \mid \mathrm{Rew}(s, f) \}$
6. *For all* $s' \in S'$, $s'$ *is reachable under* $A'$ *from* $s'_0$.

Item 1 says that the e-states are labelled by a state and a reward function specification. Item 2 says that the initial e-state is labelled with the initial state and with the original reward function specification. Item 3 says that an action is applicable in an e-state if it is applicable in the state labelling it. Item 4 explains how successor e-states and their probabilities are computed. Given an action $a$ applicable in an e-state $\langle s, \phi \rangle$, each successor e-state will be labelled by a successor state $s'$ of $s$ via $a$ in the NMRDP and by the progression of $\phi$ through $s$. The probability of that e-state is $\Pr(s, a, s')$ as in the NMRDP. Note that the cost of computing $\Pr'$ is linear in that of computing $\Pr$ and in the sum of the lengths of the formulae in $\phi$. Item 5 has been motivated before (see Section 3.6). Finally, since items 1–5 leave open the choice of many MDPs differing only in the unreachable states they contain, item 6 excludes all such irrelevant extensions. It is easy to show that this translation leads to an equivalent MDP, as defined in Definition 1. Obviously, the function $\tau$ required for Definition1 is given by $\tau(\langle s, \phi \rangle) = s$, and then the proof is a matter of checking conditions.

In our practical implementation, the labelling is one step ahead of that in the definition: we label the initial e-state with $\mathrm{RProg}(s_0, \phi_0)$ and compute the current reward and the current reward specification label by progression of predecessor reward specifications through the current state rather than through the predecessor states. As will be apparent below, this has the potential to reduce the number of states in the generated MDP.

Figure 7 shows the equivalent MDP produced for the \$FLTL version of our NMRDP example in Figure 3. Recall that for this example, the PLTL reward formula was $q \wedge \ominus \ominus p$. In \$FLTL, the allocation of rewards is described by $\Box((p \wedge \bigcirc\bigcirc q) \rightarrow \bigcirc\bigcirc \$)$. The figure also

8. Care is needed over the notion of ‘semantic equivalence’. Because rewards are additive, determining equivalence may involve arithmetic as well as theorem proving. For example, the reward function specification $\{(p \rightarrow \$ : 3), (q \rightarrow \$ : 2)\}$ is equivalent to $\{((p \wedge q) \rightarrow \$ : 5), ((p \wedge \neg q) \rightarrow \$ : 3), ((\neg p \wedge q) \rightarrow \$ : 2)\}$ although there is no one-one correspondence between the formulae in the two sets.

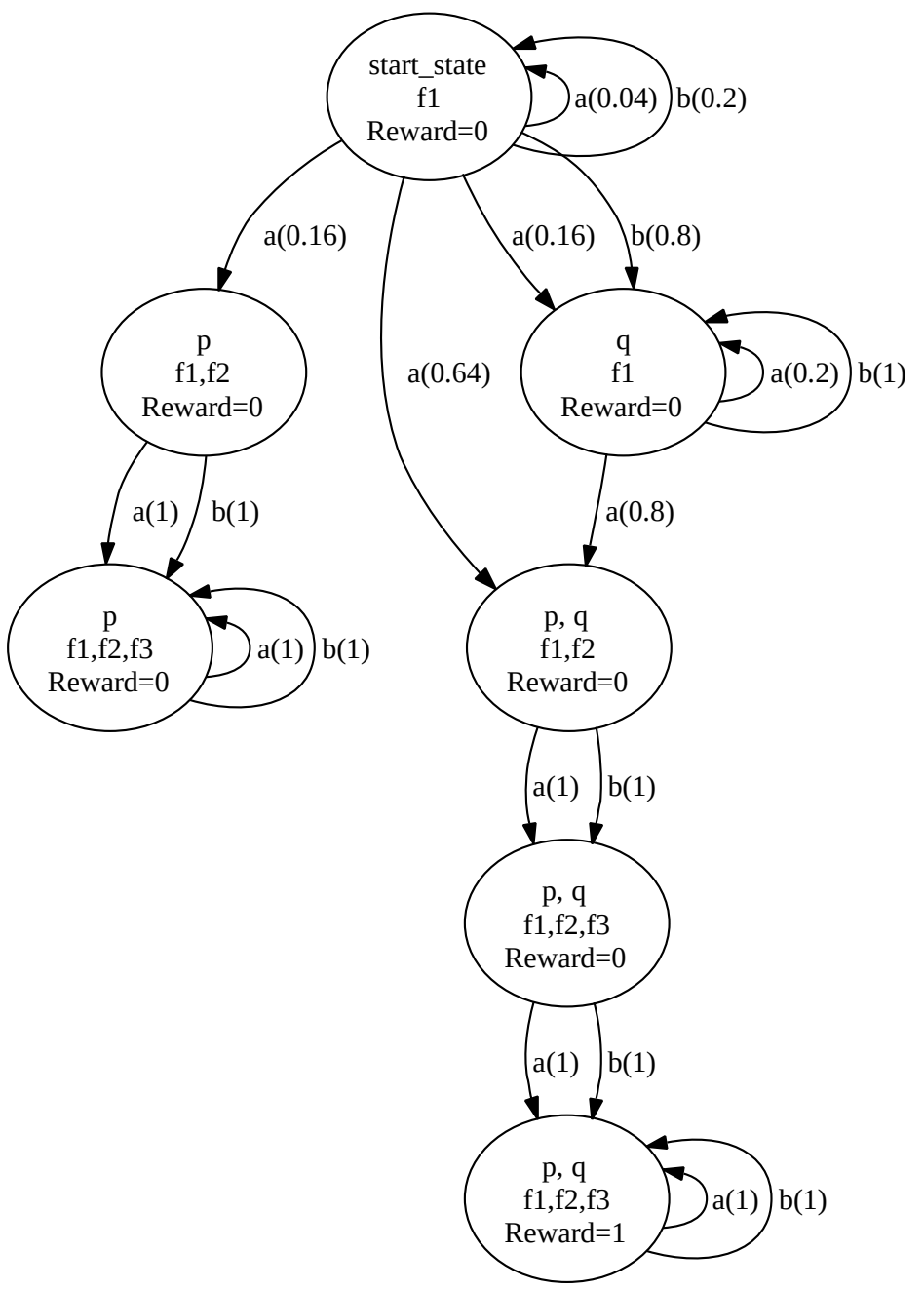

The following formulae label the e-states:
$f_1 : \Box((p \wedge \bigcirc\bigcirc q) \rightarrow \bigcirc\bigcirc \$)$
$f_2 : \bigcirc q \rightarrow \bigcirc \$$
$f_3 : q \rightarrow \$$

Figure 7: Equivalent MDP Produced by fltl

shows the relevant formulae labelling the e-states, obtained by progression of this reward formula. Note that without progressing one step ahead, there would be 3 e-states with state $\{p\}$ on the left-hand side, labelled with $\{f_1\}$, $\{f_1, f_2\}$, and $\{f_1, f_2, f_3\}$, respectively.

### 3.8 Blind Minimality

The size of the MDP obtained, i.e. the number of e-states it contains is a key issue for us, as it has to be amenable to state-based solution methods. Ideally, we would like the MDP to be of minimal size. However, we do not know of a method building the *minimal* equivalent MDP incrementally, adding parts as required by the solution method. And since in the worst case even the minimal equivalent MDP can be larger than the NMRDP by a factor exponential in the length of the reward formulae (Bacchus et al., 1996), constructing it entirely would nullify the interest of anytime solution methods.

However, as we now explain, Definition 5 leads to an equivalent MDP exhibiting a relaxed notion of minimality, and which is amenable to incremental construction. By inspection, we may observe that wherever an e-state $\langle s, \phi \rangle$ has a successor $\langle s', \phi' \rangle$ via action $a$, this means that in order to succeed in rewarding the behaviours described in $\phi$ by means of execution sequences that start by going from $s$ to $s'$ via $a$, it is necessary that the future starting with $s'$ succeeds in rewarding the behaviours described in $\phi'$. If $\langle s, \phi \rangle$ is in the minimal equivalent MDP, and if there really are such execution sequences succeeding in rewarding the behaviours described in $\phi$, then $\langle s', \phi' \rangle$ must also be in the minimal MDP. That is, construction by progression can only introduce e-states which are *a priori* needed. Note that an e-state that is *a priori* needed may not *really* be needed: there may in fact be no execution sequence using the available actions that exhibits a given behaviour. For instance, consider the response formula $\Box(p \rightarrow (\bigcirc^k q \rightarrow \bigcirc^k \$))$, i.e., every time trigger $p$ is true, we will be rewarded $k$ steps later provided $q$ is true then. Obviously, whether $p$ is true at some stage affects the way future states should be rewarded. However, if the transition relation happens to have the property that $k$ steps from a state satisfying $p$, no state satisfying $q$ can be reached, then *a posteriori* $p$ is irrelevant, and there was no need to label e-states differently according to whether $p$ was true or not – observe an occurrence of this in the example in Figure 7, and how this leads FLTL to produce an extra state at the bottom left of the Figure. To detect such cases, we would have to look perhaps quite deep into feasible futures, which we cannot do while constructing the e-states on the fly. Hence the relaxed notion which we call *blind minimality* does not always coincide with absolute minimality.

We now formalise the difference between true and blind minimality. For this purpose, it is convenient to define some functions $\rho$ and $\mu$ mapping e-states $e$ to functions from $S^*$ to $\mathbb{R}$ intuitively assigning rewards to sequences in the NMRDP starting from $\tau(e)$. Recall from Definition 1 that $\tau$ maps each e-state of the MDP to the underlying NMRDP state.

**Definition 6** *Let $D$ be an NMRDP. Let $S'$ be the set of e-states in an equivalent MDP $D'$ for $D$. Let $e$ be any reachable e-state in $S'$. Let $\Gamma'(i)$ be a sequence of e-states in $\widetilde{D'}(s'_0)$ such that $\Gamma'(i) = e$. Let $\Gamma(i)$ be the corresponding sequence in $\widetilde{D}(s_0)$ obtained under $\tau$ in the sense that, for each $j \leq i$, $\Gamma(j) = \tau(\Gamma'_j)$. Then for any $\Delta \in S^*$, we define*

$$\rho(e) : \Delta \mapsto \begin{cases} R(\Gamma(i-1); \Delta) & \textit{if } \Delta_0 = \Gamma_i \\ 0 & \textit{otherwise} \end{cases}$$

*and*

$$\mu(e) : \Delta \mapsto \begin{cases} R(\Gamma(i-1); \Delta) & \textit{if } \Delta \in \widetilde{D}(\Gamma_i) \\ 0 & \textit{otherwise} \end{cases}$$

*For any unreachable e-state $e$, we define both $\rho(e)(\Delta)$ and $\mu(e)(\Delta)$ to be 0 for all $\Delta$.*

Note carefully the difference between $\rho$ and $\mu$. The former describes the rewards assigned to *all* continuations of a given state sequence, while the latter confines rewards to *feasible* continuations. Note also that $\rho$ and $\mu$ are well-defined despite the indeterminacy in the choice of $\Gamma'(i)$, since by clause 4 of Definition 1, all such choices lead to the same values for $R$.

**Theorem 3** *Let $S'$ be the set of e-states in an equivalent MDP $D'$ for $D = \langle S, s_0, A, \Pr, R\rangle$. $D'$ is minimal iff every e-state in $S'$ is reachable and $S'$ contains no two distinct e-states $s'_1$ and $s'_2$ with $\tau(s'_1) = \tau(s'_2)$ and $\mu(s'_1) = \mu(s'_2)$.*

**Proof:** See Appendix B. □

Blind minimality is similar, except that, since there is no looking ahead, no distinction can be drawn between feasible trajectories and others in the future of $s$:

**Definition 7** *Let $S'$ be the set of e-states in an equivalent MDP $D'$ for $D = \langle S, s_0, A, \Pr, R\rangle$. $D'$ is* blind minimal *iff every e-state in $S'$ is reachable and $S'$ contains no two distinct e-states $s'_1$ and $s'_2$ with $\tau(s'_1) = \tau(s'_2)$ and $\rho(s'_1) = \rho(s'_2)$.*

**Theorem 4** *Let $D'$ be the translation of $D$ as in Definition 5. $D'$ is a blind minimal equivalent MDP for $D$.*

**Proof:** See Appendix B. □

The size difference between the blind-minimal and minimal MDPs will depend on the precise interaction between rewards and dynamics for the problem at hand, making theoretical analyses difficult and experimental results rather anecdotal. However, our experiments in Section 5 and 6 will show that from a computation time point of view, it is often preferable to work with the blind-minimal MDP than to invest in the overhead of computing the truly minimal one.

Finally, recall that syntactically different but semantically equivalent reward function specifications define the same e-state. Therefore, neither minimality nor blind minimality can be achieved in general without an equivalence check at least as complex as theorem proving for LTL. In pratical implementations, we avoid theorem proving in favour of embedding (fast) formula simplification in our progression and regression algorithms. This means that in principle we only approximate minimality and blind minimality, but this appears to be enough for practical purposes.

### 3.9 Embedded Solution/Construction

Blind minimality is essentially the best achievable with anytime state-based solution methods which typically extend their envelope one step forward without looking deeper into the future. Our translation into a blind-minimal MDP can be trivially embedded in any of these solution methods. This results in an 'on-line construction' of the MDP: the method entirely drives the construction of those parts of the MDP which it feels the need to explore, and leave the others implicit. If time is short, a suboptimal or even incomplete policy may be returned, but only a fraction of the state and expanded state spaces might be constructed. Note that the solution method should raise an exception as soon as one of the reward formulae progresses to $\bot$, i.e., as soon as an expanded state $\langle s, \phi \rangle$ is built such that $(\bot : r) \in \phi$, since this acts as a detector of unsuitable reward function specifications.

To the extent enabled by blind minimality, our approach allows for a dynamic analysis of the reward formulae, much as in PLTLSTR (Bacchus et al., 1997). Indeed, only the execution sequences feasible under a particular policy actually explored by the solution method contribute to the analysis of rewards for that policy. Specifically, the reward formulae generated by progression for a given policy are determined by the prefixes of the execution sequences feasible under this policy. This dynamic analysis is particularly useful, since relevance of reward formulae to particular policies (e.g. the optimal policy) cannot be detected a priori.

The forward-chaining planner TLPlan (Bacchus & Kabanza, 2000) introduced the idea of using FLTL to specify domain-specific *search control knowledge* and formula progression to prune unpromising sequential plans (plans violating this knowledge) from deterministic search spaces. This has been shown to provide enormous time gains, leading TLPlan to win the 2002 planning competition hand-tailored track. Because our approach is based on progression, it provides an elegant way to exploit search control knowledge, yet in the context of decision-theoretic planning. Here this results in a dramatic reduction of the fraction of the MDP to be constructed and explored, and therefore in substantially better policies by the deadline.

We achieve this as follows. We specify, via a \$-free formula $c_0$, properties which we know must be verified by paths feasible under *promising* policies. Then we simply progress $c_0$ alongside the reward function specification, making e-states triples $\langle s, \phi, c\rangle$ where $c$ is a \$-free formula obtained by progression. To prevent the solution method from applying an action that leads to the control knowledge being violated, the action applicability condition (item 3 in Definition 5) becomes: $a \in A'(\langle s, \phi, c\rangle)$ iff $a \in A(s)$ and $c \neq \bot$ (the other changes are straightforward). For instance, the effect of the control knowledge formula $\Box(p \rightarrow \bigcirc q)$ is to remove from consideration any feasible path in which $p$ is not followed by $q$. This is detected as soon as violation occurs, when the formula progresses to $\bot$. Although this paper focuses on non-Markovian rewards rather than dynamics, it should be noted that \$-free formulae can also be used to express non-Markovian constraints on the system's dynamics, which can be incorporated in our approach exactly as we do for the control knowledge.

### 3.10 Discussion

Existing approaches (Bacchus et al., 1996, 1997) advocate the use of PLTL over a finite past to specify non-Markovian rewards. In the PLTL style of specification, we describe the past conditions under which we get rewarded now, while with \$FLTL we describe the conditions on the present and future under which future states will be rewarded. While the behaviours and rewards may be the same in each scheme, the naturalness of thinking in one style or the other depends on the case. Letting the kids have a strawberry dessert because they have been good all day fits naturally into a past-oriented account of rewards, whereas promising that they may watch a movie if they tidy their room (indeed, making sense of the whole notion of promising) goes more naturally with \$FLTL. One advantage of the PLTL formulation is that it trivially enforces the principle that present rewards do not depend on future states. In \$FLTL, this responsibility is placed on the domain modeller. The best we can offer is an exception mechanism to recognise mistakes when their effects appear, or syntactic restrictions. On the other hand, the greater expressive power of \$FLTL opens the possibility of considering a richer class of decision processes, e.g. with uncertainty as to which rewards are received (the dessert or the movie) and when (some time next week, before it rains).

At any rate, we believe that \$FLTL is better suited than PLTL to solving NMRDPs using anytime state-based solution methods. While the pltlsim translation could be easily embedded in such a solution method, it loses the structure of the original formulae when considering subformulae individually. Consequently, the expanded state space easily becomes exponentially bigger than the blind-minimal one. This is problematic with the solution methods we consider, because size severely affects their performance in solution quality. The pre-processing phase of pltlmin uses PLTL formula regression to find sets of subformulae as potential labels for possible predecessor states, so that the subsequent generation phase builds an MDP representing all and only the histories which make a difference to the way actually feasible execution sequences should be rewarded. Not only does this recover the structure of the original formula, but in the best case, the MDP produced is exponentially smaller than the blind-minimal one. However, the prohibitive cost of the pre-processing phase makes it unsuitable for anytime solution methods. We do not consider that any method based on PLTL and regression will achieve a meaningful relaxed notion of minimality without a costly pre-processing phase. FLTL is an approach based on $FLTL and progression which does precisely that, letting the solution method resolve the tradeoff between quality and cost in a principled way intermediate between the two extreme suggestions above.

The structured representation and solution methods targeted by Bacchus et al. (1997) differ from the anytime state-based solution methods FLTL primarily aims at, in particular in that they do not require explicit state enumeration at all. Here, non-minimality is not as problematic as with the state-based approaches. In virtue of the size of the MDP produced, the PLTLSTR translation is, as PLTLSIM, clearly unsuitable to anytime state-based methods.[9] In another sense, too, FLTL represents a middle way, combining the advantages conferred by state-based and structured approaches, e.g. by PLTLMIN on one side, and PLTLSTR on the other. From the former FLTL inherits a meaningful notion of minimality. As with the latter, approximate solution methods can be used and can perform a restricted dynamic analysis of the reward formulae. In particular, formula progression enables even state-based methods to exploit some of the structure in '$FLTL space'. However, the gap between blind and true minimality indicates that progression alone is insufficient to always fully exploit that structure. There is a hope that PLTLSTR is able to take advantage of the full structure of the reward function, but also a possibility that it will fail to exploit even as much structure as FLTL, as efficiently. An empirical comparison of the three approaches is needed to answer this question and identify the domain features favoring one over the other.

## 4. NMRDPP

The first step towards a decent comparison of the different approaches is to have a framework that includes them all. The Non-Markovian Reward Decision Process Planner, NMRDPP, is a platform for the development and experimentation of approaches to NMRDPs. it provides an implementation of the approaches we have described in a common framework, within a single system, and with a common input language. NMRDPP is available on-line, see `http://rsise.anu.edu.au/~charlesg/nmrdpp`. It is worth noting that Bacchus et al. (1996, 1997) do not report any implementation of their approaches.

### 4.1 Input language

The input language enables the specification of actions, initial states, rewards, and search control-knowledge. The format for the action specification is essentially the same as in the SPUDD system (Hoey et al., 1999). The reward specification is one or more formulae, each associated with a name and a real number. These formulae are in either PLTL or $FLTL. Control knowledge is given in the same language as that chosen for the reward. Control knowledge formulae will have to be verified by any sequence of states feasible under the generated policies. Initial states are simply specified as part of the control knowledge or as explicit assignments to propositions.

9. It would be interesting, on the other hand, to use PLTLSTR in conjunction with symbolic versions of such methods, e.g. Symbolic LAO* (Feng & Hansen, 2002) or Symbolic RTDP (Feng, Hansen, & Zilberstein, 2003).

```
action flip
   heads (0.5)
endaction

action tilt
   heads (heads (0.9) (0.1))
endaction

heads = ff
[first, 5.0]? heads and ~prv (pdi heads)
[seq, 1.0]? (prv^2 heads) and (prv heads) and ~heads
```

Figure 8: Input for the Coin Example. `prv` (previously) stands for $\ominus$ and `pdi` (past diamond) stands for $\diamondminus$.

For instance, consider a simple example consisting of a coin showing either heads or tails ($\neg$heads). There are two actions that can be performed. The flip action changes the coin to show heads or tails with a 50% probability. The tilt action changes it with 10% probability, otherwise leaving it as it is. The initial state is tails. We get a reward of 5.0 for the very first head (this is written heads $\wedge \neg \ominus \diamondminus$heads in PLTL) and a reward of 1.0 each time we achieve the sequence heads, heads, tails ($\ominus^2$heads $\wedge \ominus$heads $\wedge \neg$heads in PLTL). In our input language, this NMRDP is described as shown in Figure 8.

### 4.2 Common framework

The common framework underlying nmrdpp takes advantage of the fact that NMRDP solution methods can, in general, be divided into the distinct phases of preprocessing, expansion, and solving. The first two are optional.

For pltlsim, *preprocessing* simply computes the set $\overline{\text{Sub}(F)}$ of subformulae of the reward formulae. For pltlmin, it also includes computing the labels $l(s)$ for each state $s$. For pltlstr, preprocessing involves computing the set $T$ of temporal variables as well as the ADDs for their dynamics and for the rewards. fltl does not require any preprocessing.

*Expansion* is the optional generation of the entire equivalent MDP prior to solving. Whether or not off-line expansion is sensible depends on the MDP solution method used. If state-based value or policy iteration is used, then the MDP needs to be expanded anyway. If, on the other hand, an anytime search algorithm or structured method is used, it is definitely a bad idea. In our experiments, we often used expansion solely for the purpose of measuring the size of the generated MDP.

*Solving* the MDP can be done using a number of methods. Currently, nmrdpp provides implementations of classical dynamic programming methods, namely state-based value and policy iteration (Howard, 1960), of heuristic search methods: state-based LAO* (Hansen & Zilberstein, 2001) using either value or policy iteration as a subroutine, and of one structured method, namely SPUDD (Hoey et al., 1999). Prime candidates for future developments are (L)RTDP (Bonet & Geffner, 2003), symbolic LAO* (Feng & Hansen, 2002), and symbolic RTDP (Feng et al., 2003).

| | |
|---|---|
| `> loadWorld('coin')` | load coin NMRDP |
| `> preprocess('sPltl')` | pltlstr preprocessing |
| `> startCPUtimer` | |
| `> spudd(0.99, 0.0001)` | solve MDP with SPUDD($\beta, \epsilon$) |
| `> stopCPUtimer` | |
| `> readCPUtimer` | report solving time |
| `1.22000` | |
| `> iterationCount` | report number of iterations |
| `1277` | |
| `> displayDot(valueToDot)` | display ADD of value function |

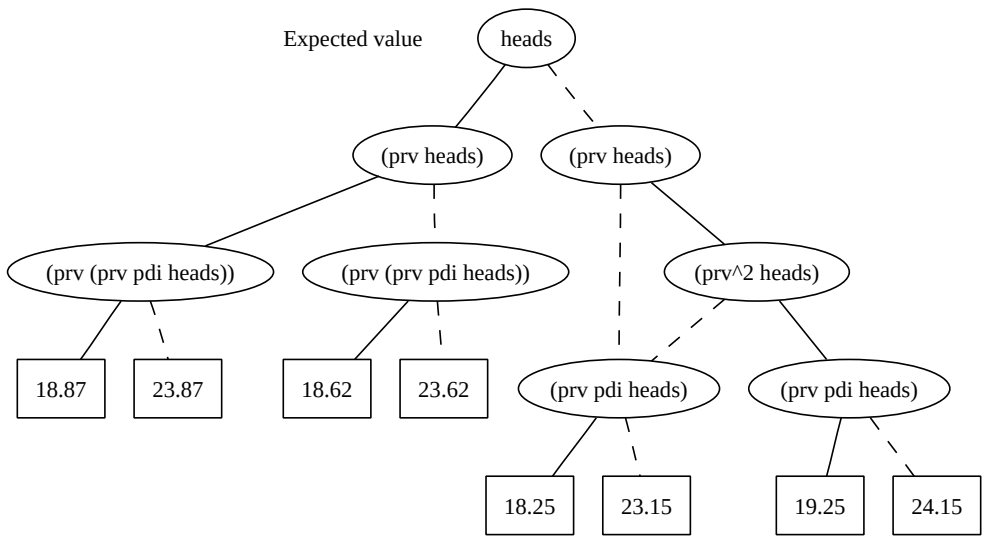

| | |
|---|---|
| `> displayDot(policyToDot)` | display policy |

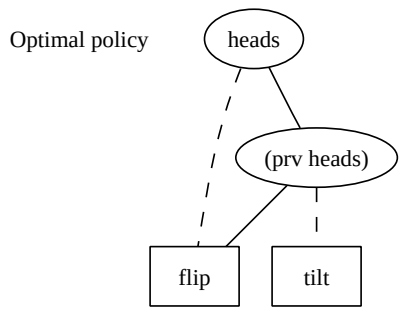

| | |
|---|---|
| `> preprocess('mPltl')` | pltlmin preprocessing |
| `> expand` | completely expand MDP |
| `> domainStateSize` | report MDP size |
| `6` | |
| `> printDomain ("") \| 'show-domain.rb'` | display postscript rendering of MDP |

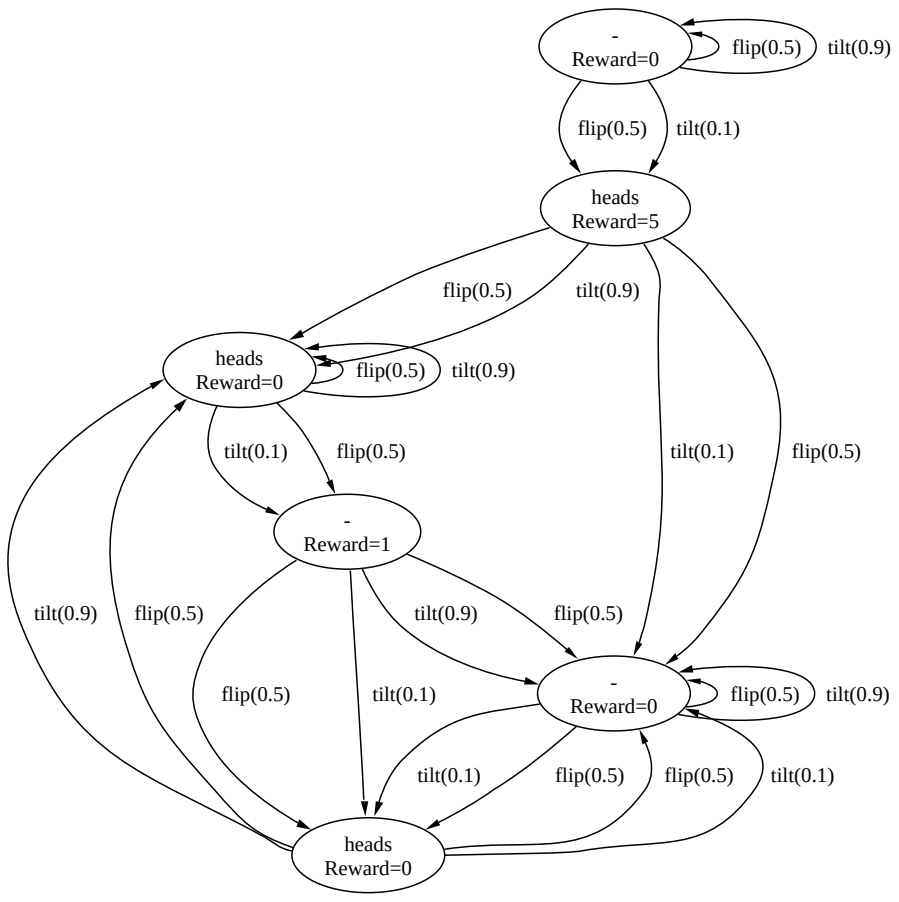

| | |
|---|---|
| `> valIt(0.99, 0.0001)` | solve MDP with VI($\beta, \epsilon$) |
| `> iterationCount` | report number of iterations |
| `1277` | |
| `> getPolicy` | output policy (textual) |
| `...` | |

Figure 9: Sample Session

### 4.3 Approaches covered

Altogether, the various types of preprocessing, the choice of whether to expand, and the MDP solution methods, give rise to quite a number of NMRDP approaches, including, but not limited to those previously mentioned (see e.g. pltlstr(a) below). Not *all* combinations are possible. E.g., state-based processing variants are incompatible with structured solution methods (the converse is possible in principle, however). Also, there is at present no structured form of preprocessing for $FLTL formulae.

pltlstr(a) is an example of an interesting variant of pltlstr, which we obtain by considering additional preprocessing, whereby the state space is explored (without explicitly enumerating it) to produce a BDD representation of the e-states reachable from the start state. This is done by starting with a BDD representing the start e-state, and repeatedly applying each action. Non-zero probabilities are converted to ones and the result "or-ed" with the last result. When no action adds any reachable e-states to this BDD, we can be sure it represents the reachable e-state space. This is then used as additional control knowledge to restrict the search. It should be noted that without this phase pltlstr makes no assumptions about the start state, and thus is left at a possible disadvantage. Similar structured reachability analysis techniques have been used in the symbolic implementation of LAO* (Feng & Hansen, 2002). However, an important aspect of what we do here is that temporal variables are also included in the BDD.

### 4.4 The nmrdpp System

nmrdpp is controlled by a command language, which is read either from a file or interactively. The command language provides commands for the different phases (preprocessing, expansion, solution) of the methods, commands to inspect the resulting policy and value functions, e.g. with rendering via DOT (AT&T Labs-Research, 2000), as well as supporting commands for timing and memory usage. A sample session, where the coin NMRDP is successively solved with pltlstr and pltlmin is shown in Figure 9.

nmrdpp is implemented in C++, and makes use of a number of supporting libraries. In particular, it relies heavily on the CUDD package for manipulating ADDs (Somenzi, 2001): action specification trees are converted into and stored as ADDs by the system, and moreover the structured algorithms rely heavily on CUDD for ADD computations. The state-based algorithms make use of the MTL – Matrix Template Library for matrix operations. MTL takes advantage of modern processor features such as MMX and SSE and provides efficient sparse matrix operations. We believe that our implementations of MDP solution methods are comparable with the state of the art. For instance, we found that our implementation of SPUDD is comparable in performance (within a factor of 2) to the reference implementation (Hoey et al., 1999). On the other hand, we believe that data structures used for regression and progression of temporal formulae could be optimised.

## 5. Experimental Analysis

We are faced with three substantially different approaches that are not easy to compare, as their performance will depend on domain features as varied as the structure in the transition model, the type, syntax, and length of the temporal reward formula, the presence of rewards unreachable or irrelevant to the optimal policy, the availability of good heuristics and control-knowledge, etc, and on the interactions between these factors. In this section, we report an experimental investigation into the influence of some of these factors and try to answer the questions raised previously:[10]

1. is the dynamics of the domain the predominant factor affecting performance?
2. is the type of reward a major factor?
3. is the syntax used to describe rewards a major factor?
4. is there an overall best method?
5. is there an overall worst method?
6. does the preprocessing phase of PLTLMIN pay, compared to PLTLSIM?
7. does the simplicity of the FLTL translation compensate for blind-minimality, or does the benefit of true minimality outweigh the cost of PLTLMIN preprocessing?
8. are the dynamic analyses of rewards in PLTLSTR and FLTL effective?
9. is one of these analyses more powerful, or are they rather complementary?

In some cases but not all, we were able to identify systematic patterns. The results in this section were obtained using a Pentium4 2.6GHz GNU/Linux 2.4.20 machine with 500MB of ram.

### 5.1 Preliminary Remarks

Clearly, FLTL and PLTLSTR(A) have great potential for exploiting domain-specific heuristics and control-knowledge; PLTLMIN less so. To avoid obscuring the results, we therefore refrained from incorporating these features in the experiments. When running LAO*, the heuristic value of a state was the crudest possible (the sum of all reward values in the problem). Performance results should be interpreted in this light – they do not necessarily reflect the practical abilities of the methods that are able to exploit these features.

We begin with some general observations. One question raised above was whether the gain during the PLTL expansion phase is worth the expensive preprocessing performed by PLTLMIN, i.e. whether PLTLMIN typically outperforms PLTLSIM. We can definitively answer this question: up to pathological exceptions, preprocessing pays. We found that expansion was the bottleneck, and that post-hoc minimisation of the MDP produced by PLTLSIM did not help much. PLTLSIM is therefore of little or no practical interest, and we decided not to report results on its performance, as it is often an order of magnitude worse than that of PLTLMIN. Unsurprisingly, we also found that PLTLSTR would typically scale to larger state spaces, inevitably leading it to outperform state-based methods. However, this effect is not uniform: structured solution methods sometimes impose excessive memory requirements which makes them uncompetitive in certain cases, for example where $\ominus^n f$, for large $n$, features as a reward formula.

10. Here is an executive summary of the answers for the executive reader. 1. no, 2. yes, 3. yes, 4. PLTLSTR and FLTL, 5. PLTLSIM, 6. yes, 7. yes and no, respectively, 8. yes, 9. no and yes, respectively.

### 5.2 Domains

Experiments were performed on four hand-coded domains (propositions + dynamics) and on random domains. Each hand-coded domain has $n$ propositions $p_i$, and a dynamics which makes every state possible and eventually reachable from the initial state in which all propositions are false. The first two such domains, SPUDD-LINEAR and SPUDD-EXPON were discussed by Hoey et al. (1999); the two others are our own.

The intention of SPUDD-LINEAR was to take advantage of the best case behaviour of SPUDD. For each proposition $p_i$, it has an action $a_i$ which sets $p_i$ to true and all propositions $p_j$, $1 \leq j < i$ to false. SPUDD-EXPON, was used by Hoey et al. (1999) to demonstrate the worst case behaviour of SPUDD. For each proposition $p_i$, it has an action $a_i$ which sets $p_i$ to true only when all propositions $p_j$, $1 \leq j < i$ are true (and sets $p_i$ to false otherwise), and sets the latter propositions to false. The third domain, called ON/OFF, has one "turn-on" and one "turn-off" action per proposition. The "turn-on-$p_i$" action only probabilistically succeeds in setting $p_i$ to true when $p_i$ was false. The turn-off action is similar. The fourth domain, called COMPLETE, is a fully connected reflexive domain. For each proposition $p_i$ there is an action $a_i$ which sets $p_i$ to true with probability $i/(n+1)$ (and to false otherwise) and $p_j$, $j \neq i$ to true or false with probability 0.5. Note that $a_i$ can cause a transition to any of the $2^n$ states.

Random domains of size $n$ also involve $n$ propositions. The method for generating their dynamics is detailed in appendix C. Let us just summarise by saying that we are able to generate random dynamics exhibiting a given degree of "structure" and a given degree of uncertainty. Lack of structure essentially measures the bushiness of the internal part of the ADDs representing the actions, and uncertainty measures the bushiness of their leaves.

### 5.3 Influence of Dynamics

The interaction between dynamics and reward certainly affects the performance of the different approaches, though not so strikingly as other factors such as the reward type (see below). We found that under the same reward scheme, varying the degree of structure or uncertainty did not generally change the relative success of the different approaches. For instance, Figures 10 and 11 show the average run time of the methods as a function of the degree of structure, resp. degree of uncertainty, for random problems of size $n = 6$ and reward $\circleddash^n \neg \circleddash \top$ (the state encountered at stage $n$ is rewarded, regardless of its properties[11]). Run-time increases slightly with both degrees, but there is no significant change in relative performance. These are typical of the graphs we obtain for other rewards.

Clearly, counterexamples to this observation exist. These are most notable in cases of extreme dynamics, for instance with the SPUDD-EXPON domain. Although for small values of $n$, such as $n = 6$, PLTLSTR approaches are faster than the others in handling the reward $\circleddash^n \neg \circleddash \top$ for virtually any type of dynamics we encountered, they perform very poorly with that reward on SPUDD-EXPON. This is explained by the fact that only a small fraction of SPUDD-EXPON states are reachable in the first $n$ steps. After $n$ steps, FLTL immediately recognises that reward is of no consequence, because the formula has progressed to $\top$. PLTLMIN discovers this fact only after expensive preprocessing. PLTLSTR, on the other hand, remains concerned by the prospect of reward, just as PLTLSIM would.

11. $\bigcirc^n \$$ in \$FLTL

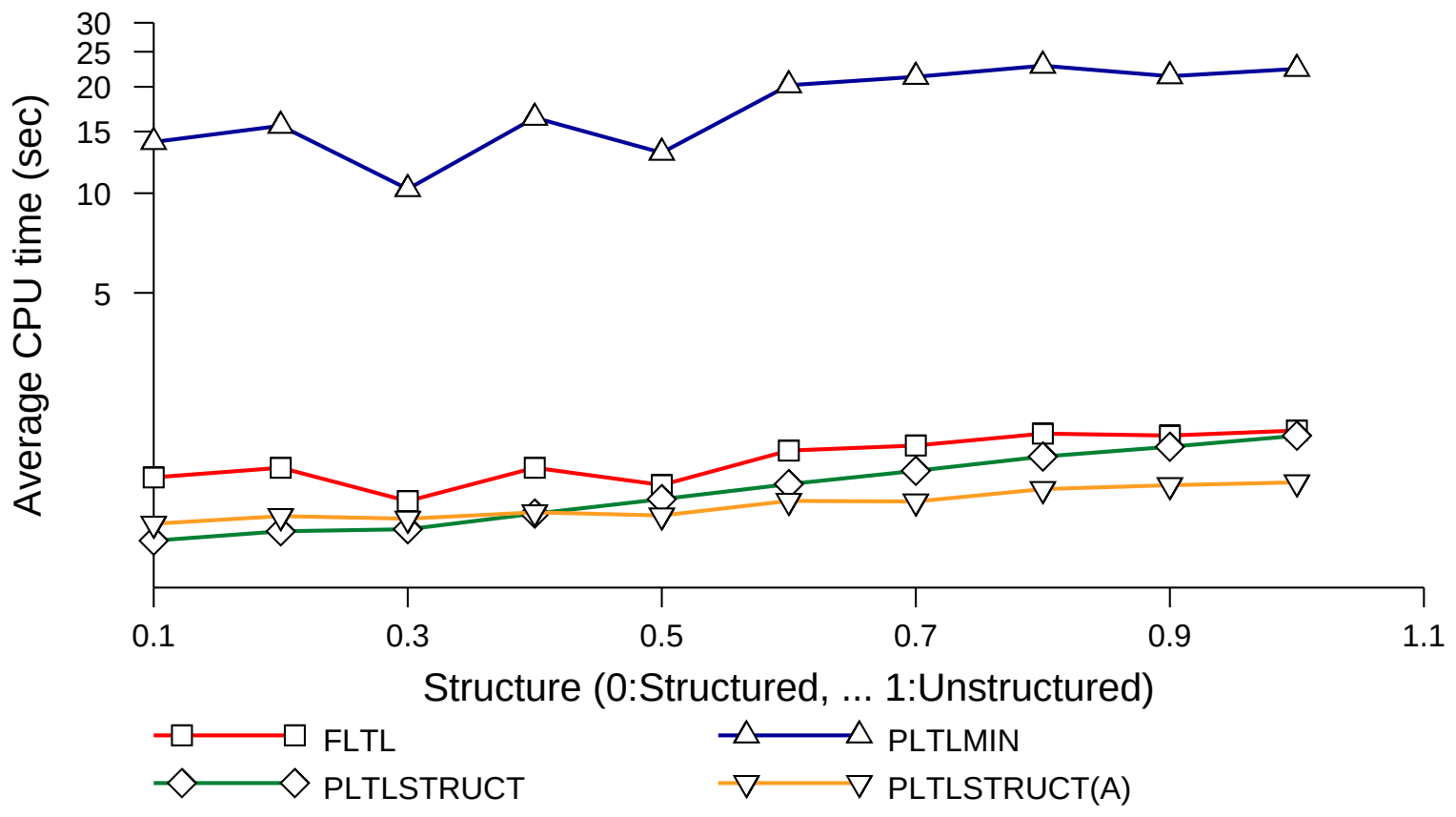


Figure 10: Changing the Degree of Structure

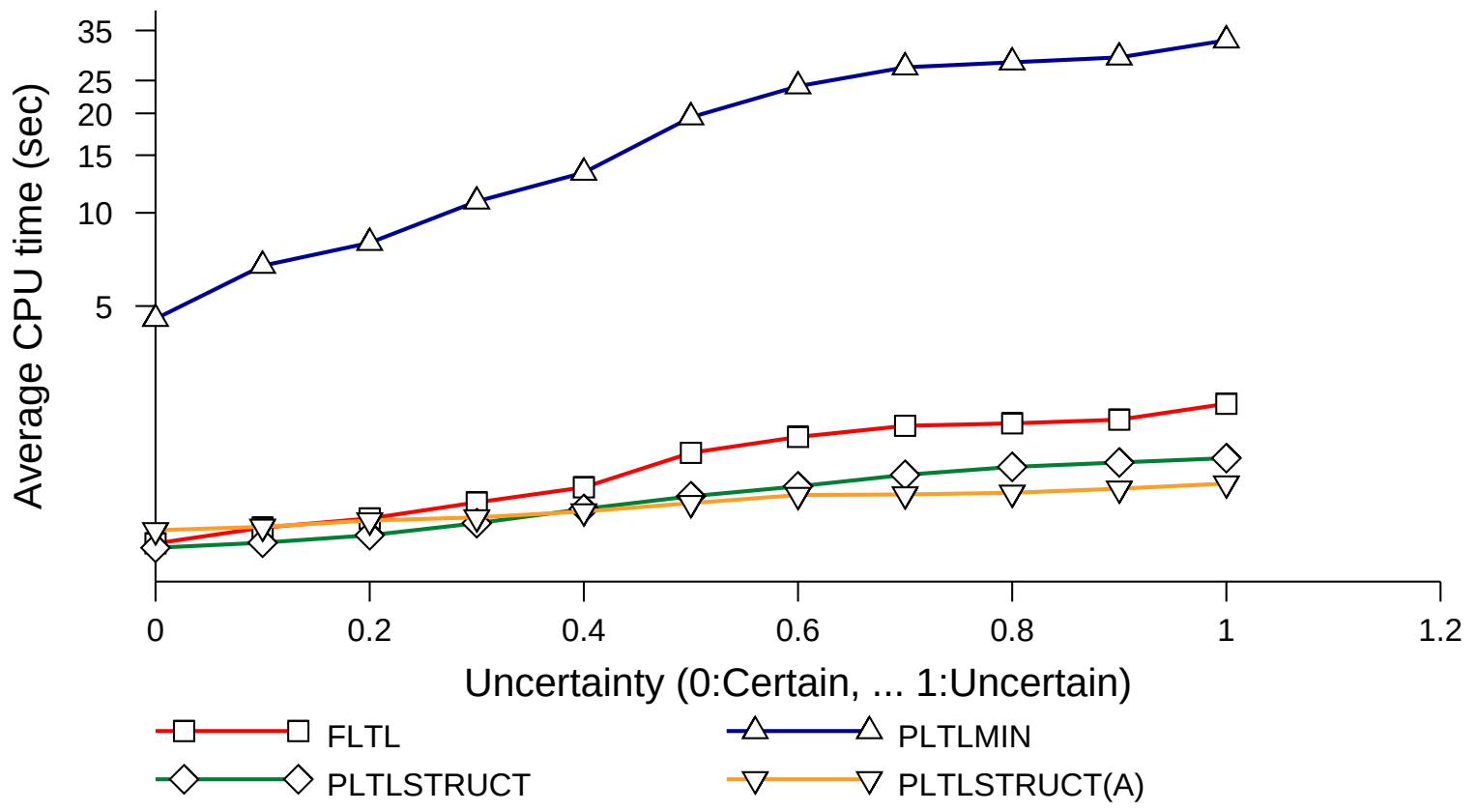


Figure 11: Changing the Degree of Uncertainty

### 5.4 Influence of Reward Types

The type of reward appears to have a stronger influence on performance than dynamics. This is unsurprising, as the reward type significantly affects the size of the generated MDP: certain rewards only make the size of the minimal equivalent MDP increase by a constant number of states or a constant factor, while others make it increase by a factor exponential in the length of the formula. Table 1 illustrates this. The third column reports the size of the minimal equivalent MDP induced by the formulae on the left hand side.[12]

A legitimate question is whether there is a direct correlation between size increase and (in)appropriateness of the different methods. For instance, we might expect the state-based methods to do particularly well in conjunction with reward types inducing a small MDP and

12. The figures are not necessarily valid for non-completely connected NMRDPs. Unfortunately, even for completely connected domains, there does not appear to be a much cheaper way to determine the MDP size than to generate it and count states.

| type | formula | size | fastest | slowest |
|---|---|---|---|---|
| first time all $p_i$s | $(\wedge_{i=1}^{n} p_i) \wedge (\neg \ominus \Diamond\!\!\!\!- \wedge_{i=1}^{n} p_i)$ | $\mathcal{O}(1)\|\|S\|\|$ | PLTLSTR(A) | PLTLMIN |
| $p_i$s in sequence from start state | $(\wedge_{i=1}^{n} \ominus^i p_i) \wedge \ominus^n \neg \ominus \top$ | $\mathcal{O}(n)\|\|S\|\|$ | FLTL | PLTLSTR |
| two consecutive $p_i$s | $\vee_{i=1}^{n-1} (\ominus p_i \wedge p_{i+1})$ | $\mathcal{O}(n^k)\|\|S\|\|$ | PLTLSTR | FLTL |
| all $p_i$s $n$ times ago | $\ominus^n \wedge_{i=1}^{n} p_i$ | $\mathcal{O}(2^n)\|\|S\|\|$ | PLTLSTR | PLTLMIN |

Table 1: Influence of Reward Type on MDP Size and Method Performance

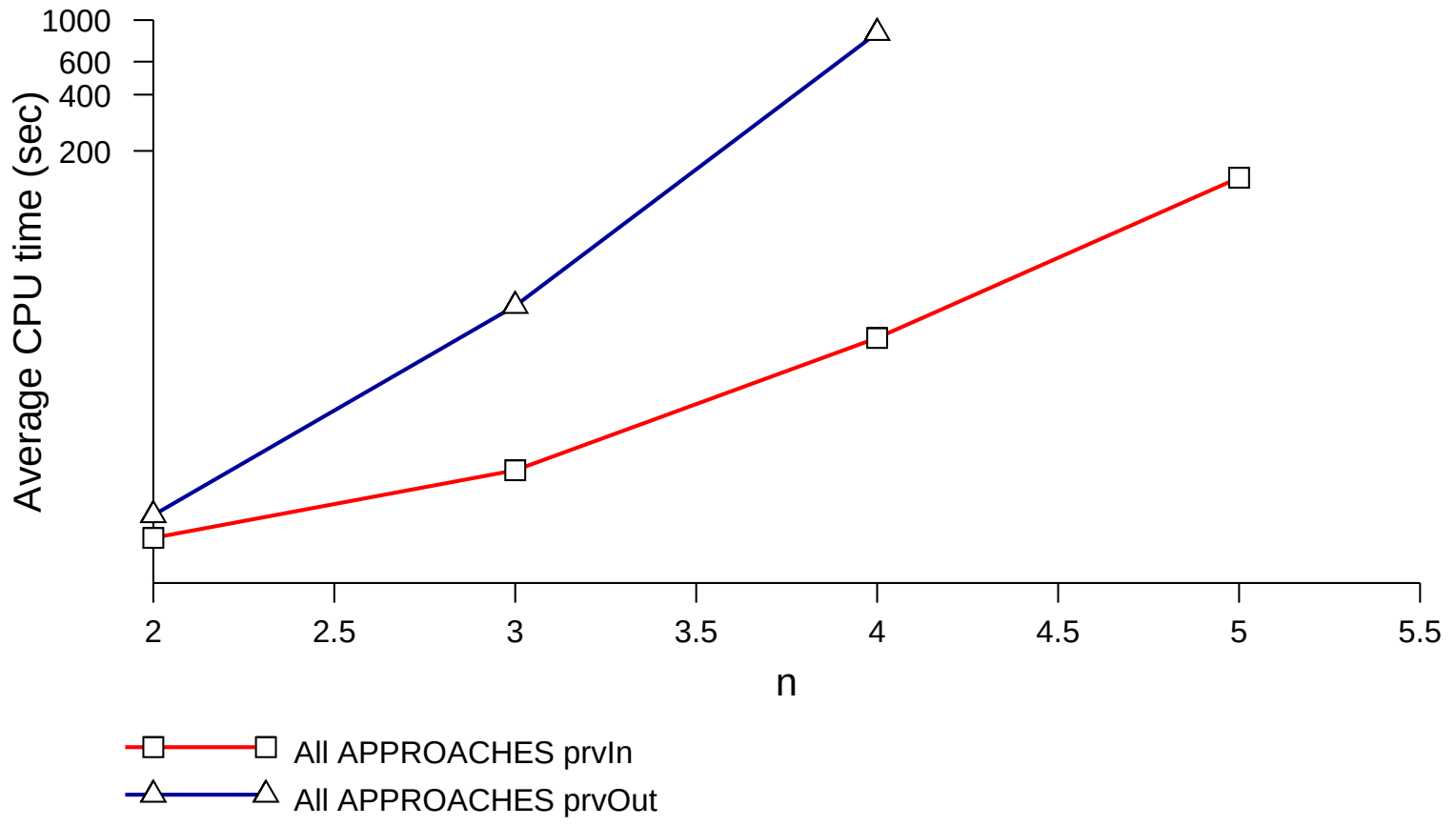


Figure 12: Changing the Syntax

otherwise badly in comparison with structured methods. Interestingly, this is not always the case. For instance, in Table 1 whose last two columns report the fastest and slowest methods over the range of hand-coded domains where $1 \leq n \leq 12$, the first row contradicts that expectation. Moreover, although PLTLSTR is fastest in the last row, for larger values of $n$ (not represented in the table), it aborts through lack of memory, unlike the other methods.

The most obvious observations arising out of these experiments is that PLTLSTR is nearly always the fastest – until it runs out of memory. Perhaps the most interesting results are those in the second row, which expose the inability of methods based on PLTL to deal with rewards specified as long sequences of events. In converting the reward formula to a set of subformulae, they lose information about the order of events, which then has to be recovered laboriously by reasoning. $FLTL progression in contrast takes the events one at a time, preserving the relevant structure at each step. Further experimentation led us to observe that all PLTL based algorithms perform poorly where reward is specified using formulae of the form $\ominus^k f$, $\Diamond\!\!\!\!-_k f$, and $\boxminus_k f$ ($f$ has been true $k$ steps ago, within the last $k$ steps, or at all of the last $k$ steps).

### 5.5 Influence of Syntax

Unsurprisingly, we find that the syntax used to express rewards, which affects the length of the formula, has a major influence on the run time. A typical example of this effect is captured in Figure 12. This graph demonstrates how re-expressing $prvOut \equiv \ominus^n(\wedge_{i=1}^{n} p_i)$

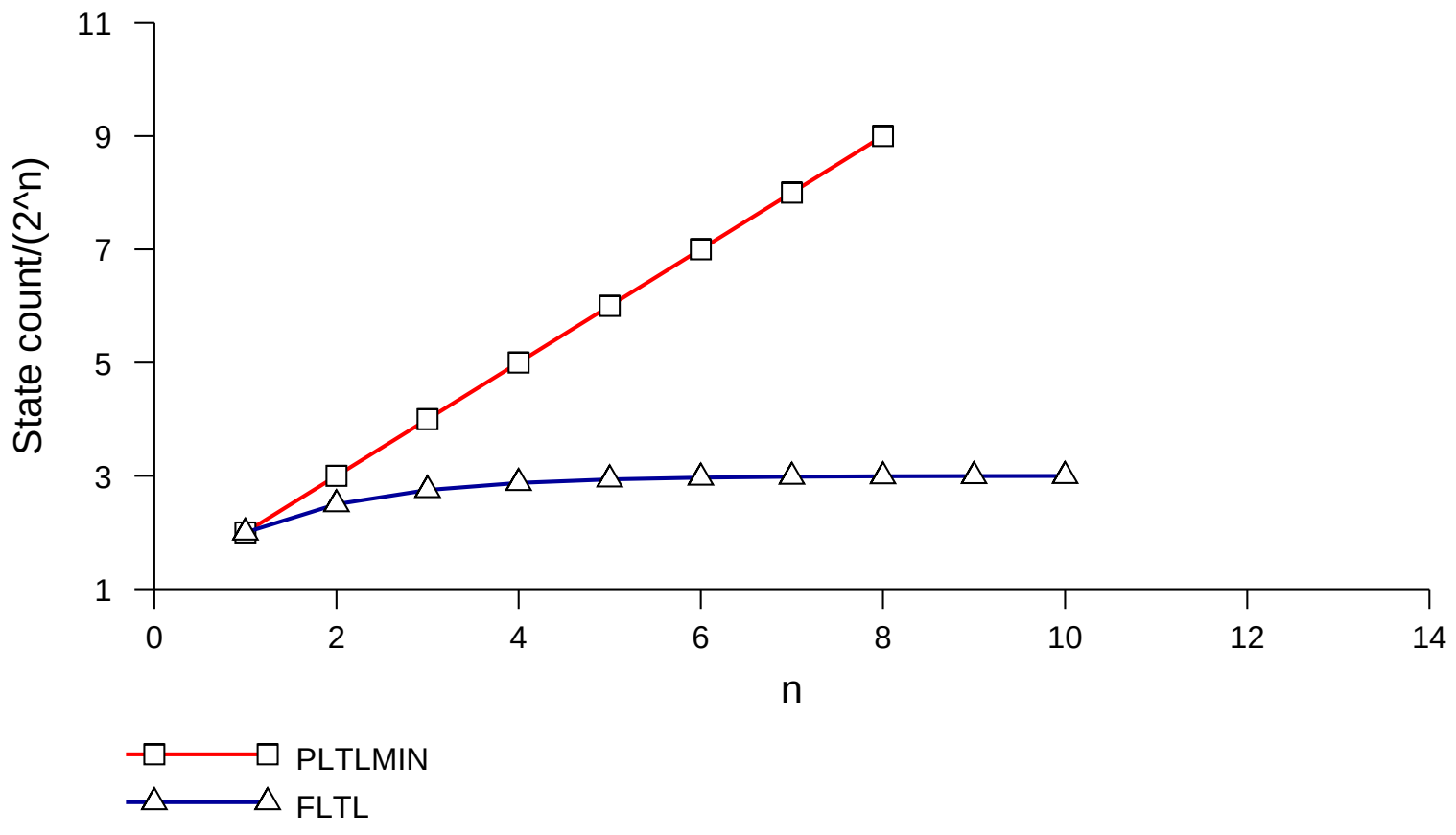


Figure 13: Effect of Multiple Rewards on MDP size

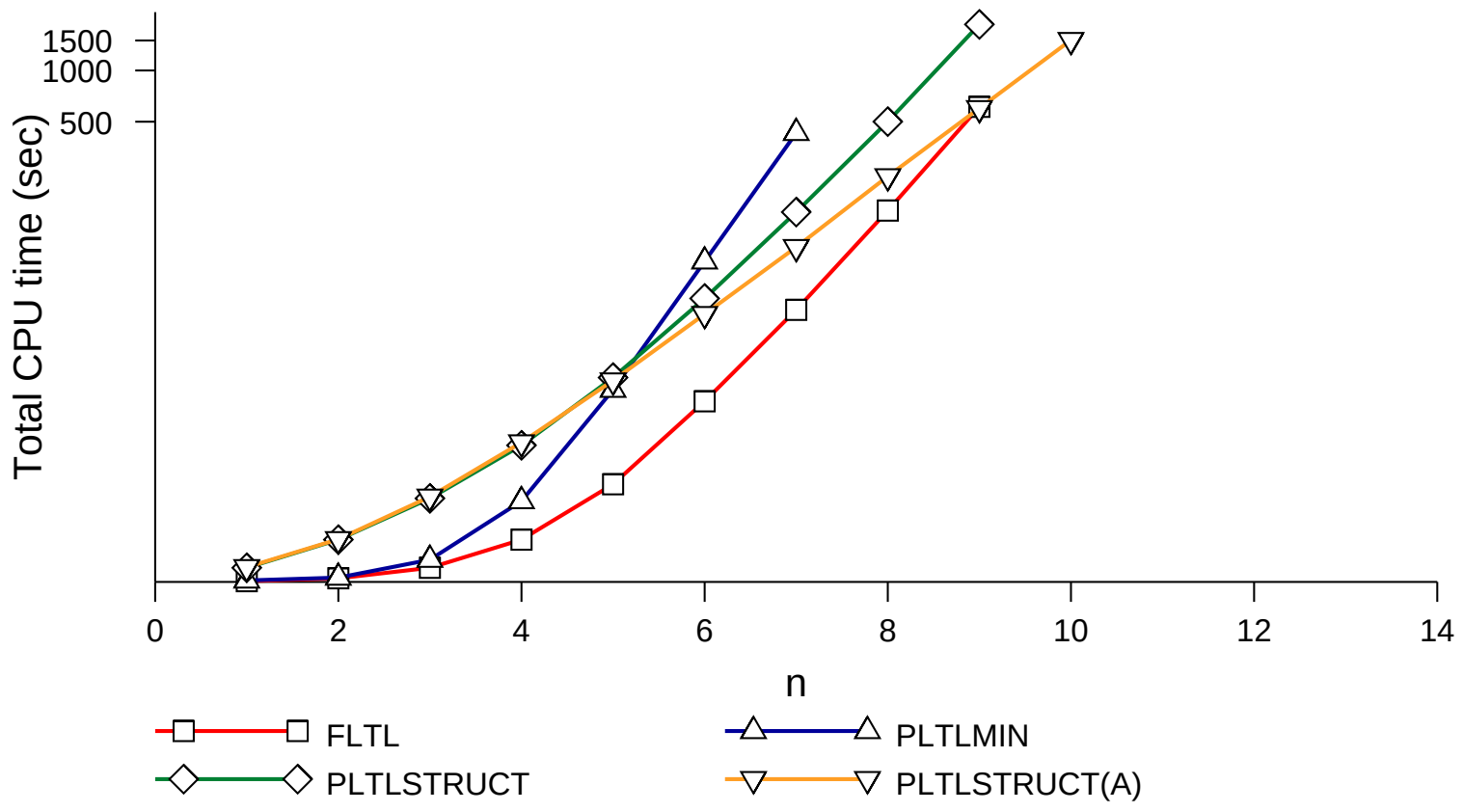


Figure 14: Effect of Multiple Rewards on Run Time

as $prvIn \equiv \wedge_{i=1}^{n} \ominus^n p_i$, thereby creating $n$ times more temporal subformulae, alters the running time of all PLTL methods. fltl is affected too as \$FLTL progression requires two iterations through the reward formula. The graph represents the averages of the running times over all the methods, for the complete domain.

Our most serious concern in relation to the PLTL approaches is their handling of reward specifications containing multiple reward elements. Most notably we found that pltlmin does not necessarily produce the minimal equivalent MDP in this situation. To demonstrate, we consider the set of reward formulae $\{f_1, f_2, \ldots, f_n\}$, each associated with the same real value $r$. Given this, PLTL approaches will distinguish unnecessarily between past behaviours which lead to identical future rewards. This may occur when the reward at an e-state is determined by the truth value of $f_1 \vee f_2$. This formula does not necessarily require e-states that distinguish between the cases in which $\{f_1 \equiv \top, f_2 \equiv \bot\}$ and $\{f_1 \equiv \bot, f_2 \equiv \top\}$ hold; however, given the above specification, pltlmin makes this distinction. For example,

taking $f_i = \ominus p_i$, Figure 13 shows that fltl leads to an MDP whose size is at most 3 times that of the NMRDP. In contrast, the relative size of the MDP produced by pltlmin is linear in $n$, the number of rewards and propositions. These results are obtained with all hand-coded domains except spudd-expon. Figure 14 shows the run-times as a function of $n$ for complete. fltl dominates and is only overtaken by pltlstr(A) for large values of $n$, when the MDP becomes too large for explicit exploration to be practical. To obtain the minimal equivalent MDP using pltlmin, a bloated reward specification of the form $\{(\ominus \vee_{i=1}^{n} (p_i \wedge_{j=1,j\neq i}^{n} \neg p_j) : r), \ldots, (\ominus \wedge_{i=1}^{n} p_i : n * r)\}$ is necessary, which, by virtue of its exponential length, is not an adequate solution.

### 5.6 Influence of Reachability

All approaches claim to have some ability to ignore variables which are irrelevant because the condition they track is unreachable:[13] pltlmin detects them through preprocessing, pltlstr exploits the ability of structured solution methods to ignore them, and fltl ignores them when progression never exposes them. However, given that the mechanisms for avoiding irrelevance are so different, we expect corresponding differences in their effects. On experimental investigation, we found that the differences in performance are best illustrated by looking at response formulae, which assert that if a trigger condition $c$ is reached then a reward will be received upon achievement of the goal $g$ in, resp. within, $k$ steps. In PLTL, this is written $g \wedge \ominus^k c$, resp. $g \wedge \diamondminus_k c$, and in \$FLTL, $\Box(c \rightarrow \bigcirc^k (g \rightarrow \$))$, resp. $\Box(c \rightarrow \Box_k (g \rightarrow \$))$

When the *goal* is unreachable, PLTL approaches perform well. As it is always false, the goal $g$ does not lead to behavioural distinctions. On the other hand, while constructing the MDP, fltl considers the successive progressions of $\bigcirc^k g$ without being able to detect that it is unreachable until it actually fails to happen. This is exactly what the blindness of blind minimality amounts to. Figure 15 illustrates the difference in performance as a function of the number $n$ of propositions involved in the spudd-linear domain, when the reward is of the form $g \wedge \ominus^n c$, with $g$ unreachable.

fltl shines when the *trigger* is unreachable. Since $c$ never happens, the formula will always progress to itself, and the goal, however complicated, is never tracked in the generated MDP. In this situation PLTL approaches still consider $\ominus^k c$ and its subformulae, only to discover, after expensive preprocessing for pltlmin, after reachability analysis for pltlstr(a), and never for pltlstr, that these are irrelevant. This is illustrated in Figure 16, again with spudd-linear and a reward of the form $g \wedge \ominus^n c$, with $c$ unreachable.

### 5.7 Dynamic Irrelevance

Earlier we claimed that one advantage of pltlstr and fltl over pltlmin and pltlsim is that the former perform a dynamic analysis of rewards capable of detecting irrelevance of variables to particular policies, e.g. to the optimal policy. Our experiments confirm this claim. However, as for reachability, whether the goal or the triggering condition in a response formula becomes irrelevant plays an important role in determining whether a

13. Here we sometimes speak of conditions and goals being ‘reachable’ or ‘achievable’ rather than ‘feasible’, although they may be temporally extended. This is to keep in line with conventional vocabulary as in the phrase ‘reachability analysis’.

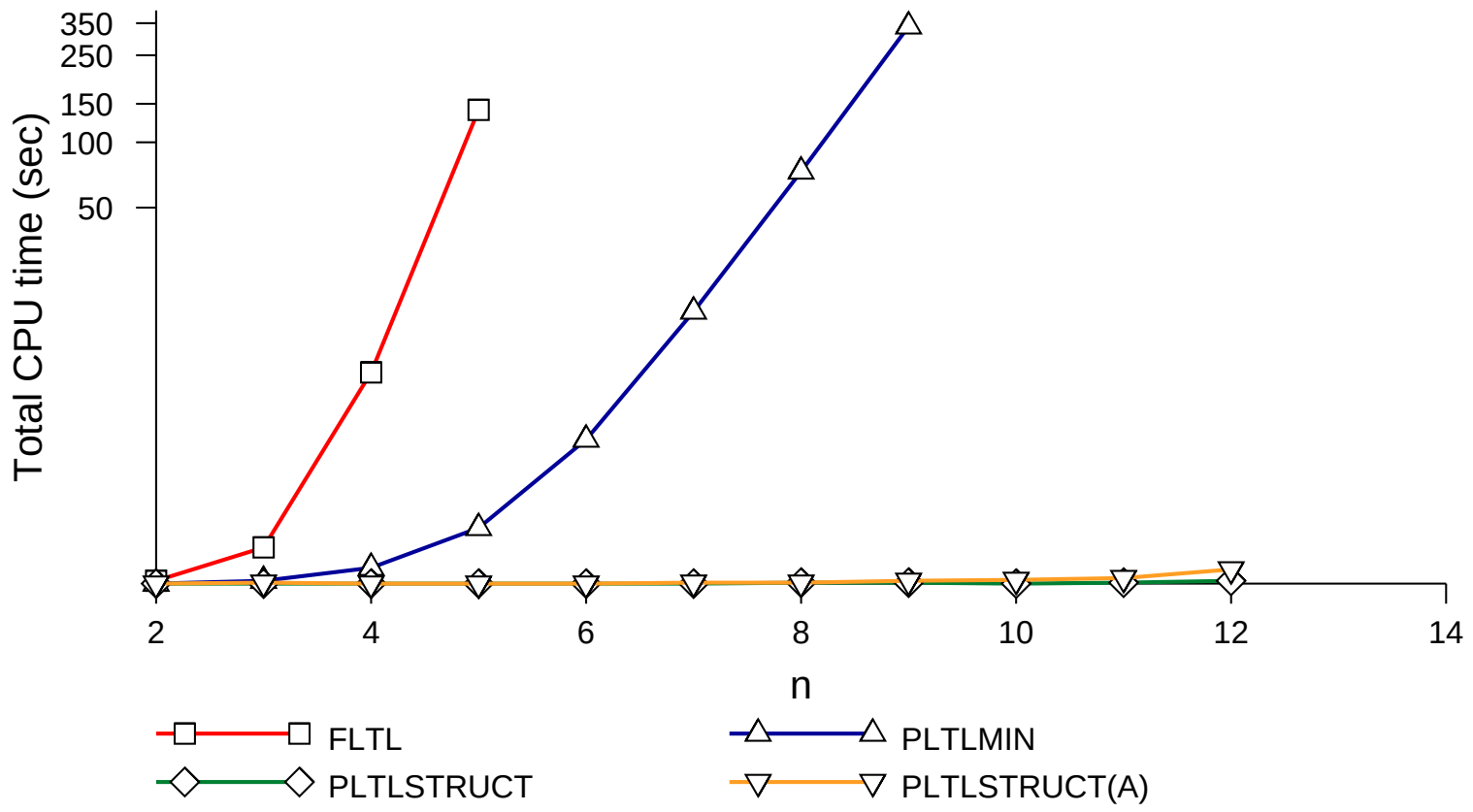


Figure 15: Response Formula with Unachievable Goal

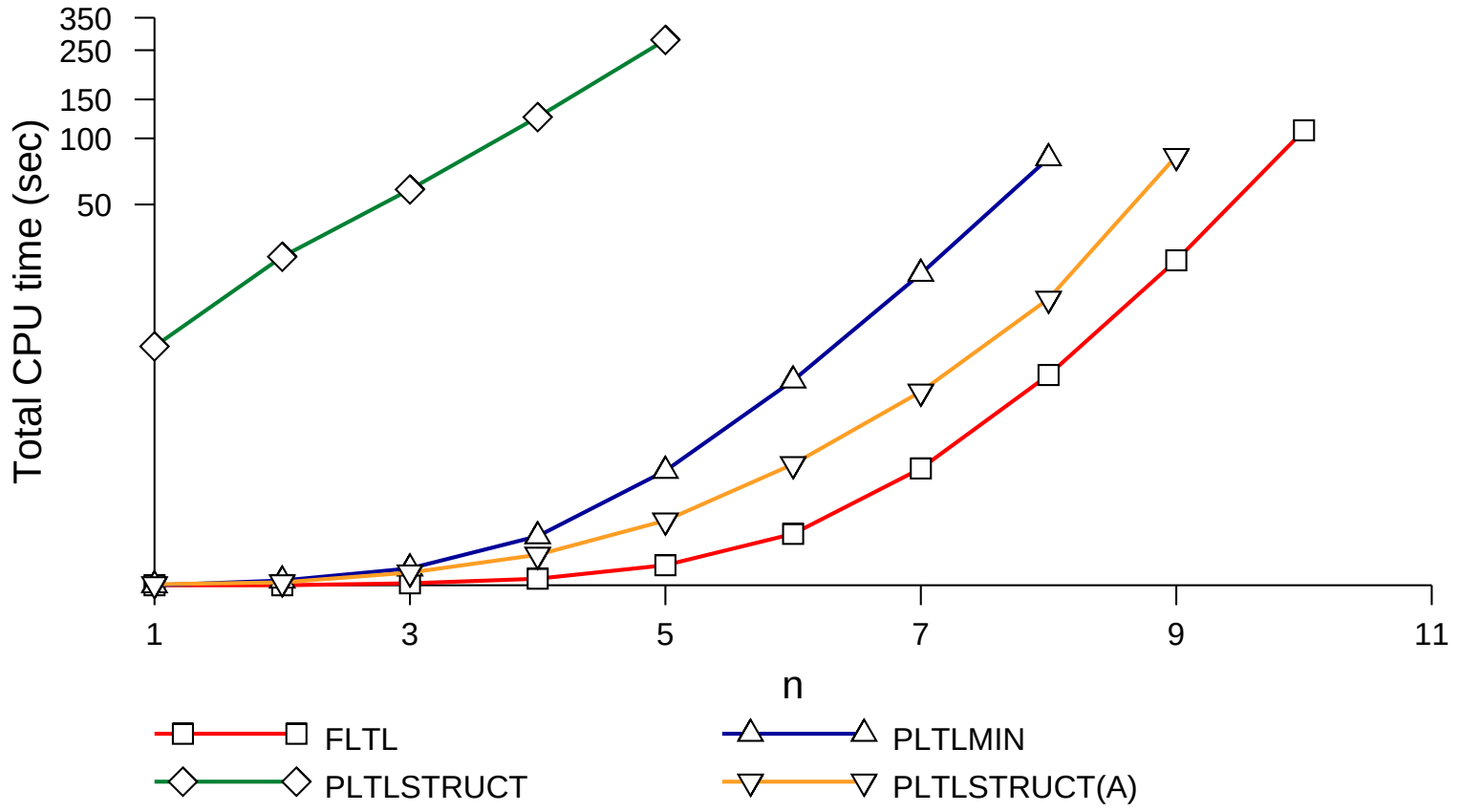


Figure 16: Response Formula with Unachievable Trigger

pltlstr or fltl approach should be taken: pltlstr is able to dynamically ignore the goal, while fltl is able to dynamically ignore the trigger.

This is illustrated in Figures 17 and 18. In both figures, the domain considered is on/off with $n = 6$ propositions, the response formula is $g \wedge \ominus^n c$ as before, here with both $g$ and $c$ achievable. This response formula is assigned a fixed reward. To study the effect of dynamic irrelevance of the goal, in Figure 17, achievement of $\neg g$ is rewarded by the value $r$ (i.e. we have $(\neg g : r)$ in PLTL). In Figure 18, on the other hand, we study the effect of dynamic irrelevance of the trigger and achievement of $\neg c$ is rewarded by the value $r$. Both figures show the runtime of the methods as $r$ increases.

Achieving the goal, resp. the trigger, is made less attractive as $r$ increases up to the point where the response formula becomes irrelevant under the optimal policy. When this happens, the run-time of pltlstr resp. fltl, exhibits an abrupt but durable improvement. The figures show that fltl is able to pick up irrelevance of the trigger, while pltlstr is able to exploit irrelevance of the goal. As expected, pltlmin whose analysis is static does not pick

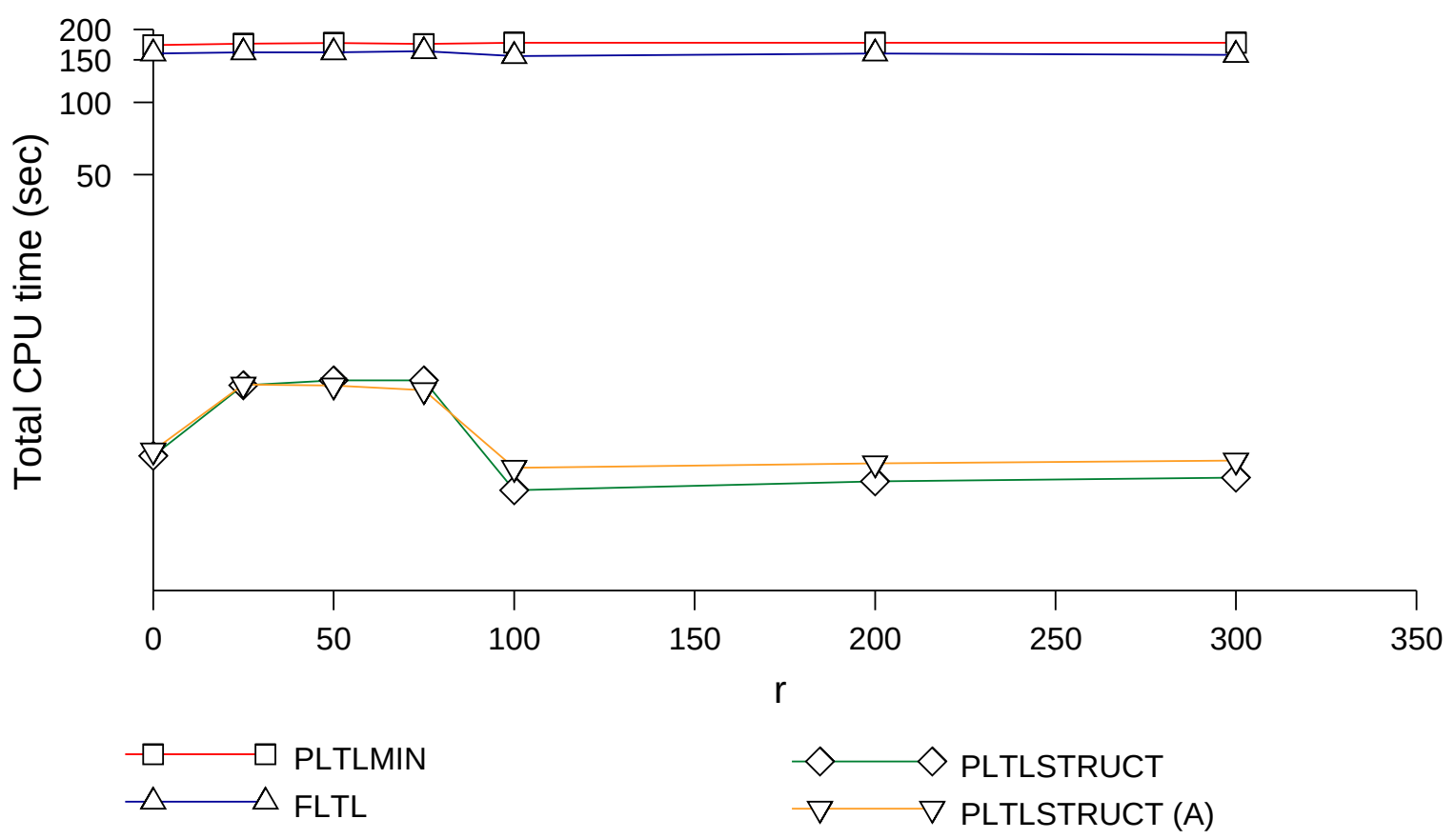


Figure 17: Response Formula with Unrewarding Goal

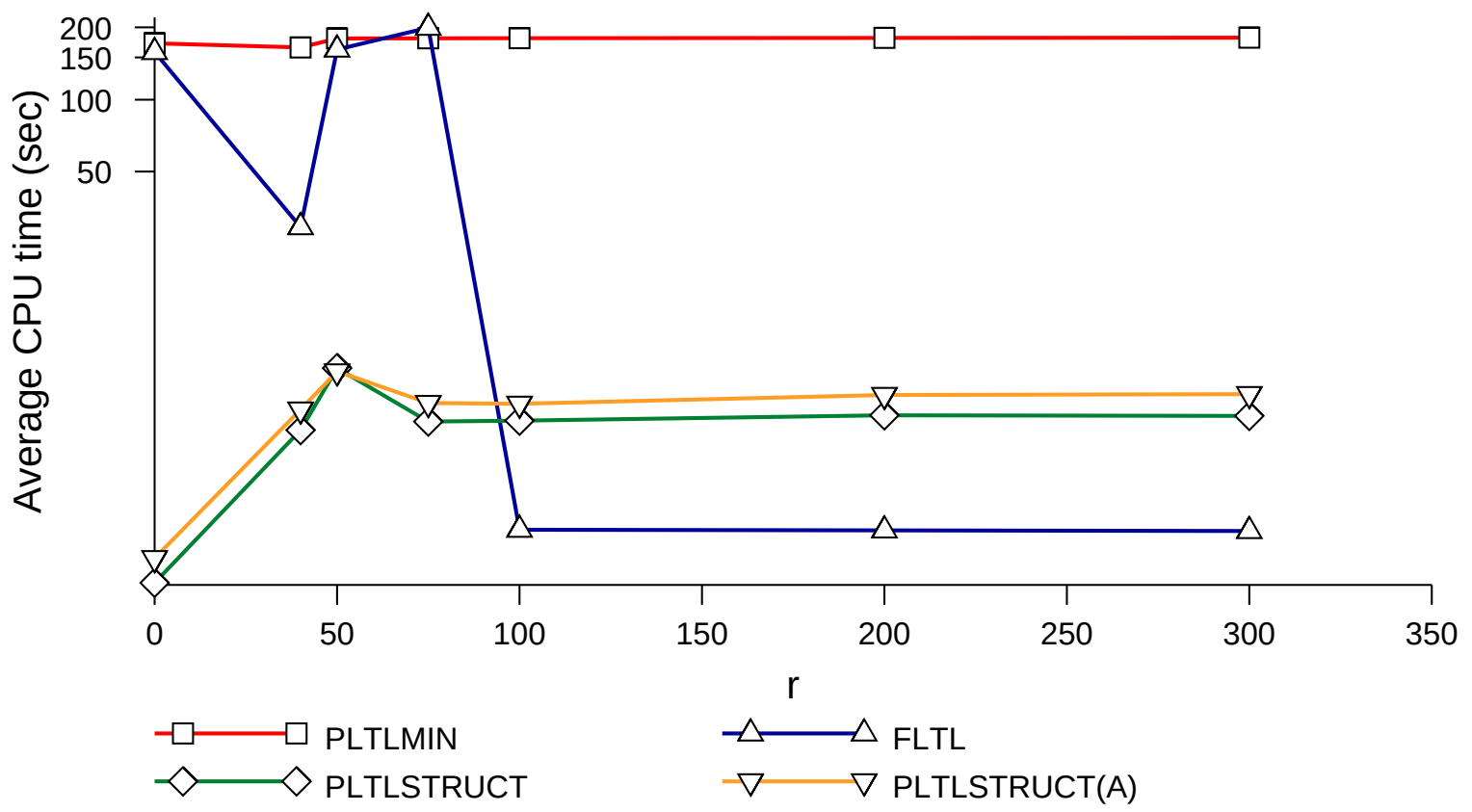


Figure 18: Response Formula with Unrewarding Trigger

up either and performs consistently badly. Note that in both figures, PLTLSTR progressively takes longer to compute as $r$ increases because value iteration requires additional iterations to converge.

## 5.8 Summary

In our experiments with artificial domains, we found PLTLSTR and FLTL preferable to state-based PLTL approaches in most cases. If one insists on using the latter, we strongly recommend preprocessing. FLTL is the technique of choice when the reward requires tracking a long sequence of events or when the desired behaviour is composed of many elements with identical rewards. For response formulae, we advise the use of PLTLSTR if the probability of reaching the goal is low or achieving the goal is very costly, and conversely, we advise the use of FLTL if the probability of reaching the triggering condition is low or if reaching it is very costly. In all cases, attention should be paid to the syntax of the reward formulae and

in particular to minimising its length. Indeed, as could be expected, we found the syntax of the formulae and the type of non-Markovian reward they encode to be a predominant factor in determining the difficulty of the problem, much more so than the features of the Markovian dynamics of the domain.

## 6. A Concrete Example

Our experiments have so far focused on artificial problems and have aimed at characterising the strengths and weaknesses of the various approaches. We now look at a concrete example in order to give a sense of the size of more interesting problems that these techniques can solve. Our example is derived from the Miconic elevator classical planning benchmark (Koehler & Schuster, 2000). An elevator must get a number of passengers from their *origin* floor to their *destination*. Initially, the elevator is *at* some arbitrary floor and no passenger is *served* nor has *boarded* the elevator. In our version of the problem, there is one single action which causes the elevator to *service* a given floor, with the effect that the unserved passengers whose origin is the serviced floor board the elevator, while the boarded passengers whose destination is the serviced floor unboard and become served. The task is to plan the elevator movement so that all passengers are eventually served.[14]

There are two variants of Miconic. In the 'simple' variant, a reward is received each time a passenger becomes served. In the 'hard' variant, the elevator also attempts to provide a range of priority services to passengers with special requirements: many passengers will prefer travelling in a single direction (either up or down) to their destination, certain passengers might be offered non-stop travel to their destination, and finally, passengers with disabilities or young children should be supervised inside the elevator by some other passenger (the supervisor) assigned to them. Here we omit the VIP and conflicting group services present in the original hard Miconic problem, as the reward formulae for those do not create additional difficulties.

Our formulation of the problem makes use of the same propositions as the PDDL description of Miconic used in the 2000 International Planning Competition: dynamic propositions record the floor the elevator is currently at and whether passengers are served or boarded, and static propositions record the origin and destination floors of passengers, as well as the categories (non-stop, direct-travel, supervisor, supervised) the passengers fall in. However, our formulation differs from the PDDL description in two interesting ways. Firstly, since we use *rewards* instead of goals, we are able to find a preferred solution even when all goals cannot simultaneously be satisfied. Secondly, because priority services are naturally described in terms of *non-Markovian* rewards, we are able to use the same action description for both the simple and hard versions, whereas the PDDL description of hard miconic requires additional actions (up, down) and complex preconditions to monitor the satisfaction of priority service constraints. The reward schemes for Miconic can be encapsulated through four different types of reward formula.

1. In the simple variant, a reward is received the first time each passenger $P_i$ is served:

14. We have experimented with stochastic variants of Miconic where passengers have some small probability of desembarking at the wrong floor. However, we find it more useful to present results for the deterministic version since it is closer to the Miconic deterministic planning benchmark and since, as we have shown before, rewards have a far more crucial impact than dynamics on the relative performance of the methods.

**PLTL:** $ServedP_i \wedge \circleddash \boxminus \neg ServedP_i$

**\$FLTL:** $\neg ServedP_i \, \mathsf{U} \, (ServedP_i \wedge \$)$

2. Next, a reward is received each time a non-stop passenger $P_i$ is served in one step after boarding the elevator:

   **PLTL:** $NonStopP_i \wedge \circleddash \circleddash \neg BoardedP_i \wedge \circleddash \circleddash \neg ServedP_i \wedge ServedP_i$

   **\$FLTL:** $\Box((NonStopP_i \wedge \neg BoardedP_i \wedge \neg ServedP_i \wedge \bigcirc\bigcirc ServedP_i) \rightarrow \bigcirc\bigcirc \$)$

3. Then, a reward is received each time a supervised passenger $P_i$ is served while having been accompanied at all times inside the elevator by his supervisor[15] $P_j$:

   **PLTL:** $SupervisedP_i \wedge SupervisorP_jP_i \wedge ServedP_i \wedge$
   $\circleddash \boxminus \neg ServedP_i \wedge \boxminus(BoardedP_i \rightarrow BoardedP_j)$

   **\$FLTL:** $\neg ServedP_i \, \mathsf{U} \, ((BoardedP_i \wedge SupervisedP_i \wedge \neg(BoardedP_j \wedge SupervisorP_jP_i) \wedge$
   $\neg ServedP_i) \vee (ServededP_i \wedge \$))$

4. Finally, reward is received each time a direct travel passenger $P_i$ is served while having travelled only in one direction since boarding, e.g., in the case of going up:

   **PLTL:** $DirectP_i \wedge ServedP_i \wedge \circleddash \neg ServedP_i \wedge$
   $((\bigvee_j \bigvee_{k>j} (AtFloor_k \wedge \circleddash AtFloor_j)) \, \mathsf{S} \, (BoardedP_i \wedge \circleddash \neg BoardedP_i))$

   **\$FLTL:** $\Box((DirectP_i \wedge BoardedP_i) \rightarrow (\neg ServedP_i \, \mathsf{U} \, ((\neg(\bigvee_j \bigvee_{k>i} AtFloor_j \wedge \bigcirc AtFloor_k) \wedge$
   $\neg ServedP_i) \vee (servedP_i \wedge \$))))$

   and similarly in the case of going down.

Experiments in this section were run on a Dual Pentium4 3.4GHz GNU/Linux 2.6.11 machine with 1GB of ram. We first experimented with the simple variant, giving a reward of 50 each time a passenger is first served. Figure 19 shows the CPU time taken by the various approaches to solve random problems with an increasing number $n$ of floors and passengers, and Figure 20 shows the number of states expanded when doing so. Each data point corresponds to just one random problem. To be fair with the structured approach, we ran PLTLSTR(A) which is able to exploit reachability from the start state. A first observation is that although PLTLSTR(A) does best for small values of $n$, it quickly runs out of memory. PLTLSTR(A) and PLTLSIM both need to track formulae of the form $\circleddash \boxminus \neg ServedP_i$ while PLTLSIM does not, and we conjecture that this is why they run out of memory earlier. A second observation is that attempts at PLTL minimisation do not pay very much here. While PLTLMIN has reduced memory because it tracks fewer subformulae, the size of the MDP it produces is identical to the size of the PLTLSIM MDP and larger than that of the FLTL MDP. This size increase is due to the fact that PLTL approaches label differently e-states in which the same passengers are served, depending on who has just become served (for those passengers, the reward formula is true at the e-state). In contrast, our FLTL implementation with progression one step ahead labels all these e-states with the reward

15. To understand the \$FLTL formula, observe that we get a reward iff $(BoardedP_i \wedge SupervisedP_i) \rightarrow (BoardedP_j \wedge SupervisorP_jP_i)$ holds until $ServedP_i$ becomes true, and recall that the formula $\neg q \, \mathsf{U} \, ((\neg p \wedge \neg q) \vee (q \wedge \$))$ rewards the holding of $p$ until the occurrence of $q$.

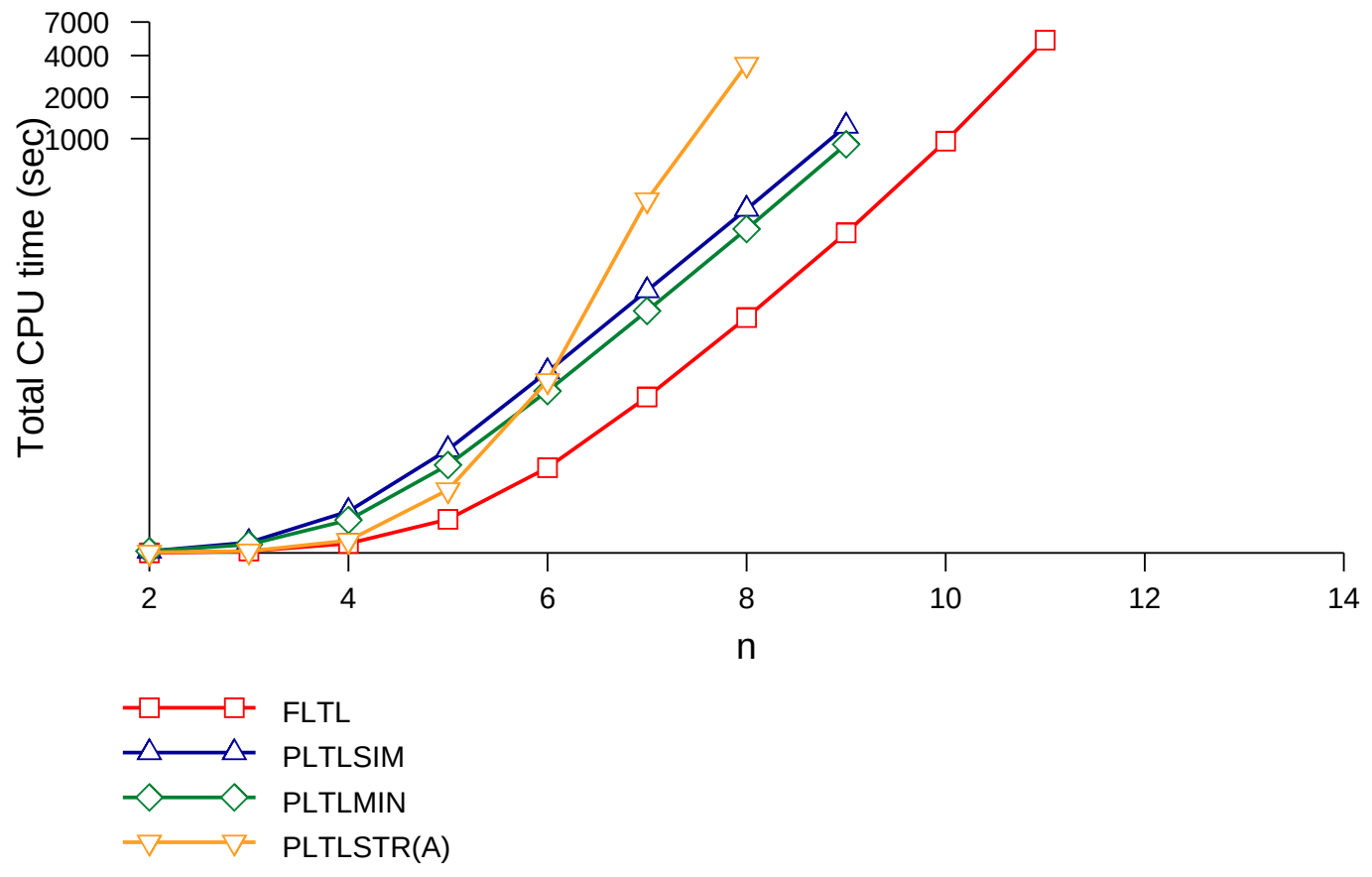


Figure 19: Simple Miconic - Run Time

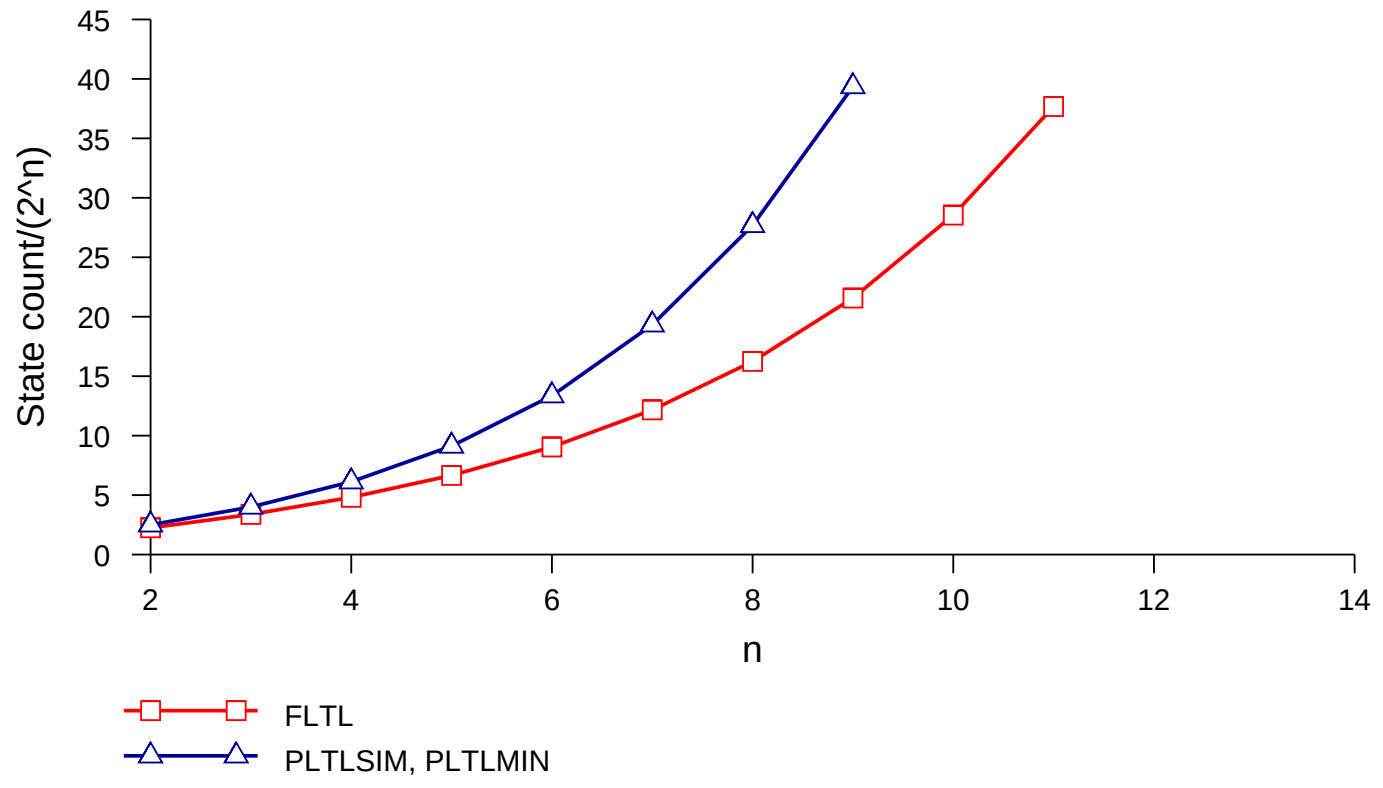


Figure 20: Simple Miconic - Number of Expanded States

formulae relevant to the passengers that still need to be served, the other formulae having progressed to $\top$. The gain in number of expanded states materialises into run time gains, resulting in FLTL eventually taking the lead.

Our second experiment illustrates the benefits of using an even extremely simple admissible heuristic in conjunction with FLTL. Our heuristic is applicable to discounted stochastic shortest path problems, and discounts rewards by the shortest time in the future in which they are possible. Here it simply amounts to assigning a fringe state to a value of 50 times the number of still unserved passengers (discounted once), and results in avoiding floors at which no passenger is waiting and which are not the destination of a boarded passenger. Figures 21 and 22 compare the run time and number of states expanded by FLTL when used in conjunction with value iteration (valIt) to when it is used in conjunction with an LAO*

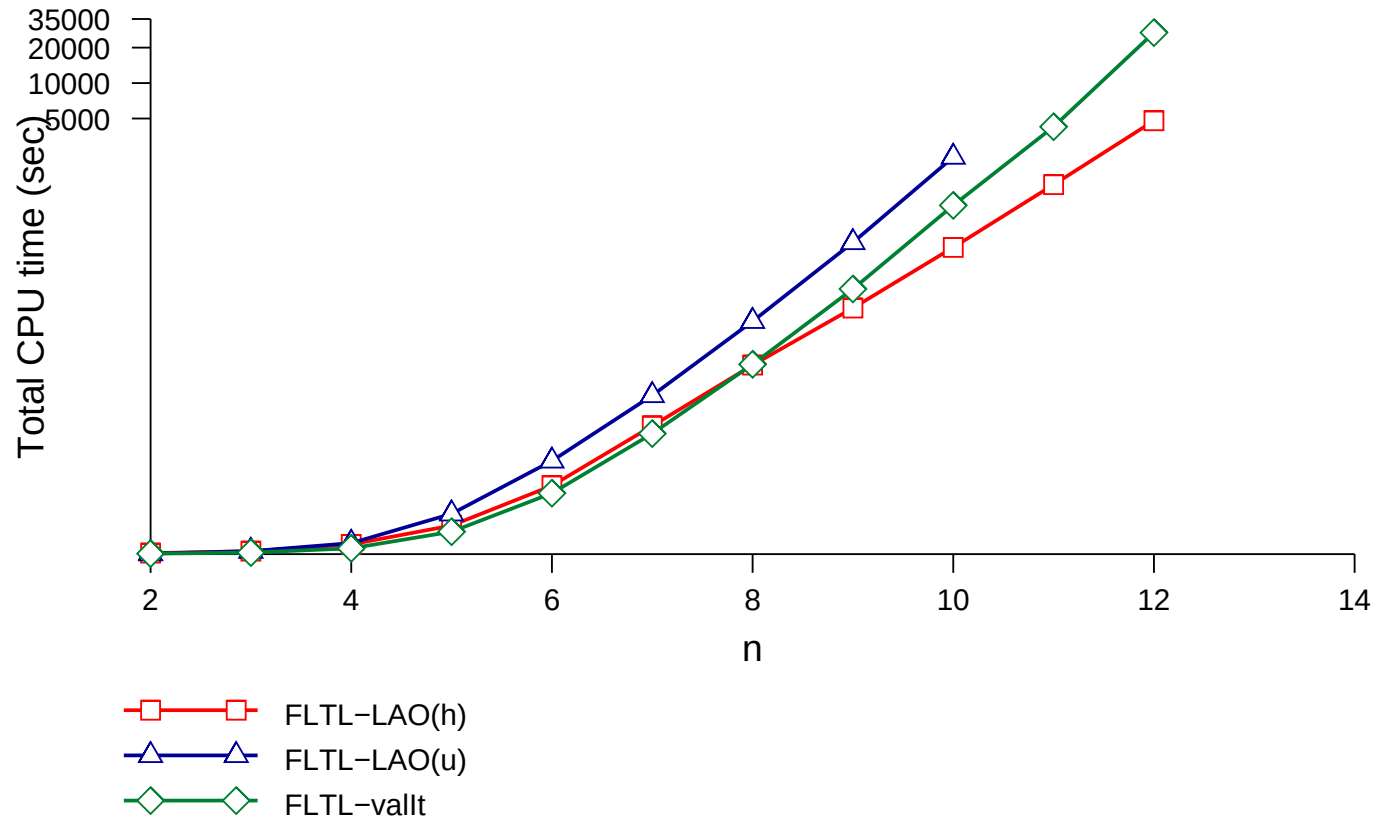


Figure 21: Effect of a Simple Heuristic on Run Time

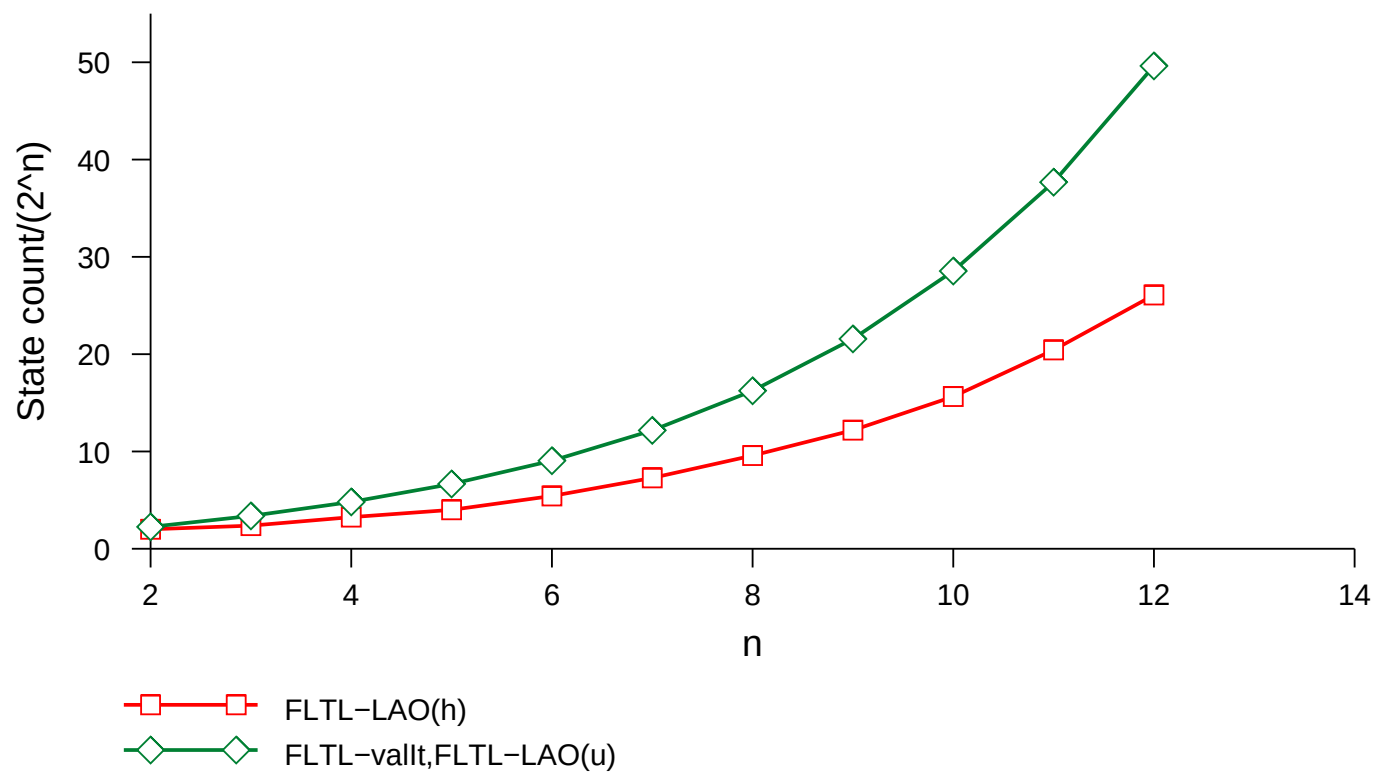


Figure 22: Effect of a Simple Heuristic on the Number of Expanded States

search informed by the above heuristic (LAO(h)). Uninformed LAO* (LAO*(u), i.e. LAO* with a heuristic of $50 * n$ at each node) is also included as a reference point to show the overhead induced by heuristic search. As can be seen from the graphs, the heuristic search generates significantly fewer states and this eventually pays in terms of run time.

In our final experiment, we considered the hard variant, giving a reward of 50 as before for service (1), a reward of 2 for non-stop travel (2), a reward of 5 for appropriate supervision (3), and a reward of 10 for direct travel (2). Regardless of the number $n$ of floors and passengers, problems only feature a single non-stop traveller, a third of passengers require supervision, and only half the passengers care about traveling direct. CPU time and number of states expanded are shown in Figures 23 and 24, respectively. As in the simple case, pltlsim and pltlstr quickly run out of memory. Formulae of type (2) and (3) create too many additional variables to track for these approaches, and the problem does not seem

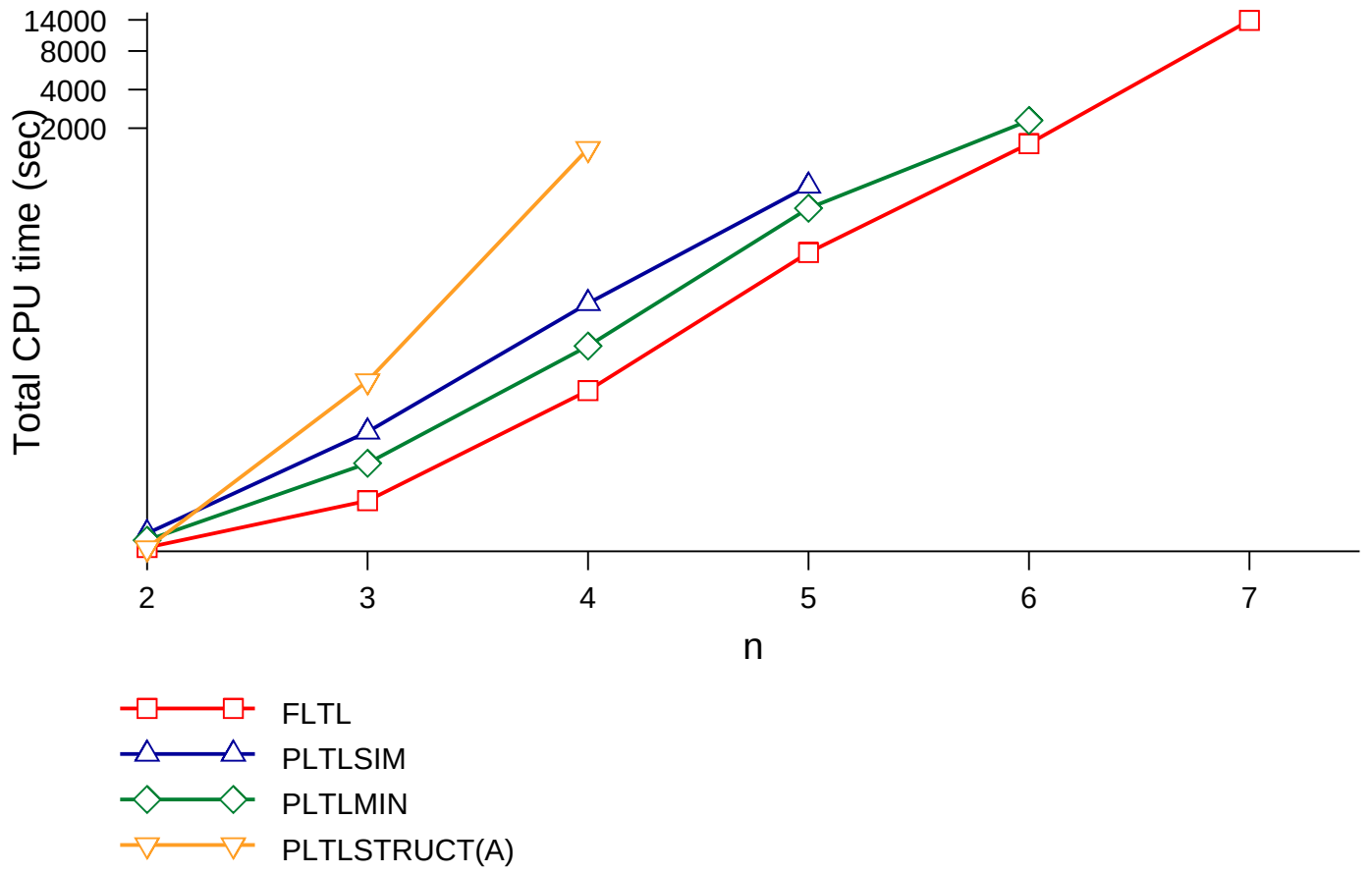


Figure 23: Hard Miconic - Run Time

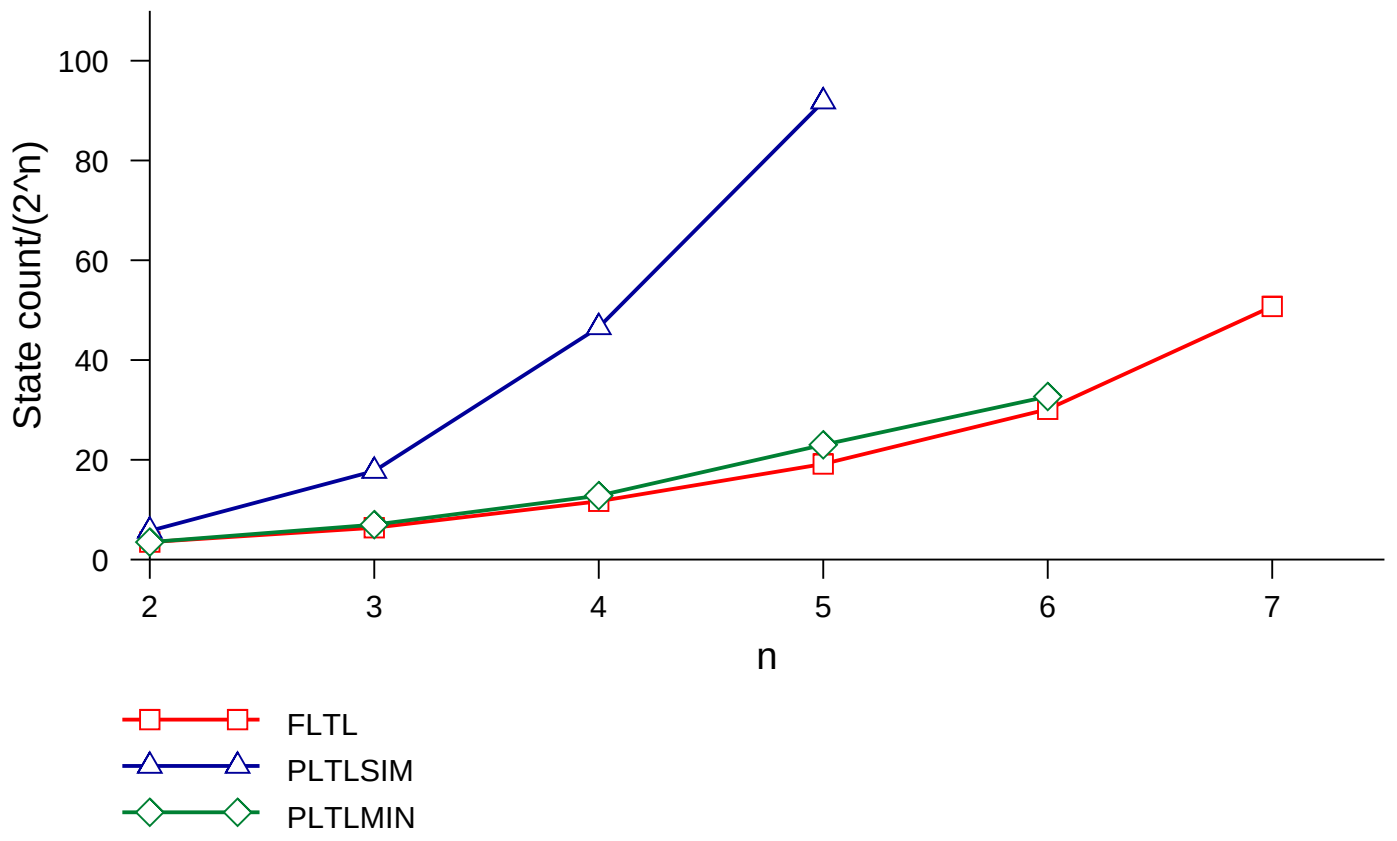


Figure 24: Hard Miconic - Number of Expanded States

to exhibit enough structure to help pltlstr. fltl remains the fastest. Here, this does not seem to be so much due to the size of the generated MDP which is just slightly below that of the pltlmin MDP, but rather to the overhead incurred by minimisation. Another observation arising from this experiment is that only very small instances can be handled in comparison to the classical planning version of the problem solved by state of the art optimal classical planners. For example, at the 2000 International Planning Competition, the PropPlan planner (Fourman, 2000) optimally solved instances of hard Miconic with 20 passengers and 40 floors in about 1000 seconds on a much less powerful machine.

## 7. nmrdpp in the Probabilistic Planning Competition

We now report on the behaviour of nmrdpp in the probabilistic track of the 4th International Planning Competition (IPC-4). Since the competition did not feature non-Markovian rewards, our original motivation in taking part was to further compare the solution methods implemented in nmrdpp in a *Markovian* setting. This objective largely underestimated the challenges raised by merely getting a planner ready for a competition, especially when that competition is the first of its kind. In the end, we decided that successfully preparing nmrdpp to attempt all problems in the competition using *one* solution method (and possibly search control knowledge), would be an honorable result.

The most crucial problem we encountered was the translation of PPDDL (Younes & Littman, 2004), the probabilistic variant of PDDL used as input language for the competition, into nmrdpp's ADD-based input language. While translating PPDDL into ADDs is possible in theory, devising a translation which is practical enough for the need of the competition (small number of variables, small, quickly generated, and easily manipulable ADDs) is another matter. mtbdd, the translator kindly made available to participants by the competition organisers, was not always able to achieve the required efficiency. At other times, the translation was quick but nmrdpp was unable to use the generated ADDs efficiently. Consequently, we implemented a state-based translator on top of the PDDL parser as a backup, and opted for a state-based solution method since it did not rely on ADDs and could operate with both translators.

The version of nmrdpp entered in the competition did the following:

1. Attempt to get a translation into ADDs using mtbdd, and if that proves infeasible, abort it and rely on the state-based translator instead.

2. Run fltl expansion of the state space, taking search control knowledge into account when available. Break after 10mn if not complete.

3. Run value iteration to convergence. Failing to achieve any useful result (e.g. because expansion was not complete enough to even reach a goal state), go back to step 2.

4. Run as many of the 30 trials as possible in the remaining time,[16] following the generated policy where defined, and falling back on the non-deterministic search control policy when available.

With Step 1 we were trying to maximise the instances in which the original ADD-based nmrdpp version could be run intact. In Step 3, it was decided not to use LAO* because when run with no good heuristic, it often incurs a significant overhead compared to value iteration.

The problems featured in the competition can be classified into goal-based or reward-based problems. In goal-based problems, a (positive) reward is only received when a goal state is reached. In reward-based problems, action performance may also incur a (usually negative) reward. Another orthogonal distinction can be made between problems from

16. On each given problem, planners had 15mn to run whatever computation they saw as appropriate (including parsing, pre-processing, and policy generation if any), and execute 30 trial runs of the generated policy from an initial state to a goal state.

domains that were not communicated in advance to the participants and those from domains that were. The latter consisted of variants of blocks world and logistics (or box world) problems, and gave the participating planners an opportunity to exploit knowledge of the domain, much as in the hand-coded deterministic planning track.

We decided to enroll NMRDPP in a control-knowledge mode and in a domain-independent mode. The only difference between the two modes is that the first uses FLTL search control knowledge written for the known domains as additional input. Our main concern in writing the control knowledge was to achieve a reasonable compromise between the size and effectiveness of the formulae. For the blocks world domain, in which the two actions pickup-from and putdown-to had a 25% chance of dropping the block onto the table, the control knowledge we used encoded a variant of the well-known GN1 near-optimal strategy for deterministic blocks world planning (Slaney & Thiébaux, 2001): whenever possible, try putting a clear block in its goal position, otherwise put an arbitrary clear block on the table. Because blocks get dropped on the table whenever an action fails, and because the success probabilities and rewards are identical across actions, optimal policies for the problem are essentially made up of optimal sequences of actions for the deterministic blocks world and there was little need for a more sophisticated strategy.[17] In the colored blocks world domain, where several blocks can share the same color and the goal only refers to the color of the blocks, the control knowledge selected an arbitrary goal state of the non-colored blocks world consistent with the colored goal specification, and then used the same strategy as for the non-colored blocks world. The performance of this strategy depends entirely on the goal-state selected and can therefore be arbitrarily bad.

Logistics problems from IPC-2 distinguish between airports and other locations within a city; trucks can drive between any two locations in a city and planes can fly between any two airports. In contrast, the box world only features cities, some of which have an airport, some of which are only accessible by truck. A priori, the map of the truck and plane connections is arbitrary. The goal is to get packages from their city of origin to their city of destination. Moving by truck has a 20% chance of resulting in reaching one of the three cities closest to the departure city rather than the intended one. The size of the box world search space turned out to be quite challenging for NMRDPP. Therefore, when writing search control knowledge, we gave up any optimality consideration and favored maximal pruning. We were helped by the fact that the box world generator produces problems with the following structure. Cities are divided into clusters, all of which are composed of at least one airport city. Furthermore each cluster has at least one hamiltonian circuit which trucks can follow. The control knowledge we used forced all planes but one, and all trucks but one in each cluster to be idle. In each cluster, the truck allowed to move could only attempt driving along the chosen hamiltonian circuit, picking up and dropping parcels as it went.

The planners participating in the competition are shown in Table 2. Planners E, G2, J1, and J2 are domain-specific: either they are tuned for blocks and box worlds, or they use domain-specific search control knowledge, or learn from examples. The other participating planners are domain-independent.

17. More sophisticated near-optimal strategies for deterministic blocks world exist (see Slaney & Thiébaux, 2001), but are much more complex to encode and might have caused time performance problems.

| Part. | Description | Reference |
|---|---|---|
| C | symbolic LAO* | (Feng & Hansen, 2002) |
| E* | first-order heuristic search in the fluent calculus | (Karabaev & Skvortsova, 2005) |
| G1 | NMRDPP without control knowledge | this paper |
| G2* | NMRDPP with control knowledge | this paper |
| J1* | interpreter of hand written classy policies | (Fern et al., 2004) |
| J2* | learns classy policies from random walks | (Fern et al., 2004) |
| J3 | version of FF replanning upon failure | (Hoffmann & Nebel, 2001) |
| P | mGPT: LRTDP with automatically extracted heuristics | (Bonet & Geffner, 2005) |
| Q | ProbaProp: conformant probabilistic planner | (Onder et al., 2006) |
| R | structured reachability analysis and structured PI | (Teichteil-Königsbuch & Fabiani, 2005) |

Table 2: Competition Participants. Domain-specific planners are starred

| dom | bw-c-nr | | | bw-nc-nr | bx-nr | | expl-bw | hanoise | zeno | tire-nr | |
|---|---|---|---|---|---|---|---|---|---|---|---|
| prob | 5 | 8 | 11 | 8 | 5-10 | 10-10 | 11 | 5-3 | 1-2-3-7 | 30-4 | total |
| **G2*** | **100** | **100** | **100** | **100** | **100** | **100** | | | | | **600** |
| J1* | 100 | 100 | 100 | 100 | 100 | 100 | | | | | 600 |
| J2* | 100 | 100 | 100 | 100 | 100 | 67 | | | | | 567 |
| E* | 100 | 100 | 100 | 100 | | | | | | | 400 |
| J3 | 100 | 100 | 100 | 100 | 100 | 100 | 9 | — | — | 23 | 632 |
| **G1** | | | | | | | — | **50** | **100** | **30** | **180** |
| R | | | | 3 | | | | 57 | 90 | 30 | 177 |
| P | | | | | | | | | 100 | 53 | 153 |
| C | | | | | | | | | 100 | ? | $\geq$ 100 |
| Q | | | | | | | | | 3 | 23 | 26 |

Table 3: Results for Goal-Based Problems. Domain-specific planners are starred. Entries are the percentage of runs in which the goal was reached. A blank indicates that the planner was unable to attempt the problem. A — indicates that the planner attempted the problem but was never able to achieve the goal. A ? indicates that the result is unavailable (due to a bug in the evaluation software, a couple of the results initially announced were found to be invalid).

| dom | bw-c-r | | | bw-nc-r | | | | | | bx-r | | | file | tire-r | |
|---|---|---|---|---|---|---|---|---|---|---|---|---|---|---|---|
| prob | 5 | 8 | 11 | 5 | 8 | 11 | 15 | 18 | 21 | 5-10 | 10-10 | 10-15 | 30-4 | 30-4 | total |
| J1* | 497 | 487 | 481 | 494 | 489 | 480 | 470 | 462 | 458 | 419 | 317 | 129 | | | 5183 |
| **G2*** | **495** | **486** | **480** | **495** | **490** | **480** | **468** | **352** | **286** | **438** | **376** | — | | | 4846 |
| E* | 496 | 492 | 486 | 495 | 490 | | | | | | | | | | 2459 |
| J2* | 497 | 486 | 482 | 495 | 490 | 480 | 468 | — | 455 | 376 | — | — | | | 4229 |
| J3 | 496 | 487 | 482 | 494 | 490 | 481 | — | — | 459 | 425 | 346 | 279 | 36 | — | 4475 |
| P | | | | 494 | 488 | 466 | 397 | | | 184 | — | | 58 | — | 2087 |
| C | | | | 495 | | | | | | | | | | ? | $\geq$ 495 |
| **G1** | | | | **495** | | | | | | | | | — | — | **495** |
| R | | | | 494 | | | | | | | | | | | 494 |
| Q | | | | 180 | | | | | | | | | | 11 | 191 |

Table 4: Results for Reward-Based Problems. Domain-specific planners are starred. Entries are the average reward achieved over the 30 runs. A blank indicates that the planner was unable to attempt the problem. A — indicates that the planner attempted the problem but did not achieve a strictly positive reward. A ? indicates that the result is unavailable.

Tables 3 and 4 show the results of the competition, which we extracted from the competition overview paper (Younes, Littman, Weissmann, & Asmuth, 2005) and from the competition web site `http://www.cs.rutgers.edu/~mlittman/topics/ipc04-pt/`. The first of those tables concerns goal-based problems and the second the reward-based problems. The entries in the tables represent the goal-achievement percentage or average reward achieved by the various planner versions (left-column) on the various problems (top two rows). Planners in the top part of the tables are domain-specific. Problems from the known domains lie on the left-hand side of the tables. The colored blocks world problems are bw-c-nr (goal-based version) and bw-c-r (reward version) with 5, 8, and 11 blocks. The non-colored blocks world problems are bw-nc-nr (goal-based version) with 8 blocks, and bw-nc-r (reward-based version) with 5, 8, 11, 15, 18, and 21 blocks. The box world problems are bx-nr (goal-based) and bx-r (reward-based), with 5 or 10 cities and 10 or 15 boxes. Problems from the unknown domains lie on the right hand side of the tables. They comprise: expl-bw, an exploding version of the 11 block blocks world problem in which putting down a block may destroy the object it is put on, zeno, a probabilistic variant of a zeno travel domain problem from the IPC-3 with 1 plane, 2 persons, 3 cities and 7 fuel levels, hanoise, a probabilistic variant of the tower of hanoi problem with 5 disks and 3 rods, file, a problem of putting 30 files in 5 randomly chosen folders, and tire, a variant a the tire world problem with 30 cities and spare tires at 4 of them, where the tire may go flat while driving.

Our planner NMRDPP in its G1 or G2 version, was able to attempt all problems, achieving a strictly positive reward in all but 4 of them. Not even FF (J3), the competition overall winner, was able to successfully attempt that many problems. NMRDPP performed particularly well on goal-based problems, achieving the goal in 100% of the runs except in expl-bw, hanoise, and tire-nr (note that for these three problems, the goal achievement probability of the optimal policy does not exceed 65%). No other planner outperformed NMRDPP on that scale. As pointed out before, FF behaves well on the probabilistic version of blocks and box world because the optimal policies are very close to those for the deterministic problem – Hoffmann (2002) analyses the reasons why the FF heuristic works well for traditional planning benchmarks such as blocks world and logistics. On the other hand, FF is unable to solve the unknown problems which have a different structure and require more substantial probabilistic reasoning, although these problems are easily solved by a number of participating planners. As expected, there is a large discrepancy between the version of NMRDPP allowed to use search control (G2) and the domain-independent version (G1). While the latter performs okay with the unknown goal-based domains, it is not able to solve any of the known ones. In fact, to except for FF, none of the participating domain-independent planners were able to solve these problems.

In the reward-based case, NMRDPP with control knoweldge behaves well on the known problems. Only the human-encoded policies (J1) performed better. Without control knowledge NMRDPP is unable to scale on those problems, while other participants such as FF and mGPT are. Furthermore NMRDPP appears to perform poorly on the two unknown problems. In both cases, this might be due to the fact that it fails to generate an optimal policy: suboptimal policies easily have a high negative score in these domains (see Younes et al., 2005). For r-tire, we know that NMRDPP did indeed generate a suboptimal policy. Additionally, it could be that NMRDPP was unlucky with the sampling-based policy evaluation process: in

tire-r in particular, there was a high variance between the costs of various trajectories in the optimal policy.

Alltogether, the competition results suggest that control knowledge is likely to be essential when solving larger problems (Markovian or not) with nmrdpp, and that, as has been observed with deterministic planners, approaches making use of control knowledge are quite powerful.

## 8. Conclusion, Related, and Future Work

In this paper, we have examined the problem of solving decision processes with non-Markovian rewards. We have described existing approaches which exploit a compact representation of the reward function to automatically translate the NMRDP into an equivalent process amenable to MDP solution methods. The computational model underlying this framework can be traced back to work on the relationship between linear temporal logic and automata in the areas of automated verification and model-checking (Vardi, 2003; Wolper, 1987). While remaining in this framework, we have proposed a new representation of non-Markovian reward functions and a translation into MDPs aimed at making the best possible use of state-based anytime heuristic search as the solution method. Our representation extends future linear temporal logic to express rewards. Our translation has the effect of embedding model-checking in the solution method. It results in an MDP of the minimal size achievable without stepping outside the anytime framework, and consequently in better policies by the deadline. We have described nmrdpp, a software platform that implements such approaches under a common interface, and which proved a useful tool in their experimental analysis. Both the system and the analysis are the first of their kind. We were able to identify a number of general trends in the behaviours of the methods and to provide advice as to which are the best suited to certain circumstances. For obvious reasons, our analysis has focused on artificial domains. Additional work should examine a wider range of domains of more practical interest, to see what form these results take in that context. Ultimately, we would like our analysis to help nmrdpp automatically select the most appropriate method. Unfortunately, because of the difficulty of translating between PLTL and $FLTL, it is likely that nmrdpp would still have to maintain both a PLTL and a $FLTL version of the reward formulae.

A detailed comparison of our approach to solving NMRDPs with existing methods (Bacchus et al., 1996, 1997) can be found in Sections 3.10 and 5. Two important aspects of future work would help take the comparison further. One is to settle the question of the appropriateness of our translation to structured solution methods. Symbolic implementations of the solution methods we consider, e.g. symbolic LAO* (Feng & Hansen, 2002), as well as formula progression in the context of symbolic state representations (Pistore & Traverso, 2001) could be investigated for that purpose. The other is to take advantage of the greater expressive power of $FLTL to consider a richer class of decision processes, for instance with uncertainty as to which rewards are received and when. Many extensions of the language are possible: adding eventualities, unrestricted negation, first-class reward propositions, quantitative time, etc. Of course, dealing with them via progression without backtracking is another matter.

We should investigate the precise relationship between our line of work and recent work on planning for temporally extended goals in non-deterministic domains. Of particular interest are 'weak' temporally extended goals such as those expressible in the Eagle language (Dal Lago et al., 2002), and temporally extended goals expressible in $\pi$-CTL* (Baral & Zhao, 2004). Eagle enables the expression of attempted reachability and maintenance goals of the form "try-reach $p$" and "try-maintain $p$", which add to the goals "do-reach $p$" and "do-maintain $p$" already expressible in CTL. The idea is that the generated policy should make every attempt at satisfying proposition $p$. Furthermore, Eagle includes recovery goals of the form "$g_1$ fail $g_2$", meaning that goal $g_2$ must be achieved whenever goal $g_1$ fails, and cyclic goals of the form "repeat $g$", meaning that $g$ should be achieved cyclically until it fails. The semantics of these goals is given in terms of variants of Büchi tree automata with preferred transitions. Dal Lago et al. (2002) present a planning algorithm based on symbolic model-checking which generates policies achieving those goals. Baral and Zhao (2004) describe $\pi$-CTL*, an alternative framework for expressing a subset of Eagle goals and a variety of others. $\pi$-CTL* is a variant of CTL* which allows for formulae involving two types of path quantifiers: quantifiers tied to the paths feasible under the generated policy, as is usual, but also quantifiers more generally tied to the paths feasible under any of the domain actions. Baral and Zhao (2004) do not present any planning algorithm. It would be very interesting to know whether Eagle and $\pi$-CTL* goals can be encoded as non-Markovian rewards in our framework. An immediate consequence would be that NMRDPP could be used to plan for them. More generally, we would like to examine the respective merits of non-deterministic planning for temporally extended goals and decision-theoretic planning with non-Markovian rewards.

In the pure probabilistic setting (no rewards), recent related research includes work on planning and controller synthesis for probabilistic temporally extended goals expressible in probabilistic temporal logics such as CSL or PCTL (Younes & Simmons, 2004; Baier et al., 2004). These logics enable expressing statements about the probability of the policy satisfying a given temporal goal exceeding a given threshold. For instance, Younes and Simmons (2004) describe a very general probabilistic planning framework, involving concurrency, continuous time, and temporally extended goals, rich enough to model generalised semi-Markov processes. The solution algorithms are not directly comparable to those presented here.

Another exciting future work area is the investigation of temporal logic formalisms for specifying heuristic functions for NMRDPs or more generally for search problems with temporally extended goals. Good heuristics are important to some of the solution methods we are targeting, and surely their value ought to depend on history. The methods we have described could be applicable to the description and processing of such heuristics. Related to this is the problem of extending search control knowledge to *fully* operate under the presence of temporally extended goals, rewards, and stochastic actions. A first issue is that branching or probabilistic logics such as CTL or PCTL variants should be preferred to FLTL when describing search control knowledge, because when stochastic actions are involved, search control often needs to refer to *some* of the possible futures and even to their probabilities.[18] Another major problem is that the GOALP modality, which is the key to the specification of reusable search control knowledge is interpreted with respect to

18. We would *not* argue, on the other hand, that CTL is necessary for representing non-Markovian *rewards*.

a fixed reachability goal[19] (Bacchus & Kabanza, 2000), and as such, is not applicable to domains with temporally extended goals, let alone rewards. Kabanza and Thiébaux (2005) present a first approach to search control in the presence of temporally extended goals in deterministic domains, but much remains to be done for a system like NMRDPP to be able to support a meaningful extension of GOALP.

Finally, let us mention that related work in the area of databases uses a similar approach to PLTLSTR to extend a database with auxiliary relations containing sufficient information to check temporal integrity constraints (Chomicki, 1995). The issues are somewhat different from those raised by NMRDPs: as there is only ever one sequence of databases, what matters is more the size of these auxiliary relations than avoiding making redundant distinctions.

## Acknowledgements

Many thanks to Fahiem Bacchus, Rajeev Goré, Marco Pistore, Ron van der Meyden, Moshe Vardi, and Lenore Zuck for useful discussions and comments, as well as to the anonymous reviewers and to David Smith for their thorough reading of the paper and their excellent suggestions. Sylvie Thiébaux, Charles Gretton, John Slaney, and David Price thank National ICT Australia for its support. NICTA is funded through the Australian Government's *Backing Australia's Ability* initiative, in part through the Australian Research Council. Froduald Kabanza is supported by the Canadian Natural Sciences and Engineering Research Council (NSERC).

## Appendix A. A Class of Reward-Normal Formulae

The existing decision procedure (Slaney, 2005) for determining whether a formula is reward-normal is guaranteed to terminate finitely, but involves the construction and comparison of automata and is rather intricate in practice. It is therefore useful to give a simple syntactic characterisation of a set of constructors for obtaining reward-normal formulae even though not all such formulae are so constructible.

We say that a formula is *material* iff it contains no \$ and no temporal operators – that is, the material formulae are the boolean combinations of atoms.

We consider four operations on behaviours representable by formulae of \$FLTL. Firstly, a behaviour may be delayed for a specified number of timesteps. Secondly, it may be made conditional on a material trigger. Thirdly, it may be started repeatedly until a material termination condition is met. Fourthly, two behaviours may be combined to form their union. These operations are easily realised syntactically by corresponding operations on formulae. Where $m$ is any material formula:

$$\begin{aligned}\mathsf{delay}[f] &= \bigcirc f \\ \mathsf{cond}[m, f] &= m \rightarrow f \\ \mathsf{loop}[m, f] &= f \,\mathsf{U}\, m \\ \mathsf{union}[f_1, f_2] &= f_1 \wedge f_2\end{aligned}$$

19. Where $f$ is an atemporal formula, GOALP($f$) is true iff $f$ is true of all goal states.

We have shown (Slaney, 2005) that the set of reward-normal formulae is closed under $\mathsf{delay}$, $\mathsf{cond}$ (for any material $m$), $\mathsf{loop}$ (for any material $m$) and $\mathsf{union}$, and also that the closure of $\{\$\}$ under these operations represents a class of behaviours closed under intersection and concatenation as well as union.

Many familiar reward-normal formulae are obtainable from \$ by applying the four operations. For example, $\Box(p \rightarrow \$)$ is $\mathsf{loop}[\bot, \mathsf{cond}[p, \$]]$. Sometimes a paraphrase is necessary. For example, $\Box((p \wedge \bigcirc q) \rightarrow \bigcirc \$)$ is not of the required form because of the $\bigcirc$ in the antecedent of the conditional, but the equivalent $\Box(p \rightarrow \bigcirc(q \rightarrow \$))$ is $\mathsf{loop}[\bot, \mathsf{cond}[p, \mathsf{delay}[\mathsf{cond}[q, \$]]]]$. Other cases are not so easy. An example is the formula $\neg p \,\mathsf{U}\, (p \wedge \$)$ which stipulates a reward the first time $p$ happens and which is not at all of the form suggested. To capture the same behaviour using the above operations requires a formula like $(p \rightarrow \$) \wedge (\bigcirc(p \rightarrow \$) \,\mathsf{U}\, p)$.

## Appendix B. Proofs of Theorems

**Property 1** Where $b \Leftrightarrow (\Gamma(i) \in B)$, $(\Gamma, i) \models_B f$ iff $(\Gamma, i+1) \models_B \mathrm{Prog}(b, \Gamma_i, f)$.

**Proof:** Induction on the structure of $f$. There are several base cases, all fairly trivial. If $f = \top$ or $f = \bot$ there is nothing to prove, as these progress to themselves and hold everywhere and nowhere respectively. If $f = p$ then if $f$ holds in $\Gamma_i$ then it progresses to $\top$ which holds in $\Gamma_{i+1}$ while if $f$ does not hold in $\Gamma_i$ then it progresses to $\bot$ which does not hold in $\Gamma_{i+1}$. The case $f = \neg p$ is similar. In the last base case, $f = \$$. Then the following are equivalent:

$(\Gamma, i) \models_B f$
$\Gamma(i) \in B$
$b$
$\mathrm{Prog}(b, \Gamma_i, f) = \top$
$(\Gamma, i+1) \models_B \mathrm{Prog}(b, \Gamma_i, f)$

Induction case 1: $f = g \wedge h$. The following are equivalent:

$(\Gamma, i) \models_B f$
$(\Gamma, i) \models_B g$ and $(\Gamma, i) \models_B h$
$(\Gamma, i+1) \models_B \mathrm{Prog}(b, \Gamma_i, g)$ and $(\Gamma, i+1) \models_B \mathrm{Prog}(b, \Gamma_i, h)$ (by induction hypothesis)
$(\Gamma, i+1) \models_B \mathrm{Prog}(b, \Gamma_i, g) \wedge \mathrm{Prog}(b, \Gamma_i, h)$
$(\Gamma, i+1) \models_B \mathrm{Prog}(b, \Gamma_i, f)$

Induction case 2: $f = g \vee h$. Analogous to case 1.

Induction case 3: $f = \bigcirc g$. Trivial by inspection of the definitions.

Induction case 4: $f = g \,\mathsf{U}\, h$. Then $f$ is logically equivalent to $h \vee (g \wedge \bigcirc(g \,\mathsf{U}\, h)$ which by cases 1, 2 and 3 holds at stage $i$ of $\Gamma$ for behaviour $B$ iff $\mathrm{Prog}(b, \Gamma_i, f)$ holds at stage i+1. □

**Theorem 1** *Let $f$ be reward-normal, and let $\langle f_0, f_1, \ldots \rangle$ be the result of progressing it through the successive states of a sequence $\Gamma$. Then, provided no $f_i$ is $\bot$, for all $i$ $\mathrm{Rew}(\Gamma_i, f_i)$ iff $\Gamma(i) \in B_f$.*

**Proof:** First, by the definition of reward-normality, if $f$ is reward-normal then $\Gamma \models_B f$ iff for all $i$, if $\Gamma(i) \in B_f$ then $\Gamma(i) \in B$. Next, if $\Gamma \models_B f$ then progressing $f$ through $\Gamma$ according to $B$ (that is, letting each $b_i$ be true iff $\Gamma(i) \in B$) cannot lead to a contradiction because by Property 1, progression is truth-preserving.

It remains, then, to show that if $\Gamma \not\models_B f$ then progressing $f$ through $\Gamma$ according to $B$ must lead eventually to $\bot$. The proof of this is by induction on the structure of $f$ and as usual the base case in which $f$ is a literal (an atom, a negated atom or $\top$, $\bot$ or \$) is trivial.

Case $f = g \wedge h$. Suppose $\Gamma \not\models_B f$. Then either $\Gamma \not\models_B g$ or $\Gamma \not\models_B h$, so by the induction hypothesis either $g$ or $h$ progresses eventually to $\bot$, and hence so does their conjunction.

Case $f = g \vee h$. Suppose $\Gamma \not\models_B f$. Then both $\Gamma \not\models_B g$ and $\Gamma \not\models_B h$, so by the induction hypothesis each of $g$ and $h$ progresses eventually to $\bot$. Suppose without loss of generality that $g$ does not progress to $\bot$ before $h$ does. Then at some point $g$ has progressed to some formula $g'$ and $f$ has progressed to $g' \vee \bot$ which simplifies to $g'$. Since $g'$ also progresses to $\bot$ eventually, so does $f$.

Case $f = \bigcirc g$. Suppose $\Gamma \not\models_B f$. Let $\Gamma = \Gamma_0; \Delta$ and let $B' = \{\gamma | \Gamma_0; \gamma \in B\}$. Then $\Delta \not\models_{B'} g$, so by the induction hypothesis $g$ progressed through $\Delta$ according to $B'$ eventually reaches $\bot$. But The progression of $f$ through $\Gamma$ according to $B$ is exactly the same after the first step, so that too leads to $\bot$.

Case $f = g \mathsf{U} h$. Suppose $\Gamma \not\models_B f$. Then there is some $j$ such that $(\Gamma, j) \not\models_B g$ and for all $i \leq j$, $(\Gamma, i) \not\models_B h$. We proceed by induction on $j$. In the base case $j = 0$, and both $\Gamma \not\models_B g$ and $\Gamma \not\models_B h$ whence by the main induction hypothesis both $g$ and $h$ will eventually progress to $\bot$. Thus $h \vee (g \wedge f')$ progresses eventually to $\bot$ for any $f'$, and in particular for $f' = \bigcirc f$, establishing the base case. For the induction case, suppose $\Gamma \models_B g$ (and of course $\Gamma \not\models_B h$). Since $f$ is equivalent to $h \vee (g \wedge \bigcirc f)$ and $\Gamma \not\models_B f$, $\Gamma \not\models_B h$ and $\Gamma \models_B g$, clearly $\Gamma \not\models_B \bigcirc f$. Where $\Delta$ and $B'$ are as in the previous case, therefore, $\Delta \not\models_{B'} f$ and the failure occurs at stage $j-1$ of $\Delta$. Therefore the hypothesis of the induction on $j$ applies, and $f$ progressed through $\Delta$ according to $B'$ goes eventually to $\bot$, and so $f$ progressed through $\Gamma$ according to $B$ goes similarly to $\bot$. □

**Theorem 3** *Let $S'$ be the set of e-states in an equivalent MDP $D'$ for $D = \langle S, s_0, A, \Pr, R \rangle$. $D'$ is minimal iff every e-state in $S'$ is reachable and $S'$ contains no two distinct e-states $s'_1$ and $s'_2$ with $\tau(s'_1) = \tau(s'_2)$ and $\mu(s'_1) = \mu(s'_2)$.*

**Proof:** Proof is by construction of the canonical equivalent MDP $D^c$. Let the set of finite prefixes of state sequences in $\widetilde{D}(s_0)$ be partitioned into equivalence classes, where $\Gamma 1(i) \equiv \Gamma 2(j)$ iff $\Gamma 1_i = \Gamma 2_j$ and for all $\Delta \in S^*$ such that $\Gamma 1(i); \Delta \in \widetilde{D}(s_0)$, $R(\Gamma 1(i); \Delta) = R(\Gamma 2(j); \Delta)$. Let $[\Gamma(i)]$ denote the equivalence class of $\Gamma(i)$. Let $\mathcal{E}$ be the set of these equivalence classes. Let $\mathcal{A}$ be the function that takes each $[\Gamma(i)]$ in $\mathcal{E}$ to $A(\Gamma_i)$. For each $\Gamma(i)$ and $\Delta(j)$ and for each $a \in \mathcal{A}([\Gamma(i)])$, let $\mathcal{T}([\Gamma(i)], a, [\Delta(j)])$ be $\Pr(\Gamma_i, a, s)$ if $[\Delta(j)] = [\Gamma(i); \langle s \rangle]$. Otherwise let $\mathcal{T}([\Gamma(i)], a, [\Delta(j)]) = 0$. Let $\mathcal{R}([\Gamma(i)])$ be $R(\Gamma(i))$. Then note the following four facts:

1. Each of the functions $A$, $T$ and $R$ is well-defined.

2. $D^c = \langle \mathcal{E}, [\langle s_0 \rangle], \mathcal{A}, \mathcal{T}, \mathcal{R} \rangle$ is an equivalent MDP for $D$ with $\tau([\Gamma(i)]) = \Gamma_i$.

3. For any equivalent MDP $D''$ of $D$ there is a mapping from a subset of the states of $D''$ onto $\mathcal{E}$.

4. $D'$ satisfies the condition that every e-state in $S'$ is reachable and $S'$ contains no two distinct e-states $s'_1$ and $s'_2$ with $\tau(s'_1) = \tau(s'_2)$ and $\mu(s'_1) = \mu(s'_2)$ iff $D^c$ is isomorphic to $D'$.

What fact 1 above amounts to is that if $\Gamma1(i) \equiv \Gamma2(j)$ then it does not matter which of the two sequences is used to define $\mathcal{A}$, $\mathcal{T}$ and $\mathcal{R}$ of their equivalence class. In the cases of $\mathcal{A}$ and $\mathcal{T}$ this is simply that $\Gamma1_i = \Gamma2_j$. In the case of $\mathcal{R}$, it is the special case $\Delta = \langle \Gamma1_i \rangle$ of the equality of rewards over extensions.

Fact 2 is a matter of checking that the four conditions of Definition 1 hold. Of these, conditions 1 ($\tau([s_0]) = s_0$) and 2 ($\mathcal{A}([\Gamma(i)]) = A(\Gamma_i)$) hold trivially by the construction. Condition 4 says that for any feasible state sequence $\Gamma \in \widetilde{D}(s_0)$, we have $\mathcal{R}([\Gamma(i)]) = R(\Gamma(i))$ for all $i$. This also is given in the construction. Condition 3 states:

> For all $s_1, s_2 \in S$, if there is $a \in A(s_1)$ such that $\Pr(s_1, a, s_2) > 0$, then for all $\Gamma(i) \in \widetilde{D}(s_0)$ such that $\Gamma_i = s_1$, there exists a unique $[\Delta(j)] \in \mathcal{E}$, $\Delta_j = s_2$, such that for all $a \in \mathcal{A}([\Gamma(i)])$, $\mathcal{T}([\Gamma(i)], a, [\Delta[j]]) = \Pr(s_1, a, s_2)$.

Suppose $\Pr(s_1, \alpha, s_2) > 0$, $\Gamma(i) \in \widetilde{D}(s_0)$ and $\Gamma_i = s_1$. Then the required $\Delta(j)$ is $\Gamma(i); \langle s_2 \rangle$, and of course $\mathcal{A}([\Gamma(i)]) = A(\Gamma_i)$, so the required condition reads:

> $[\Gamma(i); \langle s_2 \rangle]$ is the unique element $X$ of $\mathcal{E}$ with $\tau(X) = s_2$ such that for all $a \in A(\Gamma_i)$, $\mathcal{T}([\Gamma(i)], a, X) = \Pr(s_1, a, s_2)$.

To establish existence, we need that if $a \in A(\Gamma_i)$ then $\mathcal{T}([\Gamma(i)], a, [\Gamma(i); \langle s_2 \rangle]) = \Pr(\Gamma_i, a, s_2)$, which is immediate from the definition of $\mathcal{T}$ above. To establish uniqueness, suppose that $\tau(X) = s_2$ and $\mathcal{T}([\Gamma(i)], a, X) = \Pr(s_1, a, s_2)$ for all actions $a \in A(\Gamma_i)$. Since $\Pr(s_1, \alpha, s_2) > 0$, the transition probability from $[\Gamma(i)]$ to $X$ is nonzero for some action, so by the definition of $\mathcal{T}$, $X$ can only be $[\Gamma(i); \langle s_2 \rangle]$.

Fact 3 is readily observed. Let $M$ be any equivalent MDP for $D$. For any states $s_1$ and $s_2$ of $D$, and any state $X$ of $M$ such that $\tau(X) = s_1$ there is at most one state $Y$ of $M$ with $\tau(Y) = s_2$ such that some action $a \in A(s_1)$ gives a nonzero probability of transition from $X$ to $Y$. This follows from the uniqueness part of condition 3 of Definition 1 together with the fact that the transition function is a probability distribution (sums to 1). Therefore for any given finite state sequence $\Gamma(i)$ there is at most one state of $M$ reached from the start state of $M$ by following $\Gamma(i)$. Therefore $M$ induces an equivalence relation $\approx_M$ on $S^*$: $\Gamma(i) \approx_M \Delta(j)$ iff they lead to the same state of $M$ (the sequences which are not feasible in $M$ may all be regarded as equivalent under $\approx_M$). Each reachable state of $M$ has associated with it a nonempty equivalence class of finite sequences of states of $D$. Working through the definitions, we may observe that $\approx_M$ is a sub-relation of $\equiv$ (if $\Gamma(i) \approx_M \Delta(j)$ then $\Gamma(i) \equiv \Delta(j)$). Hence the function that takes the equivalence class under $\approx_M$ of each feasible sequence $\Gamma(i)$ to $[\Gamma(i)]$ induces a mapping $h$ (an epimorphism in fact) from the reachable subset of states of $M$ onto $\mathcal{E}$.

To establish Fact 4, it must be shown that in the case of $D'$ the mapping can be reversed, or that each equivalence class $[\Gamma(i)]$ in $D^c$ corresponds to exactly one element of $D'$. Suppose not (for contradiction). Then there exist sequences $\Gamma 1(i)$ and $\Gamma 2(j)$ in $\widetilde{D}(s_0)$ such that $\Gamma 1(i) \equiv \Gamma 2(j)$ but on following the two sequences from $s'_0$ we arrive at two different elements $s'_1$ and $s'_2$ of $D'$ with $\tau(s'_1) = \Gamma 1_i = \Gamma 2_j = \tau(s'_2)$ but with $\mu(s'_1) \neq \mu(s'_2)$. Therefore there exists a sequence $\Delta(k) \in \widetilde{D}(s)$ such that $R(\Gamma 1(i-1); \Delta(k)) \neq R(\Gamma 2(j-1); \Delta(k))$. But this contradicts the condition for $\Gamma 1(i) \equiv \Gamma 2(j)$. □

Theorem 3 follows immediately from facts 1–4.

**Theorem 4** *Let $D'$ be the translation of $D$ as in Definition 5. $D'$ is a blind minimal equivalent MDP for $D$.*

**Proof:** Reachability of all the e-states is obvious, as they are constructed only when reached. Each e-state is a pair $\langle s, \phi \rangle$ where $s$ is a state of $D$ and $\phi$ is a reward function specification. In fact, $s = \tau(\langle s, \phi \rangle)$ and $\phi$ determines a distribution of rewards over all continuations of the sequences that reach $\langle s, \phi \rangle$. That is, for all $\Delta$ in $S^*$ such that $\Delta_0 = s$, the reward for $\Delta$ is $\sum_{(f:r)\in\phi}\{r \mid \Delta \in B_f\}$. If $D'$ is not blind minimal, then there exist distinct e-states $\langle s, \phi \rangle$ and $\langle s, \phi' \rangle$ for which this sum is the same for all $\Delta$. But this makes $\phi$ and $\phi'$ semantically equivalent, contradicting the supposition that they are distinct.

□

## Appendix C. Random Problem Domains

Random problem domains are produced by first creating a random action specification defining the domain dynamics. Some of the experiments we conducted[20] also involved producing, in a second step, a random reward specification that had desired properties in relation to the generated dynamics.

The random generation of the domain dynamics takes as parameters the number $n$ of propositions in the domain and the number of actions to be produced, and starts by assigning some effects to each action such that each proposition is affected by exactly one action. For example, if we have 5 actions and 14 propositions, the first 4 actions may affect 3 propositions each, the 5th one only 2, and the affected propositions are all different. Once each action has some initial effects, we continue to add more effects one at a time, until a sufficient proportion of the state space is reachable – see "proportion reachable" parameter below. Each additional effect is generated by picking up a random action and a random proposition, and producing a random decision diagram according to the "uncertainty" and "structure" parameters below:

The Uncertainty parameter is the probability of a non zero/one value as a leaf node. An uncertainty of 1 will result in all leaf nodes having random values from a uniform distribution. An uncertainty of 0 will result in all leaf nodes having values 0 or 1 with an equal probability.

The Structure (or influence) parameter is the probability of a decision diagram containing a particular proposition. So an influence of 1 will result in all decision diagrams

20. None of those are included in this paper, however.

including all propositions (and very unlikely to have significant structure), while 0 will result in decision diagrams that do not depend on the values of propositions.

The Proportion Reachable parameter is a lower bound on the proportion of the entire $2^n$ state space that is reachable from the start state. The algorithm adds behaviour until this lower bound is reached. A value of 1 will result in the algorithm running until the actions are sufficient to allow the entire state space to be reachable.

A reward specification can be produced with regard to the generated dynamics such that a specified number of the rewards are reachable and a specified number are unreachable. First, a decision diagram is produced to represent which states are reachable and which are not, given the domain dynamics. Next, a random path is taken from the root of this decision diagram to a true terminal if we are generating an attainable reward, or a false terminal if we are producing an unattainable reward. The propositions encountered on this path, both negated and not, form a conjunction that is the reward formula. This process is repeated until the desired number of reachable and unreachable rewards are obtained.

## References

AT&T Labs-Research (2000). Graphviz. Available from `http://www.research.att.com/sw/tools/graphviz/`.

Bacchus, F., Boutilier, C., & Grove, A. (1996). Rewarding behaviors. In *Proc. American National Conference on Artificial Intelligence (AAAI)*, pp. 1160–1167.

Bacchus, F., Boutilier, C., & Grove, A. (1997). Structured solution methods for non-Markovian decision processes. In *Proc. American National Conference on Artificial Intelligence (AAAI)*, pp. 112–117.

Bacchus, F., & Kabanza, F. (1998). Planning for temporally extended goals. *Annals of Mathematics and Artificial Intelligence*, *22*, 5–27.

Bacchus, F., & Kabanza, F. (2000). Using temporal logic to express search control knowledge for planning. *Artificial Intelligence*, *116*(1-2).

Baier, C., Größer, M., Leucker, M., Bollig, B., & Ciesinski, F. (2004). Controller synthesis for probabilistic systems (extended abstract). In *Proc. IFIP International Conference on Theoretical Computer Science (IFIP TCS)*.

Baral, C., & Zhao, J. (2004). Goal specification in presence of nondeterministic actions. In *Proc. European Conference on Artificial Intelligence (ECAI)*, pp. 273–277.

Barto, A., Bardtke, S., & Singh, S. (1995). Learning to act using real-time dynamic programming. *Artificial Intelligence*, *72*, 81–138.

Bonet, B., & Geffner, H. (2003). Labeled RTDP: Improving the convergence of real-time dynamic programming. In *Proc. International Conference on Automated Planning and Scheduling (ICAPS)*, pp. 12–21.

Bonet, B., & Geffner, H. (2005). mGPT: A probabilistic planner based on heuristic search. *Journal of Artificial Intelligence Research*, *24*, 933–944.

Boutilier, C., Dean, T., & Hanks, S. (1999). Decision-theoretic planning: Structural assumptions and computational leverage. In *Journal of Artificial Intelligence Research*, Vol. 11, pp. 1–94.

Boutilier, C., Dearden, R., & Goldszmidt, M. (2000). Stochastic dynamic programming with factored representations. *Artificial Intelligence*, *121*(1-2), 49–107.

Calvanese, D., De Giacomo, G., & Vardi, M. (2002). Reasoning about actions and planning in LTL action theories. In *Proc. International Conference on the Principles of Knowledge Representation and Reasoning (KR)*, pp. 493–602.

Cesta, A., Bahadori, S., G, C., Grisetti, G., Giuliani, M., Loochi, L., Leone, G., Nardi, D., Oddi, A., Pecora, F., Rasconi, R., Saggase, A., & Scopelliti, M. (2003). The RoboCare project. Cognitive systems for the care of the elderly. In *Proc. International Conference on Aging, Disability and Independence (ICADI)*.

Chomicki, J. (1995). Efficient checking of temporal integrity constraints using bounded history encoding. *ACM Transactions on Database Systems*, *20*(2), 149–186.

Dal Lago, U., Pistore, M., & Traverso, P. (2002). Planning with a language for extended goals. In *Proc. American National Conference on Artificial Intelligence (AAAI)*, pp. 447–454.

Dean, T., Kaelbling, L., Kirman, J., & Nicholson, A. (1995). Planning under time constraints in stochastic domains. *Artificial Intelligence*, *76*, 35–74.

Dean, T., & Kanazawa, K. (1989). A model for reasoning about persistance and causation. *Computational Intelligence*, *5*, 142–150.

Drummond, M. (1989). Situated control rules. In *Proc. International Conference on the Principles of Knowledge Representation and Reasoning (KR)*, pp. 103–113.

Emerson, E. A. (1990). Temporal and modal logic. In *Handbook of Theoretical Computer Science*, Vol. B, pp. 997–1072. Elsevier and MIT Press.

Feng, Z., & Hansen, E. (2002). Symbolic LAO* search for factored Markov decision processes. In *Proc. American National Conference on Artificial Intelligence (AAAI)*, pp. 455–460.

Feng, Z., Hansen, E., & Zilberstein, S. (2003). Symbolic generalization for on-line planning. In *Proc. Conference on Uncertainty in Artificial Intelligence (UAI)*, pp. 209–216.

Fern, A., Yoon, S., & Givan, R. (2004). Learning domain-specific knowledge from random walks. In *Proc. International Conference on Automated Planning and Scheduling (ICAPS)*, pp. 191–198.

Fourman, M. (2000). Propositional planning. In *Proc. AIPS Workshop on Model-Theoretic Approaches to Planning*, pp. 10–17.

Gretton, C., Price, D., & Thiébaux, S. (2003a). Implementation and comparison of solution methods for decision processes with non-Markovian rewards. In *Proc. Conference on Uncertainty in Artificial Intelligence (UAI)*, pp. 289–296.

Gretton, C., Price, D., & Thiébaux, S. (2003b). NMRDPP: a system for decision-theoretic planning with non-Markovian rewards. In *Proc. ICAPS Workshop on Planning under Uncertainty and Incomplete Information*, pp. 48–56.

Haddawy, P., & Hanks, S. (1992). Representations for decision-theoretic planning: Utility functions and deadline goals. In *Proc. International Conference on the Principles of Knowledge Representation and Reasoning (KR)*, pp. 71–82.

Hansen, E., & Zilberstein, S. (2001). LAO*: A heuristic search algorithm that finds solutions with loops. *Artificial Intelligence*, *129*, 35–62.

Hoey, J., St-Aubin, R., Hu, A., & Boutilier, C. (1999). SPUDD: stochastic planning using decision diagrams. In *Proc. Conference on Uncertainty in Artificial Intelligence (UAI)*, pp. 279–288.

Hoffmann, J. (2002). Local search topology in planning benchmarks: A theoretical analysis. In *Proc. International Conference on AI Planning and Scheduling (AIPS)*, pp. 92–100.

Hoffmann, J., & Nebel, B. (2001). The FF planning system: Fast plan generation through heuristic search. *Journal of Artificial Intelligence Research*, *14*, 253–302.

Howard, R. (1960). *Dynamic Programming and Markov Processes.* MIT Press, Cambridge, MA.

Kabanza, F., & Thiébaux, S. (2005). Search control in planning for temporally extended goals. In *Proc. International Conference on Automated Planning and Scheduling (ICAPS)*, pp. 130–139.

Karabaev, E., & Skvortsova, O. (2005). A Heuristic Search Algorithm for Solving First-Order MDPs. In *Proc. Conference on Uncertainty in Artificial Intelligence (UAI)*, pp. 292–299.

Koehler, J., & Schuster, K. (2000). Elevator control as a planning problem. In *Proc. International Conference on AI Planning and Scheduling (AIPS)*, pp. 331–338.

Korf, R. (1990). Real-time heuristic search. *Artificial Intelligence*, *42*, 189–211.

Kushmerick, N., Hanks, S., & Weld, D. (1995). An algorithm for probabilistic planning. *Artificial Intelligence*, *76*, 239–286.

Lichtenstein, O., Pnueli, A., & Zuck, L. (1985). The glory of the past. In *Proc. Conference on Logics of Programs*, pp. 196–218. LNCS, volume 193.

Onder, N., Whelan, G. C., & Li, L. (2006). Engineering a conformant probabilistic planner. *Journal of Artificial Intelligence Research*, *25*, 1–15.

Pistore, M., & Traverso, P. (2001). Planning as model-checking for extended goals in non-deterministic domains. In *Proc. International Joint Conference on Artificial Intelligence (IJCAI-01)*, pp. 479–484.

Slaney, J. (2005). Semi-positive LTL with an uninterpreted past operator. *Logic Journal of the IGPL*, *13*, 211–229.

Slaney, J., & Thiébaux, S. (2001). Blocks world revisited. *Artificial Intelligence*, *125*, 119–153.

Somenzi, F. (2001). CUDD: CU Decision Diagram Package. Available from ftp://vlsi.colorado.edu/pub/.

Teichteil-Königsbuch, F., & Fabiani, P. (2005). Symbolic heuristic policy iteration algorithms for structured decision-theoretic exploration problems. In *Proc. ICAPS workshop on Planning under Uncertainty for Autonomous Systems*.

Thiébaux, S., Hertzberg, J., Shoaff, W., & Schneider, M. (1995). A stochastic model of actions and plans for anytime planning under uncertainty. *International Journal of Intelligent Systems*, *10*(2), 155–183.

Thiébaux, S., Kabanza, F., & Slaney, J. (2002a). Anytime state-based solution methods for decision processes with non-Markovian rewards. In *Proc. Conference on Uncertainty in Artificial Intelligence (UAI)*, pp. 501–510.

Thiébaux, S., Kabanza, F., & Slaney, J. (2002b). A model-checking approach to decision-theoretic planning with non-Markovian rewards. In *Proc. ECAI Workshop on Model-Checking in Artificial Intelligence (MoChArt-02)*, pp. 101–108.

Vardi, M. (2003). Automated verification = graph, logic, and automata. In *Proc. International Joint Conference on Artificial Intelligence (IJCAI)*, pp. 603–606. Invited paper.

Wolper, P. (1987). On the relation of programs and computations to models of temporal logic. In *Proc. Temporal Logic in Specification, LNCS 398*, pp. 75–123.

Younes, H. L. S., & Littman, M. (2004). PPDDL1.0: An extension to PDDL for expressing planning domains with probabilistic effects. Tech. rep. CMU-CS-04-167, School of Computer Science, Carnegie Mellon University, Pittsburgh, Pennsylvania.

Younes, H. L. S., Littman, M., Weissmann, D., & Asmuth, J. (2005). The first probabilistic track of the International Planning Competition. In *Journal of Artificial Intelligence Research*, Vol. 24, pp. 851–887.

Younes, H., & Simmons, R. G. (2004). Policy generation for continuous-time stochastic domains with concurrency. In *Proc. International Conference on Automated Planning and Scheduling (ICAPS)*, pp. 325–333.